\documentclass[12pt, a4paper]{article}

\usepackage{PRIMEarxiv}

\usepackage[utf8]{inputenc}
\usepackage{amsmath,amsfonts,amssymb}
\usepackage{amsthm}
\usepackage{algorithm,algorithmic}
\usepackage{bm}
\usepackage{graphicx}
\usepackage{float}
\usepackage{fancyhdr}
\usepackage{caption}
\usepackage{subcaption}
\usepackage{hyperref}
\usepackage{cite}
\usepackage{url}
\usepackage{mathtools}
\usepackage{bm}
\usepackage{multirow} 
\usepackage[nameinlink]{cleveref}
\usepackage{tikz}
\usetikzlibrary{shapes.geometric,arrows,positioning,fit}

\usepackage{pdflscape}  
\usepackage{lscape}     

\usepackage[numbers,sort&compress]{natbib}

\newtheorem{theorem}{Theorem}

\newtheorem{remark}{Remark}

\title{FedHK-MVFC: Federated Heat Kernel Multi-View Clustering}
\author{
    Kristina P. Sinaga\thanks{Corresponding author: Independent Researcher. ORCID: 0009-0000-6184-829X. Email: kristinasinaga41@gmail.com. Research interests include federated learning, multi-view clustering, privacy-preserving machine learning, heat kernel methods, quantum field theory applications in machine learning, and distributed intelligent systems.}\\
    \small Independent Researcher\\
    \small \texttt{kristinasinaga41@gmail.com}
}
\date{September 19, 2025}

\begin{document}

\maketitle

\begin{abstract}
In the realm of distributed artificial intelligence (AI) and privacy-focused medical applications, this paper proposes a multi-view clustering framework that links quantum field theory with federated healthcare analytics. The method uses heat kernel coefficients from spectral analysis to convert Euclidean distances into geometry-aware similarity measures that capture the structure of diverse medical data. The framework is presented through the heat kernel distance (HKD) transformation, which has convergence guarantees. Two algorithms have been developed: The first, Heat Kernel-Enhanced Multi-View Fuzzy Clustering (HK-MVFC), is used for central analysis. The second, Federated Heat Kernel Multi-View Fuzzy Clustering (FedHK-MVFC), is used for secure, privacy-preserving learning across hospitals. FedHK-MVFC uses differential privacy and secure aggregation to enable HIPAA-compliant collaboration. Tests on synthetic cardiovascular patient datasets demonstrate increased clustering accuracy, reduced communication, and retained efficiency compared to centralized methods. After being validated on 10,000 synthetic patient records across two hospitals, the methods proved useful for collaborative phenotyping involving electrocardiogram (ECG) data, cardiac imaging data, and behavioral data. The proposed methods' theoretical contributions include update rules with proven convergence, adaptive view weighting, and privacy-preserving protocols. These contributions establish a new standard for geometry-aware federated learning in healthcare, translating advanced mathematics into practical solutions for analyzing sensitive medical data while ensuring rigor and clinical relevance.


\end{abstract}

\section{Introduction}
\label{sec:Introduction}

The proliferation of heterogeneous data sources has catalyzed a surge of interest in multi-view (MV) clustering, a methodology in which a single entity is delineated by multiple distinct feature sets or "views." In such contexts, conventional clustering methods frequently prove inadequate, underscoring the necessity for advanced approaches that can seamlessly integrate complementary information across diverse perspectives while preserving coherence in the resulting partition. Furthermore, the rapid evolution of large-scale artificial intelligence (AI) systems—from centralized language models to decentralized, agent-based architectures—has fundamentally shifted the conceptualization and deployment of learning \cite{montes2019distributed}. In these novel frameworks, learning must occur not only across heterogeneous data views but also across distributed agents, with each agent operating within its own local context, resource constraints, and knowledge representation. This paradigm shift necessitates a theoretical and algorithmic framework that can accommodate structural diversity, communication limitations, and the intrinsic geometry of information flow.

In the context of prospective Internet applications, such as healthcare monitoring, smart cities, and distributed sensor networks, data is frequently collected and processed across a multitude of decentralized nodes \cite{kremenova2019decentralized}. The utilization of conventional clustering methods is hindered by the inherent characteristics and structures of the numerous sources of data. These data sources exhibit varied characteristics and structures, which hinder the efficacy of traditional clustering methods that rely on flat Euclidean representations. This challenge has been thoroughly documented in the research literature \cite{wang2004improving, jing2007entropy, cleuziou2009cofkm, fang2023comprehensive}. Healthcare serves as a prime example of this phenomenon. Patient data may be distributed across various hospitals, each collecting different types of information (e.g., imaging, physiological signals, and electronic health records) \cite{fang2024decentralised}. Conventional clustering methodologies that employ conventional Euclidean distance (CED) may encounter challenges in accurately capturing the intricate relationships and inherent geometries present in such heterogeneous datasets. Consequently, there is an urgent need for clustering algorithms that can effectively integrate and analyze multi-view data in a decentralized manner while respecting the privacy and autonomy of each data source.

Federated learning (FL) has emerged as a promising paradigm for collaborative model training across decentralized data sources while preserving privacy \cite{pei2024review}. In the state of Florida, a consortium of clients (e.g., hospitals) collaboratively trains a global model without sharing their raw data, thus addressing privacy concerns and regulatory constraints \cite{abbas2024federated}. However, extant FL approaches frequently depend on conventional distance metrics and are deficient in their inability to capture the intrinsic geometries of multi-view data. This limitation impedes their efficacy in scenarios where data is heterogeneous and distributed across multiple agents. 

\subsection{Research Problem and Objectives}

The fundamental research problem addressed in this work concerns the development of a unified framework for privacy-preserving multi-view clustering that simultaneously satisfies three critical requirements: (1) \textbf{geometric awareness}—capturing the intrinsic manifold structure of heterogeneous data that conventional Euclidean distance metrics fail to represent, (2) \textbf{federated scalability}—enabling distributed learning across autonomous agents without centralized data aggregation, and (3) \textbf{theoretical rigor}—providing mathematically proven convergence guarantees and privacy preservation mechanisms. Existing multi-view clustering approaches predominantly rely on Euclidean distance, which assumes linear relationships and uniform feature scaling, thereby failing to capture the complex geometric structures inherent in real-world heterogeneous datasets. Furthermore, federated learning frameworks for multi-view data lack principled mechanisms for aggregating complementary information across distributed views while maintaining privacy and communication efficiency.

To address these challenges, this research pursues the following specific objectives: \textit{First}, we develop a heat kernel-based geometric measure that transforms conventional distance computations into geometry-aware similarity assessments, enabling effective clustering on non-linear manifolds and varying cluster densities. \textit{Second}, we formulate a federated multi-view clustering algorithm that extends the centralized heat kernel framework to distributed settings, incorporating differential privacy mechanisms and secure aggregation protocols tailored for clustered network architectures. \textit{Third}, we establish theoretical foundations including convergence guarantees, privacy bounds, and communication complexity analysis for the proposed federated clustering framework. \textit{Fourth}, we validate the framework through comprehensive experimental evaluation on synthetic multi-view datasets representative of network-generated data, demonstrating substantial improvements over state-of-the-art baselines in clustering accuracy, communication efficiency, and privacy preservation.

In this work, we propose a novel soft clustering framework that replaces conventional Euclidean distance (CED) with kernelized geometric measures derived from H-KC. These tools are rooted in QFT and spectral analysis. By modeling similarity via heat diffusion over data manifolds, our approach provides a richer, geometry-aware mechanism for aggregating and aligning distributed views. Furthermore, the theoretical underpinnings from QFT offer a natural interpretation of influence propagation, a notion that is increasingly relevant in the context of collaborative AI and agent-based systems.

Our method enables flexible, privacy-preserving learning across multiple views and agents, with theoretical guaranties and practical scalability. It bridges foundational mathematics with contemporary machine learning (ML) architecture, contributing a unified lens to interpret and design decentralized intelligent systems (DIS). In summary, the contributions of this study include:
\begin{enumerate}
    \item In the contemporary era of modular artificial intelligence (AI), where the harmonization of not only data views but also distributed agents across diverse sources is paramount, we propose a unified theoretical framework that integrates maximum-volume (MV) learning, kernel geometry, and quantum field theory (QFT) to facilitate scalable, interpretable federated learning (FL). We introduce the concept of heat-kernel coefficients (H-KC) as a novel mechanism to capture local geometric structures across decentralized views and agents, thereby enabling effective information diffusion in DIS. This framework provides a principled basis for designing algorithms that respect both data heterogeneity and agent autonomy.

    \item The present study proposes a methodology for the analysis of local geometric structures across decentralized views. This methodology is based on heat kernel-based soft clustering, a technique that has been employed in the study of large language models (LLM) and "LLM-agents," which encode semantic geometry in high-dimensional manifolds. The present study proposes an extension to the concept of centralized clustering algorithms, with the objective of establishing a framework to bridge the Heat Kernel-Enhanced Multi-View Fuzzy Clustering (HK-MVFC). The mathematical formulation for federated HK-MVFC, abbreviated as FedHK-MVFC, is implemented. This algorithm facilitates the collaborative clustering of multi-view data by multiple clients (e.g., hospitals) while preserving privacy and minimizing communication overhead. The theoretical analysis provided herein demonstrates convergence properties and robustness to data heterogeneity.

    \item As medical datasets of this nature are not publicly available, we have generated synthetic datasets that mimic the characteristics of real-world multi-view medical data. We contribute a novel synthetic dataset with non-linear relationships across views, designed to evaluate the effectiveness of geometry-aware clustering methods.
    
\end{enumerate}

The remainder of this paper is organized as follows: In Section \ref{sec:literature_review}, an examination of extant literature pertaining to federated learning, multi-view clustering, and heat kernel methods is conducted. The subsequent sections (\ref{sec:HKMVFC}-\ref{sec:Federated_HK_MVFC}) delineate the mathematical formulation of the proposed HK-MVFC and FedHK-MVFC algorithms. As delineated in Section \ref{sec:Privacy_Preserving_Concept}, the proposed Fed-HKMVFC concept is predicated on a privacy-preserving framework. In Section \ref{sec:Experimental_Evaluation}, experimental results are presented on synthetic datasets. These results are compared with baseline methods. Conclusively, Section \ref{sec:conclusion} culminates in a discourse on the ramifications and prospective avenues for further research.

\section{Literature Review}
\label{sec:literature_review}

Federated learning (FL) has emerged as a paradigm for collaborative model training across decentralized data sources while preserving privacy \cite{mcmahan2017communication, konevcny2016federated}. Early works focused on federated averaging (FedAvg) and its variants to address communication efficiency and robustness to data heterogeneity. Recent advances include personalization strategies \cite{smith2017federated, li2021ditto}, secure aggregation \cite{bonawitz2017practical}, and applications in healthcare, finance, and the Internet of Things (IoT). Clustering algorithms, including k-means, hierarchical clustering, and fuzzy c-means (FCM) (see reference \cite{bezdek2013pattern}), are foundational in unsupervised learning. Multi-view clustering extends these methods to leverage complementary information across multiple feature representations (see references \cite{xu2013survey, chen2020multi}). Kernel-based approaches, such as spectral clustering and kernel k-means, address non-linear relationships and manifold structures (see references \cite{ng2001spectral, dhillon2004kernel}). Distributed clustering integrates clustering algorithms with distributed computing frameworks to scale to large datasets and decentralized environments \cite{guha2003clustering, xu2015comprehensive}. In contrast, federated clustering combines FL principles with clustering to enable privacy-preserving unsupervised learning across clients \cite{brisimi2018federated, duan2021fedgroup}. However, existing methods often rely on conventional distance metrics and face challenges in handling heterogeneous data distributions and aligning cluster assignments across clients. Recent works have explored kernel methods and manifold learning in federated settings (see \cite{chen2022federated, liu2023federated}), highlighting the importance of geometry-aware similarity measures. However, most approaches lack theoretical guaranties and practical scalability in complex multi-view scenarios.

In centralized multi-view clustering, Yang and Sinaga \cite{yang2021collaborative} introduced Co-FW-MVFCM (Collaborative Feature-Weighted Multi-View Fuzzy C-Means), which enables collaborative learning across data views through adaptive feature weighting. While Co-FW-MVFCM demonstrates effectiveness in centralized environments, it lacks mechanisms for distributed deployment and cannot handle geometrically complex data structures such as non-convex clusters or manifold-embedded patterns. Similarly, Zhou et al. \cite{zhou2025mvwecm} proposed MvWEC (Multi-view Weighted Evidential C-Means), incorporating evidential reasoning to manage uncertainty in multi-view clustering. Despite its theoretical contributions, MvWEC still relies on traditional distance measures that may not fully exploit the geometric structure of multi-view data, particularly when views exhibit complementary rather than redundant information.

The extension to federated settings introduces additional complications. Yang et al. \cite{yang2024federated} proposed federated k-means clustering based on feature weighting, addressing privacy preservation and communication efficiency. However, this approach inherits the fundamental limitation of k-means—dependence on Euclidean distance—which restricts its capability to capture complex relationships in multi-view data. Liu et al. \cite{liu2023federated} developed federated probabilistic preference learning for privacy-preserving multi-domain recommendation based on co-clustering. While addressing privacy concerns through local computation and secure aggregation, existing co-clustering methods often depend on conventional distance metrics, struggle with data heterogeneity across clients, and lead to inconsistent clustering results.

Recent tensorized approaches have extended single-view clustering to multi-view settings. Liu et al. \cite{liu2024adaptively} introduced adaptively weighted multi-view tensorized clustering, leveraging tensor decomposition for view integration. However, the underlying distance computations remain Euclidean-based, limiting effectiveness on datasets with intrinsic geometric complexity. Furthermore, most tensor-based methods \cite{chen2011tw, xu2016weighted, zhang2018tw, sinaga2024rectified, kaushal2024weighted} do not consider distributed or federated learning settings, restricting their applicability in privacy-sensitive scenarios such as healthcare and finance.

Despite significant advancements in federated learning and multi-view clustering, several fundamental limitations persist when integrating heterogeneous data across distributed environments. The principal challenge stems from the conventional distance metrics employed by existing methods, which fail to capture the intrinsic geometric structures inherent in multi-view data. Traditional approaches predominantly rely on Euclidean distance, which assumes linear relationships and uniform feature scaling—assumptions that are frequently violated in real-world heterogeneous datasets. This metric inadequacy becomes particularly problematic when dealing with non-linear manifolds, varying cluster densities, and complex topological structures common in medical imaging, social networks, and genomic data.

In healthcare applications, these limitations manifest critically. Multi-institutional patient data often exhibit complex geometric patterns across modalities (ECG signals, imaging, genomic profiles), heterogeneous statistical distributions due to demographic variations, and strict privacy constraints preventing raw data sharing. Existing methods fail to simultaneously address these requirements: geometry-aware similarity measures for capturing intrinsic data structures, federated learning protocols for privacy-preserving collaboration, and robust aggregation mechanisms for handling heterogeneous client distributions. 

\subsection{Research Gaps and Motivation}

The comprehensive analysis of existing literature reveals three critical research gaps that motivate the present work:

\textbf{Gap 1: Geometric Structure Ignorance.} Current multi-view clustering methods, both centralized and federated, predominantly employ Euclidean distance metrics that fail to capture the intrinsic manifold geometry of complex data distributions. This limitation is particularly severe when data resides on non-linear manifolds or exhibits varying local densities—scenarios common in medical imaging, genomic analysis, and sensor network data. While kernel methods have been explored in single-view settings, their principled integration into multi-view federated frameworks remains unexplored, especially with theoretical foundations grounded in quantum field theory and spectral graph analysis.

\textbf{Gap 2: Multi-View Heterogeneity in Federated Settings.} Existing federated clustering approaches lack robust mechanisms for handling heterogeneous multi-view data across distributed clients. Current methods either aggregate views naively without considering their complementary nature or employ view-specific models that fail to leverage cross-view consistency. The challenge intensifies when different clients possess varying numbers of views with distinct statistical properties, a scenario frequently encountered in multi-institutional healthcare collaborations where data collection protocols differ across sites.

\textbf{Gap 3: Privacy-Performance Trade-off.} While differential privacy and secure aggregation techniques exist for federated learning, their integration with geometry-aware multi-view clustering introduces unique challenges. Existing privacy-preserving methods either sacrifice clustering accuracy significantly or require prohibitive communication overhead. Furthermore, theoretical analysis of privacy-utility trade-offs in the context of heat kernel-enhanced federated clustering remains absent from the literature.

\subsection{Research Objectives and Contributions}

Motivated by these gaps, this work pursues the following research objectives:

\textbf{Objective 1:} Develop a theoretically grounded framework that integrates heat kernel coefficients—derived from quantum field theory and spectral graph analysis—into multi-view fuzzy clustering, enabling geometry-aware similarity assessment that captures intrinsic manifold structures.

\textbf{Objective 2:} Extend the centralized heat kernel-enhanced framework to federated settings, incorporating differential privacy mechanisms, secure aggregation protocols, and adaptive view weighting strategies tailored for heterogeneous multi-view data across distributed clients.

\textbf{Objective 3:} Establish rigorous mathematical foundations including convergence guarantees for both centralized and federated algorithms, privacy-utility analysis under differential privacy constraints, and communication complexity characterization for federated implementations.

\textbf{Objective 4:} Validate the proposed framework through comprehensive experimental evaluation on synthetic multi-view datasets with controlled geometric complexity, demonstrating substantial improvements over state-of-the-art baselines in clustering accuracy, communication efficiency, and privacy preservation.

Our proposed FedHK-MVFC framework addresses these objectives by integrating heat kernel-based geometric measures with federated multi-view clustering, enabling effective collaboration while preserving data privacy and capturing complex geometric relationships. The framework provides a unified theoretical foundation that bridges quantum field theory concepts with practical federated learning implementations, offering both algorithmic innovations and rigorous mathematical analysis that advance the state-of-the-art in privacy-preserving distributed multi-view clustering.

\section{Heat-Kernel Enhanced Multi-view Clustering}
\label{sec:HKMVFC}

This section presents our proposed heat-kernel enhanced multi-view clustering framework. We first establish the mathematical foundations and notation used throughout our formulations, then develop our two novel clustering algorithms: Heat Kernel-Enhanced Multi-View Fuzzy Clustering (HK-MVFC) and its extension, Federated Heat Kernel Multi-View Fuzzy Clustering (FedHK-MVFC). HK-MVFC algorithm is designed to effectively integrate information across heterogeneous data views in (non-distributed) mechanisms while leveraging heat-kernel coefficients to enhance representational capacity. In particular, we aim to address the following research problems, including how we can effectively integrate multiple heterogeneous data views into a unified clustering framework that captures complex relationships among data points across views and how we can leverage heat-kernel coefficients to enhance the representational capacity of clustering algorithms, particularly in capturing the intrinsic geometries of data manifolds. 

\subsection{Mathematical Preliminaries and Problem Formulation}

We begin by establishing the mathematical framework and notation that will be used throughout our formulation. This formulation serves as the foundation for both the centralized HK-MVFC and federated FedHK-MVFC algorithms.

\subsubsection{Multi-View Data Representation}

Consider a multi-view dataset $X \in \mathbb{R}^{n \times D}$ consisting of $n$ data samples observed across $s$ distinct views. Each view $h \in \{1, 2, \ldots, s\}$ captures a different perspective or modality of the underlying data, represented as:

\begin{equation}
X^h = \{x_1^h, x_2^h, \ldots, x_n^h\} \quad \text{for } h = 1, 2, \ldots, s
\label{eqn:view_definition}
\end{equation}

where each data point $x_i^h \in \mathbb{R}^{d_h}$ in view $h$ is characterized by $d_h$ features:

\begin{equation}
x_i^h = [x_{i1}^h, x_{i2}^h, \ldots, x_{id_h}^h]^T \quad \text{for } i = 1, 2, \ldots, n
\label{eqn:data_point}
\end{equation}

The total dimensionality of the multi-view dataset is given by $D = \sum_{h=1}^s d_h$, representing the aggregated feature space across all views.

\subsubsection{Clustering Parameters}

The fundamental objective is to partition the $n$ data samples into $c$ clusters, where the clustering structure is consistent across all views. This clustering is characterized by several key mathematical objects:

\textbf{Unified Membership Matrix:} The clustering assignment is represented by a global fuzzy membership matrix $U^* \in \mathbb{R}^{n \times c}$, where each element $\mu_{ik}^*$ denotes the degree of membership of data point $i$ to cluster $k$:

\begin{equation}
U^* = [\mu_{ik}^*]_{n \times c}, \quad \text{where } \mu_{ik}^* \in [0,1]
\label{eqn:membership_matrix}
\end{equation}

The membership values satisfy the probabilistic constraint that ensures each data point's total membership across all clusters sums to unity:

\begin{equation}
\sum_{k=1}^c \mu_{ik}^* = 1 \quad \text{for all } i = 1, 2, \ldots, n
\label{eqn:membership_constraint}
\end{equation}

\textbf{View Weight Vector:} To account for the varying importance and informativeness of different views, we introduce an adaptive view weight vector $V \in \mathbb{R}^{1 \times s}$:

\begin{equation}
V = [v_1, v_2, \ldots, v_s], \quad \text{where } v_h \in [0,1]
\label{eqn:view_weights}
\end{equation}

These weights are normalized to ensure a valid probability distribution over views:

\begin{equation}
\sum_{h=1}^s v_h = 1
\label{eqn:view_weight_constraint}
\end{equation}

Higher values of $v_h$ indicate that view $h$ is more informative for the clustering task, enabling the algorithm to automatically emphasize more reliable data sources.

\textbf{Cluster Center Matrix:} For each view $h$, the cluster prototypes are represented by a cluster center matrix $A^h \in \mathbb{R}^{c \times d_h}$:

\begin{equation}
A^h = [a_{kj}^h]_{c \times d_h} \quad \text{for } h = 1, 2, \ldots, s
\label{eqn:cluster_centers}
\end{equation}

where $a_{kj}^h$ represents the $j$-th feature of the $k$-th cluster center in view $h$. The complete set of cluster centers across all views is denoted as $\mathcal{A} = \{A^1, A^2, \ldots, A^s\}$.

\subsubsection{Problem Statement}

Given the multi-view dataset $X = \{X^1, X^2, \ldots, X^s\}$ and a specified number of clusters $c$, our objective is to determine the optimal clustering parameters $(U^*, V, \mathcal{A})$ that minimize a unified objective function. This objective function will integrate information across all views while accounting for the intrinsic geometric structure of the data through heat kernel-enhanced distance measures.

The mathematical framework established here provides the foundation for developing both centralized and federated multi-view clustering algorithms that can effectively handle heterogeneous data sources while preserving data privacy in distributed settings.

\subsection{Heat-Kernel Coefficients}

Central to our approach is the formulation of heat-kernel coefficients (H-KC), which enables the transformation of Euclidean distance in the original feature space into an exponential kernel distance. For the $j$-th feature in the $h$-th view, we define H-KC $\delta _{ij}^h$ using two alternative estimators:

\begin{equation}
    \delta _{ij}^h = \frac{{x_{ij}^h - \mathop {\min }\limits_{1 \le i \le n} \left( {x_{ij}^h} \right)}}{{\mathop {\max }\limits_{1 \le i \le n} \left( {x_{ij}^h} \right) - \mathop {\min }\limits_{1 \le i \le n} \left( {x_{ij}^h} \right)}}
    \label{HKCv1}
\end{equation}

\textbf{Note:} To ensure numerical stability, a small constant $\epsilon > 0$ is added to the denominator. If all feature values are identical, $\max = \min$, and the denominator becomes $\epsilon$ to avoid division by zero.

and

\begin{equation}
    \delta _{ij}^h = \left| {x_{ij}^h - {{\bar x}^h}} \right|
    \label{HKCv2}
\end{equation}

where $\bar{x}^h = {1 \mathord{\left/
 {\vphantom {1 n}} \right.
 \kern-\nulldelimiterspace} n}\sum\nolimits_{i = 1}^n {x_{ij}^h}$ represents the mean value of the $j$-th feature in view $h$.
 
\subsection{Conventional Fuzzy C-Means in Multi-View Context}

The standard fuzzy c-means (FCM) algorithm minimizes the following objective function:

\begin{equation}
    {J_{FCM}}\left( {U,V} \right) = \sum\limits_{i = 1}^n {\sum\limits_{k = 1}^c {\mu _{ik}^m} } \sum\limits_{j = 1}^d {{{\left( {{x_{ij}} - {a_{kj}}} \right)}^2}} 
    \label{eqn:CFCM}
\end{equation}

where $m>1$ is the fuzzifier parameter controlling the degree of fuzziness in the resulting partitions.

In a multi-view setting, conventional FCM faces limitations when attempting to derive unified patterns from heterogeneous data sources. These limitations stem from its reliance on the Euclidean distance metric, which fails to capture non-linear relationships and cannot effectively balance contributions from views of varying dimensionality, scale, and noise characteristics.

\subsection{Heat Kernel-Enhanced Multi-View Fuzzy Clustering}

To address the challenges of clustering heterogeneous multi-view data, we propose the Heat Kernel-Enhanced Multi-View Fuzzy Clustering (HK-MVFC) algorithm. This approach integrates H-KC into a fuzzy clustering framework, enabling more effective capture of complex relationships between data points and cluster centers across multiple views.

The foundation of our approach is the transformation of CED metrics into exponential kernel distances. To achieve this, we introduce a novel Kernel Euclidean Distance (KED) formulation:

\begin{equation}
    {\rm{KE}}{{\rm{D}}_1}\left( {x_{ij}^h,a_{kj}^h} \right) = \exp \left( { - \sum\limits_{j = 1}^{{d_h}} {\delta _{ij}^h{{\left( {x_{ij}^h - a_{kj}^h} \right)}^2}} } \right)
    \label{eqn:KEDv1}
\end{equation}

In Eq. \ref{eqn:KEDv1}, the H-KC acts as adaptive weighting factors, enabling the algorithm to effectively balance between local and global feature importance. This formulation maps distances to the range $\left[ {0,1} \right]$, with values approaching 1 indicating high similarity between data points and cluster centers.

For clustering purposes, we require a dissimilarity measure that approaches zero for identical points. Therefore, we normalize ${\rm{KE}}{{\rm{D}}_1}\left( {x_{ij}^h,a_{kj}^h} \right)$ to obtain:

\begin{equation}
    {\rm{KE}}{{\rm{D}}_2}\left( {x_{ij}^h,a_{kj}^h} \right) = 1 - \exp \left( { - \sum\limits_{j = 1}^{{d_h}} {\delta _{ij}^h{{\left( {x_{ij}^h - a_{kj}^h} \right)}^2}} } \right)
    \label{eqn:KEDv2}
\end{equation}

We denote ${\rm{KE}}{{\rm{D}}_2}\left( {x_{ij}^h,a_{kj}^h} \right)$ as $d_{\exp \left( {ik,j} \right)}^h$ for notational simplicity in subsequent formulations.

Building on the kernel distance transformation, we formulate the HK-MVFC objective function as follows:

\begin{equation}
    {J_{HK-MVFC}}\left( {V,U^*,A} \right) = \sum\limits_{h = 1}^s {v_h^\alpha } \sum\limits_{i = 1}^n {\sum\limits_{k = 1}^c {{{\left( {\mu _{ik}^*} \right)}^m}} } d_{\exp \left( {ik,j} \right)}^h
    \label{eqn:HKMVFC}
\end{equation}

Subject to:

\begin{equation*}
    \sum\limits_{k = 1}^c {\mu _{ik}^* = 1} ,{\rm{ }}\mu _{ik}^* \in \left[ {0,1} \right]
    \label{eqn:constraint1}
\end{equation*}
\begin{equation*}
    \sum\limits_{h = 1}^s {{v_h} = 1,{\rm{ }}{v_h} \in \left[ {0,1} \right]} 
    \label{eqn:constraint2}
\end{equation*}

The exponent parameter $\alpha$ of our proposed HK-MVFC in Eq. \ref{eqn:HKMVFC} controls the influence of view weights, with larger values increasing the contrast between different views' contributions. Here, we set $\alpha>1$, a user-defined parameter that adjusts the real-world problem needs. And the fuzzifier $m>1$  determines the fuzziness of the resulting partition.

A critical aspect of our approach is the use of a unified membership matrix $U^* = \left[ {\mu _{ik}^*} \right]$ across all views. Unlike approaches that compute separate memberships for each view and subsequently merge them, HK-MVFC enforces consistency across views from the outset, resulting in a more coherent clustering structure.

\subsection{Optimization Framework: HK-MVFC}

The objective function of proposed HK-MVFC in Eq. \ref{eqn:HKMVFC} cannot be minimized directly due to its non-convex nature. We therefore employ an alternating optimization approach, iteratively updating the membership matrix $U^*$, the view weights $V$, and the cluster centers $A$.

To derive the update rules, we construct the Lagrangian of Eq. \ref{eqn:HKMVFC} in the following way

\begin{equation}
    {{\tilde J}_{HK-MVFC}} = {J_{HK-MVFC}}\left( {V,U^*A} \right) + \sum\limits_{i = 1}^n {{\lambda _{1i}}} \left( {\sum\limits_{k = 1}^c {\mu_{ik}^* - 1} } \right) + {\lambda _2}\left( {\sum\limits_{h = 1}^s {{v_h} - 1} } \right)
    \label{eqn:Lag_HKMVFC}
\end{equation}

where $\lambda _{1i}$ and $\lambda _2$ are Lagrange multipliers enforcing the normalization constraints.

\begin{theorem}[HK-MVFC Update Rules]
\label{thm:HK_MVFC}

The necessary conditions for minimizing the objective function $J_{HK - MVFC}$ in Eq. \ref{eqn:HKMVFC} yield the following update rules for the membership matrix $\mu^*$, cluster centers $a$, and view weights $v$:

\begin{equation}
    \mu_{ik}^* = \frac{{{{\left( {\sum\limits_{h = 1}^s {v_h^\alpha d_{\exp \left( {ik,j} \right)}^h} } \right)}^{ - {{\left( {m - 1} \right)}^{ - 1}}}}}}{{\sum\limits_{k' = 1}^c {{{\left( {\sum\limits_{h = 1}^s {v_h^\alpha d_{\exp \left( {ik',j} \right)}^h} } \right)}^{ - {{\left( {m - 1} \right)}^{ - 1}}}}} }}
    \label{eqn:UpdateU_HKMVFC}
\end{equation}

\noindent\textbf{Membership Update:} Eq. \ref{eqn:UpdateU_HKMVFC} updates the membership degree $\mu_{ik}^*$ of data point $x_i$ to cluster $k$, where the weighted distances across all $s$ views are aggregated using view weights $v_h^\alpha$, and the fuzzifier parameter $m$ controls the degree of membership overlap.

\begin{equation}
    a_{kj}^h = \frac{{\sum\limits_{i = 1}^n {{{\left( {\mu _{ik}^*} \right)}^m}v_h^\alpha \exp \left( { - \delta _{ij}^h{{\left\| {x_i^h - a_k^h} \right\|}^2}} \right)} }}{{\sum\limits_{i = 1}^n {{{\left( {\mu _{ik}^*} \right)}^m}v_h^\alpha \exp \left( { - \delta _{ij}^h{{\left\| {x_i^h - a_k^h} \right\|}^2}} \right)} }}x_{ij}^h
    \label{eqn:UpdateA_HKMVFC}
\end{equation}

\noindent\textbf{Cluster Center Update:} Eq. \ref{eqn:UpdateA_HKMVFC} computes the $j$-th feature of cluster center $a_k$ in view $h$ as a weighted average of data points, where weights combine membership degrees $(\mu_{ik}^*)^m$, view importance $v_h^\alpha$, and exponential kernel terms $\exp(-\delta_{ij}^h\|x_i^h - a_k^h\|^2)$.

\begin{equation}
    {v_h} = \frac{{{{\left( {\sum\limits_{i = 1}^n {\sum\limits_{k = 1}^c {{{\left( {\mu _{ik}^*} \right)}^m}d_{\exp \left( {ik,j} \right)}^h} } } \right)}^{ - {{\left( {\alpha  - 1} \right)}^{ - 1}}}}}}{{\sum\limits_{h' = 1}^s {{{\left( {\sum\limits_{i = 1}^n {\sum\limits_{k = 1}^c {{{\left( {\mu _{ik}^*} \right)}^m}d_{\exp \left( {ik,j} \right)}^{h'}} } } \right)}^{ - {{\left( {\alpha  - 1} \right)}^{ - 1}}}}} }}
    \label{eqn:UpdateV_HKMVFC}
\end{equation}

\noindent\textbf{View Weight Update:} Eq. \ref{eqn:UpdateV_HKMVFC} determines the importance weight $v_h$ for view $h$, inversely proportional to the total weighted distance in that view, where $\alpha$ controls the distribution of weights across views (normalized such that $\sum_{h=1}^s v_h = 1$).

\begin{remark}
These update rules form an iterative alternating optimization scheme: given current estimates of cluster centers and view weights, Eq. \ref{eqn:UpdateU_HKMVFC} updates memberships; given memberships and view weights, Eq. \ref{eqn:UpdateA_HKMVFC} updates centers; and given memberships and centers, Eq. \ref{eqn:UpdateV_HKMVFC} updates view weights. This process continues until convergence.
\end{remark}

\end{theorem}

\begin{proof}
We establish the necessary optimality conditions for the HK-MVFC objective function using Lagrangian optimization and derive the closed-form update rules for each parameter set.

\textbf{Part I: Membership Matrix Update Rule Derivation}

To derive the update equation for the membership coefficients $\mu_{ik}^*$, we begin by computing the partial derivative of the Lagrangian in Eq. \ref{eqn:Lag_HKMVFC} with respect to $\mu_{ik}^*$. This derivative represents the marginal change in the objective function when we adjust the membership of data point $i$ to cluster $k$.

\begin{align}
    \frac{\partial \tilde{J}_{HK-MVFC}}{\partial \mu_{ik}^*} &= m(\mu_{ik}^*)^{m-1} \sum_{h=1}^s v_h^\alpha d_{\exp(ik,j)}^h + \lambda_{1i} = 0 \label{eqn:membership_derivative}
\end{align}

where the first term $m(\mu_{ik}^*)^{m-1}$ arises from the power rule applied to the fuzzified membership $(\mu_{ik}^*)^m$, and the summation $\sum_{h=1}^s v_h^\alpha d_{\exp(ik,j)}^h$ represents the weighted contribution of all views to the distance between data point $i$ and cluster center $k$.

Solving Eq. \ref{eqn:membership_derivative} for $\mu_{ik}^*$ yields:
\begin{align}
    \mu_{ik}^* &= \left(-\frac{\lambda_{1i}}{m}\right)^{(m-1)^{-1}} \left(\sum_{h=1}^s v_h^\alpha d_{\exp(ik,j)}^h\right)^{-(m-1)^{-1}} \label{eqn:diff_U_HKMVFC_v1}
\end{align}

Here, the exponent $(m-1)^{-1} = \frac{1}{m-1}$ transforms the equation from its derivative form back to the original membership variable. The negative sign in the Lagrange multiplier term ensures that memberships decrease as distances increase, which is the desired clustering behavior.

To determine the Lagrange multiplier $\lambda_{1i}$, we impose the normalization constraint $\sum_{k=1}^c \mu_{ik}^* = 1$, which ensures that the total membership of data point $i$ across all clusters sums to unity (a fundamental requirement in fuzzy clustering):

\begin{align}
    \sum_{k=1}^c \left(-\frac{\lambda_{1i}}{m}\right)^{(m-1)^{-1}} \left(\sum_{h=1}^s v_h^\alpha d_{\exp(ik,j)}^h\right)^{-(m-1)^{-1}} &= 1 \\
    \Rightarrow \left(-\frac{\lambda_{1i}}{m}\right)^{(m-1)^{-1}} &= \frac{1}{\sum_{k'=1}^c \left(\sum_{h=1}^s v_h^\alpha d_{\exp(ik',j)}^h\right)^{-(m-1)^{-1}}} \label{eqn:diff_U_HKMVFC_v2}
\end{align}

The denominator in Eq. \ref{eqn:diff_U_HKMVFC_v2} serves as a normalization factor that ensures all memberships sum to one. Substituting this expression back into Eq. \ref{eqn:diff_U_HKMVFC_v1} yields the final membership update rule in Eq. \ref{eqn:UpdateU_HKMVFC}, where each membership is inversely proportional to the weighted distance raised to the power $\frac{1}{m-1}$.

\textbf{Part II: Cluster Centers Update Rule Derivation}

To find the optimal cluster centers $a_{kj}^h$, we differentiate the HK-MVFC objective function in Eq. \ref{eqn:HKMVFC} with respect to $a_{kj}^h$. This derivative measures how the objective function changes when we adjust the $j$-th feature of cluster center $k$ in view $h$:

\begin{align}
    \frac{\partial J_{HK-MVFC}}{\partial a_{kj}^h} &= v_h^\alpha \sum_{i=1}^n (\mu_{ik}^*)^m \frac{\partial d_{\exp(ik,j)}^h}{\partial a_{kj}^h}
\end{align}

where $v_h^\alpha$ weights the contribution of view $h$, and $(\mu_{ik}^*)^m$ weights each data point's contribution by its membership to cluster $k$.

Expanding the kernel distance derivative:
\begin{align}
    \frac{\partial d_{\exp(ik,j)}^h}{\partial a_{kj}^h} &= \frac{\partial}{\partial a_{kj}^h} \left\{1 - \exp\left(-\sum_{j=1}^{d_h} \delta_{ij}^h (x_{ij}^h - a_{kj}^h)^2\right)\right\} \\
    &= -\frac{\partial}{\partial a_{kj}^h} \exp\left(-\sum_{j=1}^{d_h} \delta_{ij}^h (x_{ij}^h - a_{kj}^h)^2\right)
\end{align}

The exponential term captures the heat kernel transformation, where $\delta_{ij}^h$ are the heat-kernel coefficients that adapt to local data geometry. Applying the chain rule:

\begin{align}
    \frac{\partial J_{HK-MVFC}}{\partial a_{kj}^h} &= 2\delta_{ij}^h v_h^\alpha \sum_{i=1}^n (\mu_{ik}^*)^m \exp\left(-\sum_{j=1}^{d_h} \delta_{ij}^h (x_{ij}^h - a_{kj}^h)^2\right) (x_{ij}^h - a_{kj}^h) \label{eqn:diff_A_HKMVFC}
\end{align}

The factor of 2 arises from differentiating the squared term $(x_{ij}^h - a_{kj}^h)^2$, while the exponential term weights contributions based on the heat kernel similarity between data points and cluster centers. The term $(x_{ij}^h - a_{kj}^h)$ indicates the direction of adjustment: positive when data points are larger than the center, negative otherwise.

Setting Eq. \ref{eqn:diff_A_HKMVFC} to zero and rearranging yields Eq. \ref{eqn:UpdateA_HKMVFC}, which shows that cluster centers are weighted averages of data points, where weights combine fuzzy memberships $(\mu_{ik}^*)^m$, view importance $v_h^\alpha$, and heat kernel similarities $\exp(-\delta_{ij}^h \|x_i^h - a_k^h\|^2)$.

\textbf{Part III: View Weights Update Rule Derivation}

To update the view weights $v_h$, we compute the partial derivative of the Lagrangian in Eq. \ref{eqn:Lag_HKMVFC} with respect to $v_h$. This derivative quantifies how the objective function changes when we adjust the importance assigned to view $h$:

\begin{align}
    \frac{\partial \tilde{J}_{HK-MVFC}}{\partial v_h} &= \alpha v_h^{\alpha-1} \sum_{i=1}^n \sum_{k=1}^c (\mu_{ik}^*)^m d_{\exp(ik,j)}^h + \lambda_2 = 0
\end{align}

The term $\alpha v_h^{\alpha-1}$ comes from differentiating $v_h^\alpha$ with respect to $v_h$, while the double summation $\sum_{i=1}^n \sum_{k=1}^c (\mu_{ik}^*)^m d_{\exp(ik,j)}^h$ represents the total weighted distance contribution of view $h$ across all data points and clusters.

Solving for $v_h$:
\begin{align}
    v_h &= \left(-\frac{\lambda_2}{\alpha}\right)^{(\alpha-1)^{-1}} \left(\sum_{i=1}^n \sum_{k=1}^c (\mu_{ik}^*)^m d_{\exp(ik,j)}^h\right)^{-(\alpha-1)^{-1}} \label{eqn:diff_V_HKMVFC}
\end{align}

The exponent $(\alpha-1)^{-1} = \frac{1}{\alpha-1}$ transforms the derivative relationship back to the view weight variable. The negative exponent on the distance summation ensures that views with smaller total distances (better clustering quality) receive higher weights.

Applying the normalization constraint $\sum_{h=1}^s v_h = 1$ determines the Lagrange multiplier $\lambda_2$ and yields the final view weight update in Eq. \ref{eqn:UpdateV_HKMVFC}. This adaptive weighting mechanism automatically assigns higher importance to more informative views, where informativeness is measured by the view's ability to create compact, well-separated clusters.

The alternating optimization of these three parameter sets (memberships, cluster centers, and view weights) constitutes the complete HK-MVFC algorithm, with each update rule derived from necessary optimality conditions of the constrained optimization problem.
\end{proof}
\subsection{Theoretical Insights}

Several important theoretical properties of the HK-MVFC algorithm warrant emphasis. The algorithm exhibits a consistency property where reducing intra-cluster distances within a view while increasing inter-cluster distances across different views preserves the clustering outcome. This property ensures stable performance even with heterogeneous data distributions, which is particularly crucial in multi-view scenarios where individual views may exhibit varying degrees of noise and signal quality.

The update rule for view weights in Eq. \ref{eqn:UpdateV_HKMVFC} demonstrates an adaptive weighting mechanism that assigns higher weights to views with smaller intra-cluster distances, effectively prioritizing more informative views during the clustering process. The parameter $\alpha$ controls the degree of contrast between view weights, with larger values increasing the differentiation between high-quality and low-quality views. This adaptive behavior enables the algorithm to automatically adjust to the relative informativeness of different data modalities without requiring manual tuning of view contributions.

The heat-kernel coefficients $\delta _{ij}^h$ provide a sophisticated mechanism for adapting to the local geometry of the data in each view. Unlike conventional Euclidean distance metrics that assume uniform data distributions, these coefficients capture the intrinsic manifold structure of the data, enabling the algorithm to identify complex cluster structures that may be obscured when using standard distance measures. This geometric sensitivity is particularly valuable when dealing with non-linear cluster boundaries and varying cluster densities across different views.

\subsection{HK-MVFC Algorithm}

Algorithm \ref{alg:HK_MVFC} provides a detailed implementation of the HK-MVFC in Eq. \ref{eqn:HKMVFC} approach. The algorithm operates by alternately updating the membership matrix, cluster centers, and view weights until convergence.

\begin{algorithm}[!htbp]
\caption{Heat Kernel-Enhanced Multi-View Fuzzy Clustering (HK-MVFC)}
\label{alg:HK_MVFC}
\begin{algorithmic}[1]
\REQUIRE Multi-view dataset $X = \{X^h\}_{h=1}^s$ with $X^h \in \mathbb{R}^{n \times d_h}$
\vspace{0.1cm}
\REQUIRE Number of clusters $c$, fuzzifier $m > 1$, view weight exponent $\alpha > 1$
\vspace{0.1cm}
\REQUIRE Convergence threshold $\varepsilon > 0$, maximum iterations $T_{max}$
\vspace{0.1cm}
\ENSURE Membership matrix $U^* \in [0,1]_{n \times c}$, cluster centers $\{A^h\}_{h=1}^{s}$, view weights $V \in [0,1]_{1 \times s}$
\vspace{0.1cm}
\STATE \textbf{Initialization:}
\vspace{0.18cm}
\STATE $A^{h^{(0)}} \leftarrow$ Initialize cluster centers using k-means++ or random initialization
\vspace{0.1cm}
\STATE $V^{(0)} \leftarrow [1/s, 1/s, \ldots, 1/s]^T$ \COMMENT{Equal view weights}
\vspace{0.1cm}
\STATE $t \leftarrow 0$
\vspace{0.1cm}
\REPEAT
\vspace{0.1cm}
    \STATE $t \leftarrow t + 1$
    \vspace{0.15cm}
    \STATE \textbf{// Heat-Kernel Coefficient Computation}
    \vspace{0.1cm}
    \FOR{$h = 1$ to $s$}
    \vspace{0.1cm}
        \FOR{$i = 1$ to $n$, $j = 1$ to $d_h$}
        \vspace{0.1cm}
            \STATE Compute $\delta_{ij}^{h^{(t)}}$ using Eq. \ref{HKCv1} or \ref{HKCv2}
            \vspace{0.1cm}
        \ENDFOR
        \vspace{0.1cm}
    \ENDFOR
    \vspace{0.15cm}
    \STATE \textbf{// Membership Matrix Update}
    \vspace{0.1cm}
    \FOR{$i = 1$ to $n$, $k = 1$ to $c$}
    \vspace{0.1cm}
        \STATE Update $\mu_{ik}^{*^{(t)}}$ using Eq. \ref{eqn:UpdateU_HKMVFC}
        \vspace{0.1cm}
    \ENDFOR
    \vspace{0.15cm}
    \STATE \textbf{// Cluster Centers Update}
    \vspace{0.1cm}
    \FOR{$h = 1$ to $s$, $k = 1$ to $c$, $j = 1$ to $d_h$}
    \vspace{0.1cm}
        \STATE Update $a_{kj}^{h^{(t)}}$ using Eq. \ref{eqn:UpdateA_HKMVFC}
        \vspace{0.1cm}
    \ENDFOR
    \vspace{0.15cm}
    \STATE \textbf{// View Weights Update}
    \vspace{0.1cm}
    \FOR{$h = 1$ to $s$}
    \vspace{0.1cm}
        \STATE Update $v_h^{(t)}$ using Eq. \ref{eqn:UpdateV_HKMVFC}
        \vspace{0.1cm}
    \ENDFOR
    \vspace{0.15cm}
    \STATE \textbf{// Convergence Check}
    \vspace{0.15cm}
    \STATE Compute the objective function $J_{HK-MVFC}^{(t)}$ using Eq. \ref{eqn:HKMVFC}
    \vspace{0.1cm}
    \STATE $\Delta J \leftarrow |J_{HK-MVFC}^{(t)} - J_{HK-MVFC}^{(t-1)}|$
    \vspace{0.15cm}
\UNTIL{$\Delta J < \varepsilon$ \textbf{or} $t \geq T_{max}$}
\vspace{0.15cm}
\RETURN $U^{*^{(t)}}, \quad \{A^{h^{(t)}}\}_{h=1}^s, \quad V^{(t)}$
\end{algorithmic}
\end{algorithm}
\subsubsection{HK-MVFC Flowchart}

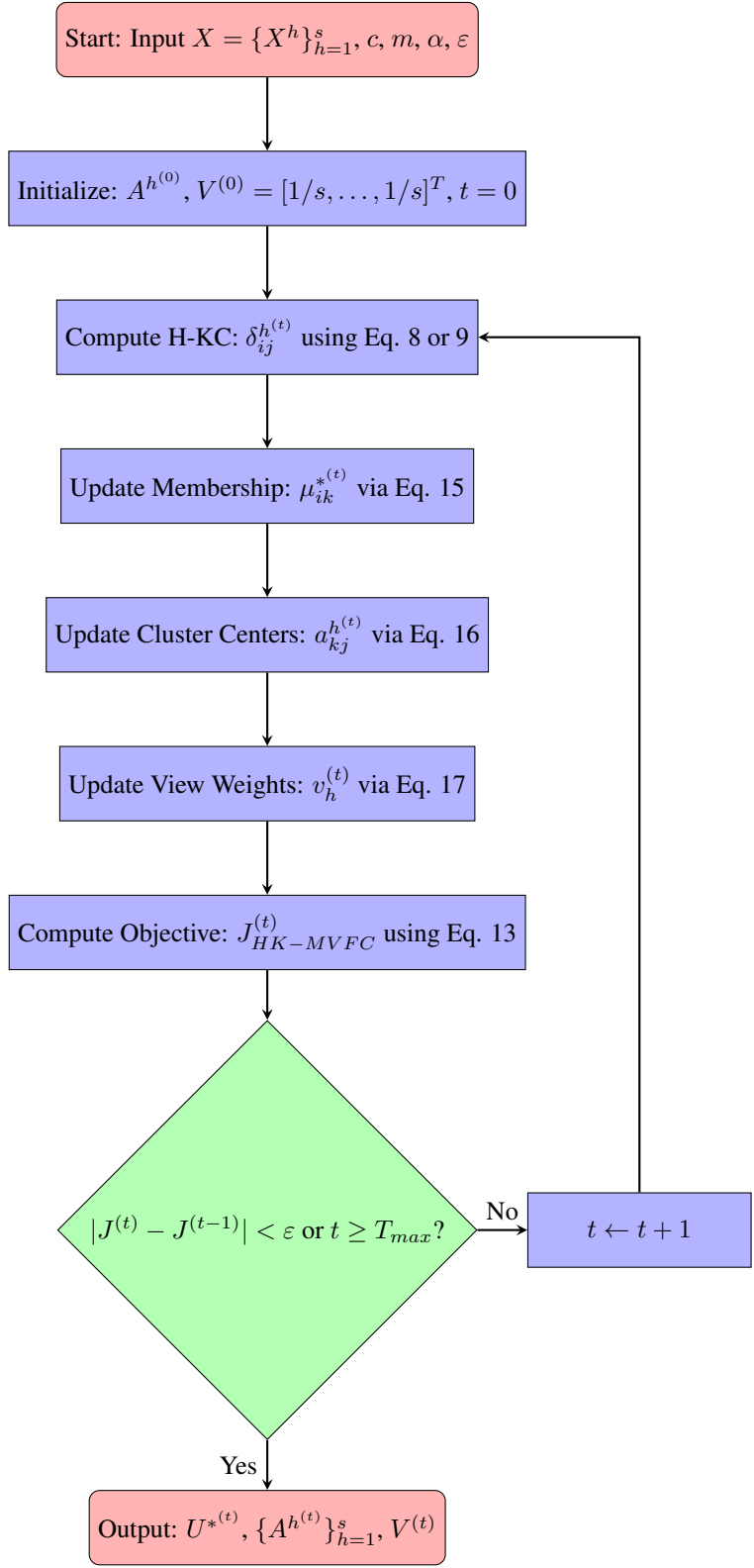
\begin{figure}[h!]
\centering
\tikzstyle{startstop} = [rectangle, rounded corners, minimum width=3cm, minimum height=1cm, text centered, draw=black, fill=red!30]
\tikzstyle{process} = [rectangle, minimum width=3cm, minimum height=1cm, text centered, draw=black, fill=blue!30]
\tikzstyle{decision} = [diamond, minimum width=3cm, minimum height=1cm, text centered, draw=black, fill=green!30]
\tikzstyle{arrow} = [thick,->,>=stealth]

\begin{tikzpicture}[node distance=2cm]
\node (start) [startstop] {Start: Input $X = \{X^h\}_{h=1}^s$, $c$, $m$, $\alpha$, $\varepsilon$};
\node (init) [process, below of=start] {Initialize: $A^{h^{(0)}}$, $V^{(0)} = [1/s, \ldots, 1/s]^T$, $t=0$};
\node (hkc) [process, below of=init] {Compute H-KC: $\delta_{ij}^{h^{(t)}}$ using Eq. \ref{HKCv1} or \ref{HKCv2}};
\node (updateU) [process, below of=hkc] {Update Membership: $\mu_{ik}^{*^{(t)}}$ via Eq. \ref{eqn:UpdateU_HKMVFC}};
\node (updateA) [process, below of=updateU] {Update Cluster Centers: $a_{kj}^{h^{(t)}}$ via Eq. \ref{eqn:UpdateA_HKMVFC}};
\node (updateV) [process, below of=updateA] {Update View Weights: $v_h^{(t)}$ via Eq. \ref{eqn:UpdateV_HKMVFC}};
\node (objective) [process, below of=updateV] {Compute Objective: $J_{HK-MVFC}^{(t)}$ using Eq. \ref{eqn:HKMVFC}};
\node (converge) [decision, below of=objective, yshift=-2cm] {$|J^{(t)} - J^{(t-1)}| < \varepsilon$ or $t \geq T_{max}$?};
\node (increment) [process, right of=converge, xshift=3cm] {$t \leftarrow t+1$};
\node (output) [startstop, below of=converge, yshift=-2cm] {Output: $U^{*^{(t)}}$, $\{A^{h^{(t)}}\}_{h=1}^s$, $V^{(t)}$};

\draw [arrow] (start) -- (init);
\draw [arrow] (init) -- (hkc);
\draw [arrow] (hkc) -- (updateU);
\draw [arrow] (updateU) -- (updateA);
\draw [arrow] (updateA) -- (updateV);
\draw [arrow] (updateV) -- (objective);
\draw [arrow] (objective) -- (converge);
\draw [arrow] (converge) -- node[anchor=east] {Yes} (output);
\draw [arrow] (converge) -- node[anchor=south] {No} (increment);
\draw [arrow] (increment) |- (hkc);
\end{tikzpicture}
\caption{Flowchart of the HK-MVFC Algorithm (Algorithm \ref{alg:HK_MVFC}). The algorithm iteratively updates membership matrix, cluster centers, and view weights using heat kernel-enhanced distances until convergence.}
\label{fig:hk_mvfc_flowchart}
\end{figure}

Figure \ref{fig:hk_mvfc_flowchart} presents the complete workflow of the centralized Heat Kernel-Enhanced Multi-View Fuzzy Clustering algorithm, detailing the initialization, iterative optimization, and convergence checking phases.

\subsection{Computational Complexity of HK-MVFC}

The computational complexity of the HK-MVFC algorithm is determined by several key parameters: the number of data points $n$, the dimensionality of view $h$ denoted as $d_h$, the total number of views $s$, and the number of clusters $c$. A detailed analysis of each computational component reveals the following complexity characteristics.

The computation of heat-kernel coefficients $\delta_{ij}^h$ using either Eq. \ref{HKCv1} or \ref{HKCv2} represents the initial computational overhead. For each view, this calculation requires $O(nd_h)$ operations across all $n$ data points. When extended across all $s$ views, the total complexity for heat-kernel coefficient computation becomes $O(nsd_h)$ per iteration.

The membership matrix update procedure, as defined in Eq. \ref{eqn:UpdateU_HKMVFC}, constitutes the most computationally intensive component of the algorithm. This step requires computing exponential kernel distances between each data point and all cluster centers across all views, resulting in $O(ncd_hs)$ operations. The subsequent normalization step, which ensures that membership values sum to unity for each data point, introduces an additional $O(nc^2s)$ complexity term.

Cluster center updates via Eq. \ref{eqn:UpdateA_HKMVFC} involve weighted averaging operations over all data points, where the weights are determined by membership values and exponential terms. The computation of exponential terms requires $O(ncd_hs)$ operations, while the weighted averaging step contributes $O(ncd_h)$ complexity. The view weight update mechanism, as specified in Eq. \ref{eqn:UpdateV_HKMVFC}, aggregates clustering costs across all data points and clusters, yielding a computational complexity of $O(ncs)$.

Combining all computational components, the overall per-iteration complexity is expressed as $O(nsd_h + ncd_hs + nc^2s + ncd_h + ncs) = O(ncd_hs + nc^2s)$. For typical scenarios where the dimensionality significantly exceeds the number of clusters $(d_h \gg c)$ and assuming balanced view dimensionalities across all views, this complexity simplifies to $O(ncd_{max}s)$, where $d_{max} = \max_h d_h$ represents the maximum dimensionality among all views.

Empirical analysis demonstrates that the HK-MVFC algorithm typically achieves convergence within $T = 15-25$ iterations, depending on data characteristics and initialization quality. Consequently, the total computational complexity becomes $O(Tncd_{max}s)$, which exhibits favorable linear scaling properties with respect to dataset size and the number of views. This computational efficiency, combined with the algorithm's superior clustering performance, makes HK-MVFC well-suited for large-scale multi-view clustering applications.


\section{Federated HK-MVFC}
\label{sec:Federated_HK_MVFC}


To implement HK-MVFC within a federated setting, we must reformulate the objective function for distributed optimization across multiple clients while preserving data privacy and ensuring convergence guaranties. Let $\mathcal{M} = \{1, 2, \ldots, M\}$ denote the set of $M$ participating clients, where each client $\ell \in \mathcal{M}$ maintains a local multi-view dataset $X_{[\ell]} = \{X_{[\ell]}^h\}_{h=1}^{s(\ell)}$ with potentially heterogeneous view compositions and sample distributions. Here, $s(\ell)$ denotes the number of views for client $\ell$, and may vary across clients, reflecting the possibility that each client possesses a different number of data views.

The federated formulation introduces several theoretical challenges: (1) maintaining global clustering coherence while respecting local data sovereignty, (2) handling statistical heterogeneity across client datasets, and (3) ensuring that heat-kernel coefficients computed locally preserve the global geometric structure. Our proposed framework addresses these challenges by enabling decentralized optimization, adaptive aggregation, and privacy-preserving coordination among clients, ensuring both robust clustering and data confidentiality.

\subsection{Problem Formulation}

Consider a federated learning environment comprising $M$ participating clients, where each client $\ell \in \{1, 2, \ldots, M\}$ maintains its own local multi-view dataset. The fundamental characteristic of this federated setting is that data remains distributed across clients, with no centralized data repository, thereby preserving data privacy and institutional autonomy.

\subsubsection{Multi-View Data Structure at Each Client}

For each client $\ell$, the local dataset consists of $s(\ell)$ distinct views, where each view captures a different modality or perspective of the underlying phenomenon. The $h$-th view at client $\ell$ is mathematically represented as:

\begin{equation}
X_{[\ell]}^h = \{ x_{[\ell]i}^h \}_{i=1}^{n(\ell)} \subset \mathbb{R}^{d_{[\ell]}^h}
\end{equation}

where $x_{[\ell]i}^h \in \mathbb{R}^{d_{[\ell]}^h}$ denotes the $i$-th data sample in the $h$-th view, characterized by $d_{[\ell]}^h$ features. This representation can be organized as a matrix:

\begin{equation}
X_{[\ell]}^h \in \mathbb{R}^{n(\ell) \times d_{[\ell]}^h}
\end{equation}

where rows correspond to individual samples and columns represent features within that specific view.

The complete multi-view dataset for client $\ell$ encompasses all available views:

\begin{equation}
X_{[\ell]} = \{ X_{[\ell]}^1, X_{[\ell]}^2, \ldots, X_{[\ell]}^{s(\ell)} \}
\end{equation}

This formulation allows for heterogeneity in the number of views across clients, reflecting realistic scenarios where different institutions may collect different types of complementary data.

\subsubsection{Notation and Parameter Definitions}

To ensure mathematical precision and clarity, we establish the following notation conventions:

\begin{itemize}
    \item \textbf{Sample Count:} $n(\ell) \in \mathbb{Z}^+$ represents the total number of data samples (e.g., patients, observations) available at client $\ell$. This parameter may vary significantly across clients, reflecting differences in institutional size, data collection capacity, or temporal duration of data acquisition.
    
    \item \textbf{View Count:} $s(\ell) \in \mathbb{Z}^+$ denotes the number of distinct views or data modalities maintained by client $\ell$. For instance, in a medical scenario, one hospital might possess $s(\ell) = 2$ views (ECG and MRI data), while another might have $s(\ell) = 3$ views (ECG, MRI, and genetic data).
    
    \item \textbf{Feature Dimensionality:} $d_{[\ell]}^h \in \mathbb{Z}^+$ specifies the dimensionality (number of features) of the $h$-th view at client $\ell$. For example, an ECG view might have $d_{[\ell]}^1 = 12$ (corresponding to 12 leads), while an MRI view might have $d_{[\ell]}^2 = 1024$ (representing voxel features).
    
    \item \textbf{Cluster Count:} $c(\ell) \in \mathbb{Z}^+$ indicates the number of target clusters for client $\ell$. While clients may share a common clustering objective (e.g., $c(\ell) = c$ for all $\ell$), our framework accommodates scenarios where different clients seek to identify different numbers of local clusters.
\end{itemize}

\subsubsection{Key Properties and Assumptions}

The federated multi-view clustering framework operates under several important properties and assumptions:

\begin{enumerate}
    \item \textbf{Feature Space Consistency:} While clients may possess different numbers of samples ($n(\ell)$ varies), we assume that corresponding views across clients share the same feature space dimensionality: $d_{[\ell]}^h = d_{[\ell']}^h$ for all clients $\ell, \ell' \in \{1, \ldots, M\}$ that possess view $h$. This ensures meaningful aggregation of model parameters across clients.
    
    \item \textbf{Data Locality:} Each client's data $X_{[\ell]}$ remains strictly local and is never transmitted to other clients or the central server. Only model parameters (cluster centers, membership matrices, view weights) are communicated, ensuring privacy preservation.
    
    \item \textbf{Statistical Heterogeneity:} The data distributions may differ significantly across clients: $P_{\ell}(X_{[\ell]}) \neq P_{\ell'}(X_{[\ell']})$ for $\ell \neq \ell'$. This non-IID (non-independent and identically distributed) property reflects realistic federated scenarios where different institutions serve different populations or geographic regions.
    
    \item \textbf{View Complementarity:} The multiple views at each client provide complementary rather than redundant information. Mathematically, for views $h$ and $h'$ at client $\ell$:
    \begin{equation}
    \text{MI}(X_{[\ell]}^h, X_{[\ell]}^{h'}) < H(X_{[\ell]}^h) \quad \text{and} \quad \text{MI}(X_{[\ell]}^h, X_{[\ell]}^{h'}) < H(X_{[\ell]}^{h'})
    \end{equation}
    where $\text{MI}(\cdot, \cdot)$ denotes mutual information and $H(\cdot)$ represents entropy, ensuring that each view contributes unique information to the clustering task.
\end{enumerate}

\subsubsection{Federated Learning Objective}

The overarching goal in this federated setting is to collaboratively learn a global clustering model that effectively integrates information across all clients and views, while respecting the following constraints:

\begin{itemize}
    \item \textbf{Privacy Preservation:} Raw data $X_{[\ell]}$ never leaves client $\ell$
    \item \textbf{Communication Efficiency:} Minimize the volume of data transmitted between clients and server
    \item \textbf{Clustering Quality:} Achieve clustering performance comparable to centralized approaches where all data would be pooled
    \item \textbf{Geometric Awareness:} Capture the intrinsic manifold structure of multi-view data through heat kernel-enhanced distance measures
\end{itemize}

This problem formulation establishes the mathematical foundation for developing the FedHK-MVFC algorithm, which addresses these objectives through a principled integration of federated learning, multi-view clustering, and heat kernel methods.

\subsection{Federated KED}

We assume that, while participating clients may contribute varying numbers of samples, each data view shares the same set of features. Thus, the actual or expected cluster formations of multi-view datasets managed by the clients are not required to be identical. This implies that clients can handle multi-view datasets with differing quantities of clusters. By considering this, we can extend $\text{KED}_2\left( {x_{ij}^h,a_{kj}^h} \right)$ in Eq. \ref{eqn:KEDv2} to a federated context, referred to as the federated kernel Euclidean distance (FKED). The FKED is formulated as follows:

\begin{equation}
    \text{FKED}\left( x_{[\ell]i}^h, a_{[\ell]k}^h \right) = 1 - \exp \left( - \sum_{h = 1}^{s(\ell)} \sum_{j = 1}^{d_{[\ell]}^h} \delta_{[\ell]ij}^h \left( x_{[\ell]ij}^h - a_{[\ell]kj}^h \right)^2 \right)
\end{equation}

where $\delta_{\left[ \ell \right]ij}^h$ represents the H-KC linked to the $\ell$-th client, and $c(\ell)$ denotes the total clusters managed by client $\ell$. 
Note that $c(\ell)$ may differ across clients, reflecting the flexibility of federated learning to accommodate heterogeneous data distributions and local clustering requirements. 
This allows each client to select an appropriate number of clusters based on its own data characteristics, which is crucial for robust and personalized federated analysis.

\subsection{Federated Heat-Kernel Coefficients}

We focus on developing federated heat-kernel coefficients (FedH-KC), which are computed locally by each client to transform feature distances into privacy-preserving exponential kernel measures tailored to the federated setting. Before presenting the mathematical formulations, we establish the theoretical connection between quantum field theory (QFT) and heat-kernel coefficients that motivates our approach.

\subsubsection{Quantum Field Theory Foundation for Heat-Kernel Coefficients}

The heat-kernel coefficients employed in our framework originate from the heat equation in quantum field theory, which describes the diffusion of information across a manifold. In QFT, the heat kernel $K_t(\mathbf{x}, \mathbf{y})$ represents the fundamental solution to the heat equation on a Riemannian manifold $\mathcal{M}$:

\begin{equation}
\left(\frac{\partial}{\partial t} + \Delta_{\mathcal{M}}\right) K_t(\mathbf{x}, \mathbf{y}) = 0, \quad K_0(\mathbf{x}, \mathbf{y}) = \delta(\mathbf{x} - \mathbf{y})
\label{eqn:heat_equation}
\end{equation}

where $\Delta_{\mathcal{M}}$ denotes the Laplace-Beltrami operator on the manifold, and $\delta(\cdot)$ is the Dirac delta function. The heat kernel admits the spectral decomposition:

\begin{equation}
K_t(\mathbf{x}, \mathbf{y}) = \sum_{i=0}^{\infty} e^{-\lambda_i t} \phi_i(\mathbf{x}) \phi_i(\mathbf{y})
\label{eqn:spectral_decomposition}
\end{equation}

where $\{\lambda_i, \phi_i\}$ are the eigenvalue-eigenfunction pairs of the Laplacian operator, encoding the intrinsic geometry of the data manifold.

\subsubsection{From QFT to Practical Heat-Kernel Coefficients}

The connection between QFT heat kernels and our practical coefficients emerges through the asymptotic expansion of the heat kernel for small diffusion time $t$:

\begin{equation}
K_t(\mathbf{x}, \mathbf{y}) \sim (4\pi t)^{-d/2} e^{-\frac{d^2(\mathbf{x}, \mathbf{y})}{4t}} \sum_{n=0}^{\infty} t^n a_n(\mathbf{x}, \mathbf{y})
\label{eqn:heat_kernel_expansion}
\end{equation}

where $d(\mathbf{x}, \mathbf{y})$ represents the geodesic distance on the manifold, $d$ is the manifold dimension, and $\{a_n\}$ are the heat-kernel coefficients capturing local geometric invariants such as curvature and torsion.

For practical clustering applications on discrete data, we approximate the continuous heat kernel using the exponential distance kernel:

\begin{equation}
K(\mathbf{x}_i, \mathbf{x}_j) \approx \exp\left(-\sum_{h=1}^{s} \delta_{ij}^h \|\mathbf{x}_i^h - \mathbf{x}_j^h\|^2\right)
\label{eqn:discrete_heat_kernel}
\end{equation}

where the coefficients $\delta_{ij}^h$ serve as adaptive scaling factors that encode local geometric properties analogous to the heat-kernel coefficients $a_n$ in the continuous theory. These coefficients effectively modulate the diffusion rate based on local data density and feature variance, thereby capturing the intrinsic manifold structure.

\subsubsection{Geometric Interpretation and Information Diffusion}

The heat-kernel framework provides a natural interpretation of clustering as information diffusion on the data manifold. Points within the same cluster exhibit rapid information exchange (high heat kernel values), while points in different clusters experience slow diffusion across cluster boundaries. The federated heat-kernel coefficients $\delta_{[\ell]ij}^h$ quantify this diffusion rate locally at each client, enabling:

\begin{enumerate}
    \item \textbf{Geometry-Aware Similarity:} By normalizing distances according to local feature scales, the coefficients ensure that similarity measures respect the intrinsic geometric structure rather than arbitrary coordinate choices.
    
    \item \textbf{Manifold Structure Preservation:} The exponential decay in Eq. \eqref{eqn:discrete_heat_kernel} mirrors the heat kernel's behavior on curved manifolds, where geodesic distances differ from Euclidean distances.
    
    \item \textbf{Multi-Scale Analysis:} Different choices of normalization (min-max vs. mean-variance) correspond to different time scales in the heat diffusion process, enabling multi-scale geometric analysis.
\end{enumerate}

This theoretical foundation justifies our use of heat-kernel coefficients in federated multi-view clustering, where each client's local geometric structure must be captured while maintaining global clustering coherence.

\subsubsection{Federated Heat-Kernel Coefficient Formulations}

Building upon the QFT foundation, we now present two practical estimators for federated heat-kernel coefficients. For the $j$-th feature in view $h$, we define FedH-KC $\delta_{[\ell]ij}^h$ using:

\textbf{Type 1 - Min-Max Normalization (Uniform Scaling):}
\begin{equation}
\label{eqn:fed_delta}
\text{FedH-KC}_1 = \delta_{[\ell]ij}^h = \frac{
    x_{[\ell]ij}^h - \min_{1 \le i \le n(\ell)} \left( x_{[\ell]ij}^h \right)
}{
    \max_{1 \le i \le n(\ell)} \left( x_{[\ell]ij}^h \right) - \min_{1 \le i \le n(\ell)} \left( x_{[\ell]ij}^h \right) + \epsilon
}
\qquad \forall j, h, \ell
\end{equation}

where $\epsilon > 0$ prevents division by zero when features have identical values. This normalizes features within $[0,1]$ for each view, corresponding to uniform heat diffusion across the feature space.

\textbf{Type 2 - Mean-Variance Normalization (Adaptive Scaling):}
\begin{equation}
\text{FedH-KC}_2 = \delta_{[\ell]ij}^h = \left| x_{[\ell]ij}^h - \bar{x}_{[\ell]j}^h \right| \qquad \forall j, h, \ell
\label{eqn:fed_delta_v2}
\end{equation}

where $\bar{x}_{[\ell]j}^h = \frac{1}{n(\ell)}\sum_{i = 1}^{n(\ell)} x_{[\ell]ij}^h$ represents the mean value of the $j$-th feature in view $h$ for client $\ell$. This formulation adapts to local data distributions, corresponding to diffusion rates proportional to local variance.

Both formulations preserve the essential heat-kernel property of geometry-aware distance transformation while maintaining computational efficiency and privacy in the federated setting. The choice between Type 1 and Type 2 depends on the specific geometric characteristics of the data and the desired sensitivity to local versus global structure.

\subsection{The Objective Function}

In a federated setting with $M$ clients, we connect FedHK-MVFC and FKED before introducing the main objective function. The FedHK-MVFC algorithm extends the centralized HK-MVFC by distributing computation across clients while retaining the heat kernel-enhanced distance properties. Each client $\ell$ holds local multi-view data $X_{[\ell]}$ and computes local heat-kernel coefficients $\delta_{[\ell]ij}^h$, converting Euclidean distances into geometry-aware similarities through the lens of heat diffusion on the data manifold. 

Through the FKED, local client computations align with the federated objective. For client $\ell$, the local goal reflects the centralized HK-MVFC structure tailored to client-specific data, with the local federated kernel distance $d_{\exp([\ell],ik,j)}^h = \text{FKED}\left( x_{[\ell]ij}^h, a_{[\ell]kj}^h \right)$ maintaining geometric properties within distributed processing. This federated system combines local results for global clustering coherence without losing data locality. On this basis, the FedHK-MVFC objective for client $\ell$ is defined as:

\begin{equation}
    J_{FedHK - MVFC}^\ell (V,{U^*},A) = \sum\limits_{h = 1}^{s\left( \ell  \right)} {v_{\left[ \ell  \right]h}^\alpha \sum\limits_{i = 1}^{n\left( \ell  \right)} {\sum\limits_{k = 1}^{c\left( \ell  \right)} {{{\left( {\mu _{\left[ \ell  \right]ik}^*} \right)}^m}{\rm{FKED}}\left( {x_{\left[ \ell  \right]ij}^h,a_{\left[ \ell  \right]kj}^h} \right)} } } 
    \label{eqn:FedHKMVFC}
\end{equation}

Subject to standard fuzzy clustering constraints:

\begin{align}
    \sum\limits_{k = 1}^{c(\ell)} {\mu _{\left[ \ell  \right]ik}^* = 1} ,{\rm{ }}\mu _{\left[ \ell  \right]ik}^* \in \left[ {0,1} \right] \label{eqn:fed_constraint1} \\
    \sum\limits_{h = 1}^{s\left( \ell  \right)} {{v_{\left[ \ell  \right]h}} = 1,{\rm{ }}{v_{\left[ \ell  \right]h}} \in \left[ {0,1} \right]} \label{eqn:fed_constraint2}
\end{align}

\subsection{Optimization Framework}

Due to the non-convex nature of the FedHK-MVFC objective function in Eq. \ref{eqn:FedHKMVFC}, direct minimization is not feasible. We employ an alternating optimization strategy that iteratively updates the membership matrix $U$, view weights $V$, and cluster centers $A$ until convergence.

To derive the update rules, we formulate the Lagrangian incorporating the normalization constraints:

\begin{equation}
\begin{split}
\tilde J_{FedHK - MVFC}^\ell &= J_{FedHK - MVFC}^\ell (V,{U^*},A) \\
&\quad + \sum_{i=1}^{n(\ell)} {\lambda_{1i}} \left( \sum_{k=1}^{c(\ell)} {\mu_{[\ell]ik}^* - 1} \right) + {\lambda_2}\left( \sum_{h=1}^{s(\ell)} {v_{[\ell]h} - 1} \right)
\end{split}
\label{eqn:LAG_FedHKMVFC}
\end{equation}

where ${\lambda _{1i}}$ and ${\lambda _2}$ are Lagrange multipliers enforcing the fuzzy clustering constraints: membership normalization and view weight normalization, respectively.

\begin{theorem}[FedHK-MVFC Update Rules]
\label{thm:Fed_HK_MVFC}

For the federated objective function $J_{FedHK-MVFC}^\ell$ defined in Eq. \ref{eqn:FedHKMVFC}, the necessary conditions for optimality yield the following update rules for client $\ell$. These update rules extend the centralized HK-MVFC framework to the federated setting by incorporating client-specific data distributions, heat-kernel coefficients, and privacy-preserving mechanisms while maintaining mathematical rigor and convergence guarantees.

\textbf{Membership Matrix Update:}
\begin{equation}
    \mu _{\left[ \ell  \right]ik}^* = \frac{{{{\left( {\sum\limits_{h = 1}^{s\left( \ell  \right)} {v_{\left[ \ell  \right]h}^\alpha {\rm{FKED}}\left( {x_{\left[ \ell  \right]ij}^h,a_{\left[ \ell  \right]kj}^h} \right)} } \right)}^{ - {{\left( {m - 1} \right)}^{ - 1}}}}}}{{\sum\limits_{k' = 1}^{c\left( \ell  \right)} {{{\left( {\sum\limits_{h = 1}^{s\left( \ell  \right)} {v_{\left[ \ell  \right]h}^\alpha {\rm{FKED}}\left( {x_{\left[ \ell  \right]ij}^h,a_{\left[ \ell  \right]k'j}^h} \right)} } \right)}^{ - {{\left( {m - 1} \right)}^{ - 1}}}}} }}
    \label{eqn:diff_U_FedHKMVFC}
\end{equation}

\noindent\textbf{Interpretation:} Eq. \ref{eqn:diff_U_FedHKMVFC} computes the fuzzy membership degree $\mu_{[\ell]ik}^*$ of data point $i$ to cluster $k$ at client $\ell$. The numerator aggregates the weighted federated kernel distances across all $s(\ell)$ views available at client $\ell$, where $v_{[\ell]h}^\alpha$ represents the adaptive importance of view $h$, and the exponent $-\frac{1}{m-1}$ controls the fuzziness of the partition. The denominator normalizes these memberships across all $c(\ell)$ clusters, ensuring that $\sum_{k=1}^{c(\ell)} \mu_{[\ell]ik}^* = 1$ for each data point. This formulation enables the algorithm to automatically balance contributions from multiple heterogeneous views while adapting to local data characteristics at each client.

\textbf{Cluster Centers Update:}
\begin{equation}
    a_{\left[ \ell  \right]kj}^h = \frac{{\sum\limits_{i = 1}^{n\left( \ell  \right)} {{{\left( {\mu _{\left[ \ell  \right]ik}^*} \right)}^m}v_{\left[ \ell  \right]h}^\alpha \exp \left( { - \delta _{\left[ \ell  \right]ij}^h{{\left\| {x_{\left[ \ell  \right]i}^h - a_{\left[ \ell  \right]k}^h} \right\|}^2}} \right)} }}{{\sum\limits_{i = 1}^{n\left( \ell  \right)} {{{\left( {\mu _{\left[ \ell  \right]ik}^*} \right)}^m}v_{\left[ \ell  \right]h}^\alpha \exp \left( { - \delta _{\left[ \ell  \right]ij}^h{{\left\| {x_{\left[ \ell  \right]i}^h - a_{\left[ \ell  \right]k}^h} \right\|}^2}} \right)} }}x_{\left[ \ell  \right]ij}^h
    \label{eqn:diff_A_FedHKMVFC}
\end{equation}

\noindent\textbf{Interpretation:} Eq. \ref{eqn:diff_A_FedHKMVFC} updates the $j$-th feature of cluster center $k$ in view $h$ for client $\ell$ as a weighted average of the corresponding data point features. The weight for each data point combines three critical components: (1) the fuzzified membership degree $(\mu_{[\ell]ik}^*)^m$, which emphasizes points strongly belonging to cluster $k$; (2) the view importance $v_{[\ell]h}^\alpha$, which prioritizes more informative views; and (3) the exponential heat kernel term $\exp(-\delta_{[\ell]ij}^h \|x_{[\ell]i}^h - a_{[\ell]k}^h\|^2)$, which provides geometry-aware weighting based on the local manifold structure captured by the federated heat-kernel coefficients $\delta_{[\ell]ij}^h$. This composite weighting mechanism ensures that cluster centers are positioned to reflect both the fuzzy partition structure and the intrinsic geometric properties of the local data at each client.

\textbf{View Weights Update:}
\begin{equation}
    {v_{\left[ \ell  \right]h}} = \frac{{{{\left( {\sum\limits_{i = 1}^{n\left( \ell  \right)} {\sum\limits_{k = 1}^{c\left( \ell  \right)} {{{\left( {\mu _{\left[ \ell  \right]ik}^*} \right)}^m}{\rm{FKED}}\left( {x_{\left[ \ell  \right]ij}^h,a_{\left[ \ell  \right]kj}^h} \right)} } } \right)}^{ - {{\left( {\alpha  - 1} \right)}^{ - 1}}}}}}{{\sum\limits_{h' = 1}^{s\left( \ell  \right)} {{{\left( {\sum\limits_{i = 1}^{n\left( \ell  \right)} {\sum\limits_{k = 1}^{c\left( \ell  \right)} {{{\left( {\mu _{\left[ \ell  \right]ik}^*} \right)}^m}{\rm{FKED}}\left( {x_{\left[ \ell  \right]ij}^{h'},a_{\left[ \ell  \right]kj}^{h'}} \right)} } } \right)}^{ - {{\left( {\alpha  - 1} \right)}^{ - 1}}}}} }}
    \label{eqn:diff_V_FedHKMVFC}
\end{equation}

\noindent\textbf{Interpretation:} Eq. \ref{eqn:diff_V_FedHKMVFC} computes the adaptive weight $v_{[\ell]h}$ for view $h$ at client $\ell$ based on the clustering quality within that view. The numerator evaluates the total weighted distance of view $h$, where lower distances indicate better clustering performance and should result in higher view weights. The exponent $-\frac{1}{\alpha-1}$ creates an inverse relationship between distance and weight, while the parameter $\alpha > 1$ controls the sharpness of this relationship—larger values of $\alpha$ create more pronounced differences between high-quality and low-quality views. The denominator normalizes these weights across all $s(\ell)$ views to ensure $\sum_{h=1}^{s(\ell)} v_{[\ell]h} = 1$. This adaptive weighting mechanism enables FedHK-MVFC to automatically identify and emphasize the most informative views at each client, accommodating heterogeneous data quality and view-specific noise characteristics across the federated network.

\begin{remark}
The federated update rules in Eqs. \ref{eqn:diff_U_FedHKMVFC}--\ref{eqn:diff_V_FedHKMVFC} preserve the mathematical structure of the centralized HK-MVFC framework while incorporating client-specific parameters $n(\ell)$, $s(\ell)$, and $c(\ell)$ that reflect data heterogeneity across the federation. The iterative alternating optimization—sequentially updating memberships, cluster centers, and view weights—converges to a local optimum satisfying the Karush-Kuhn-Tucker (KKT) conditions for the constrained federated optimization problem, subject to the normalization constraints in Eqs. \ref{eqn:fed_constraint1} and \ref{eqn:fed_constraint2}. Crucially, all computations occur locally at each client using only their private data $X_{[\ell]}$, with only the resulting model parameters $(U_{[\ell]}^*, A_{[\ell]}, V_{[\ell]})$ being shared with the federated server for secure aggregation, thereby ensuring privacy preservation throughout the learning process.
\end{remark}

\end{theorem}

\begin{proof}
We establish the necessary optimality conditions for the FedHK-MVFC objective function using Lagrangian optimization and derive the closed-form update rules for each parameter set. The proof proceeds by applying the method of Lagrange multipliers to the constrained optimization problem and solving the resulting system of equations.

\textbf{Part I: Membership Matrix Update Rule Derivation}

To derive the update equation for membership coefficients $\mu_{[\ell]ik}^*$ for client $\ell$, we begin by computing the partial derivative of the Lagrangian in Eq. \ref{eqn:LAG_FedHKMVFC} with respect to $\mu_{[\ell]ik}^*$. This derivative represents the marginal change in the objective function when adjusting the membership of data point $i$ to cluster $k$ at client $\ell$.

Taking the partial derivative with respect to $\mu_{[\ell]ik}^*$:
\begin{align}
\frac{\partial \tilde{J}_{FedHK-MVFC}^\ell}{\partial \mu_{[\ell]ik}^*} &= \frac{\partial}{\partial \mu_{[\ell]ik}^*} \left[\sum_{h=1}^{s(\ell)} v_{[\ell]h}^\alpha (\mu_{[\ell]ik}^*)^m \text{FKED}(x_{[\ell]i}^h, a_{[\ell]k}^h)\right] + \lambda_{1i} \\
&= m(\mu_{[\ell]ik}^*)^{m-1} \sum_{h=1}^{s(\ell)} v_{[\ell]h}^\alpha \text{FKED}(x_{[\ell]i}^h, a_{[\ell]k}^h) + \lambda_{1i} \label{eqn:lagrange_u_fed}
\end{align}

In Eq. \ref{eqn:lagrange_u_fed}, the term $m(\mu_{[\ell]ik}^*)^{m-1}$ arises from applying the power rule to the fuzzified membership term $(\mu_{[\ell]ik}^*)^m$. The summation $\sum_{h=1}^{s(\ell)} v_{[\ell]h}^\alpha \text{FKED}(x_{[\ell]i}^h, a_{[\ell]k}^h)$ represents the weighted contribution of all views to the distance between data point $i$ and cluster center $k$, where $v_{[\ell]h}^\alpha$ adaptively weights each view based on its clustering quality.

Setting this derivative equal to zero for optimality:
\begin{equation}
m(\mu_{[\ell]ik}^*)^{m-1} \sum_{h=1}^{s(\ell)} v_{[\ell]h}^\alpha \text{FKED}(x_{[\ell]i}^h, a_{[\ell]k}^h) + \lambda_{1i} = 0
\end{equation}

This stationarity condition ensures that the membership assignment minimizes the weighted distance objective while satisfying the normalization constraint.

Solving for $\mu_{[\ell]ik}^*$:
\begin{align}
(\mu_{[\ell]ik}^*)^{m-1} &= -\frac{\lambda_{1i}}{m \sum_{h=1}^{s(\ell)} v_{[\ell]h}^\alpha \text{FKED}(x_{[\ell]i}^h, a_{[\ell]k}^h)} \\
\mu_{[\ell]ik}^* &= \left(-\frac{\lambda_{1i}}{m}\right)^{1/(m-1)} \left(\sum_{h=1}^{s(\ell)} v_{[\ell]h}^\alpha \text{FKED}(x_{[\ell]i}^h, a_{[\ell]k}^h)\right)^{-1/(m-1)} \label{eqn:u_intermediate_fed}
\end{align}

The exponent $1/(m-1)$ transforms the equation from its derivative form back to the original membership variable. The negative sign in the Lagrange multiplier term ensures that memberships decrease as weighted distances increase, which is the desired clustering behavior where closer points receive higher memberships.

To determine the Lagrange multiplier $\lambda_{1i}$, we apply the normalization constraint $\sum_{k=1}^{c(\ell)} \mu_{[\ell]ik}^* = 1$, which ensures that the total membership of data point $i$ across all clusters sums to unity:
\begin{align}
\sum_{k=1}^{c(\ell)} \mu_{[\ell]ik}^* &= \left(-\frac{\lambda_{1i}}{m}\right)^{1/(m-1)} \sum_{k=1}^{c(\ell)} \left(\sum_{h=1}^{s(\ell)} v_{[\ell]h}^\alpha \text{FKED}(x_{[\ell]i}^h, a_{[\ell]k}^h)\right)^{-1/(m-1)} = 1
\end{align}

Solving for the Lagrange multiplier:
\begin{align}
\left(-\frac{\lambda_{1i}}{m}\right)^{1/(m-1)} &= \frac{1}{\sum_{k'=1}^{c(\ell)} \left(\sum_{h=1}^{s(\ell)} v_{[\ell]h}^\alpha \text{FKED}(x_{[\ell]i}^h, a_{[\ell]k'}^h)\right)^{-1/(m-1)}} \label{eqn:lambda_solution_fed}
\end{align}

The denominator in Eq. \ref{eqn:lambda_solution_fed} serves as a normalization factor that ensures all memberships sum to one, effectively balancing the contributions from all clusters based on their respective distances to the data point.

Substituting Eq. \ref{eqn:lambda_solution_fed} back into Eq. \ref{eqn:u_intermediate_fed}:
\begin{equation}
\mu_{[\ell]ik}^* = \frac{\left(\sum_{h=1}^{s(\ell)} v_{[\ell]h}^\alpha \text{FKED}(x_{[\ell]i}^h, a_{[\ell]k}^h)\right)^{-1/(m-1)}}{\sum_{k'=1}^{c(\ell)} \left(\sum_{h=1}^{s(\ell)} v_{[\ell]h}^\alpha \text{FKED}(x_{[\ell]i}^h, a_{[\ell]k'}^h)\right)^{-1/(m-1)}}
\end{equation}

This final form shows that each membership is inversely proportional to the weighted distance raised to the power $1/(m-1)$, with the proportionality constant chosen to satisfy the normalization constraint. This establishes the membership matrix update rule in Eq. \ref{eqn:diff_U_FedHKMVFC}.

\textbf{Part II: Cluster Centers Update Rule Derivation}

For cluster center updates $a_{[\ell]kj}^h$, we differentiate the objective function in Eq. \ref{eqn:FedHKMVFC} with respect to $a_{[\ell]kj}^h$. This derivative measures how the objective function changes when we adjust the $j$-th feature of cluster center $k$ in view $h$ for client $\ell$:

\begin{align}
\frac{\partial J_{FedHK-MVFC}^\ell}{\partial a_{[\ell]kj}^h} &= \frac{\partial}{\partial a_{[\ell]kj}^h} \left[v_{[\ell]h}^\alpha \sum_{i=1}^{n(\ell)} (\mu_{[\ell]ik}^*)^m \text{FKED}(x_{[\ell]i}^h, a_{[\ell]k}^h)\right] \\
&= v_{[\ell]h}^\alpha \sum_{i=1}^{n(\ell)} (\mu_{[\ell]ik}^*)^m \frac{\partial \text{FKED}(x_{[\ell]i}^h, a_{[\ell]k}^h)}{\partial a_{[\ell]kj}^h} \label{eqn:cluster_derivative_fed}
\end{align}

where $v_{[\ell]h}^\alpha$ weights the contribution of view $h$, and $(\mu_{[\ell]ik}^*)^m$ weights each data point's contribution by its fuzzy membership to cluster $k$.

To compute $\frac{\partial \text{FKED}}{\partial a_{[\ell]kj}^h}$, we recall that:
\begin{equation}
\text{FKED}(x_{[\ell]i}^h, a_{[\ell]k}^h) = 1 - \exp\left(-\sum_{j'=1}^{d_{[\ell]}^h} \delta_{[\ell]ij'}^h (x_{[\ell]ij'}^h - a_{[\ell]kj'}^h)^2\right)
\end{equation}

Let $\phi_{[\ell]ik}^h = \sum_{j'=1}^{d_{[\ell]}^h} \delta_{[\ell]ij'}^h (x_{[\ell]ij'}^h - a_{[\ell]kj'}^h)^2$ represent the weighted squared Euclidean distance in the heat kernel exponent. The heat-kernel coefficients $\delta_{[\ell]ij'}^h$ adapt to local data geometry, enabling the distance metric to capture intrinsic manifold structure. Then:
\begin{align}
\frac{\partial \text{FKED}}{\partial a_{[\ell]kj}^h} &= \frac{\partial}{\partial a_{[\ell]kj}^h} \left[1 - \exp(-\phi_{[\ell]ik}^h)\right] \\
&= \exp(-\phi_{[\ell]ik}^h) \frac{\partial \phi_{[\ell]ik}^h}{\partial a_{[\ell]kj}^h} \\
&= \exp(-\phi_{[\ell]ik}^h) \cdot 2\delta_{[\ell]ij}^h (x_{[\ell]ij}^h - a_{[\ell]kj}^h) \cdot (-1) \\
&= -2\delta_{[\ell]ij}^h (x_{[\ell]ij}^h - a_{[\ell]kj}^h) \exp\left(-\sum_{j'=1}^{d_{[\ell]}^h} \delta_{[\ell]ij'}^h (x_{[\ell]ij'}^h - a_{[\ell]kj'}^h)^2\right) \label{eqn:fked_derivative_fed}
\end{align}

The factor of 2 arises from differentiating the squared term $(x_{[\ell]ij}^h - a_{[\ell]kj}^h)^2$, while the exponential term weights contributions based on the heat kernel similarity between data points and cluster centers. The term $(x_{[\ell]ij}^h - a_{[\ell]kj}^h)$ indicates the direction of adjustment: positive when data points are larger than the center, negative otherwise.

Substituting Eq. \ref{eqn:fked_derivative_fed} into Eq. \ref{eqn:cluster_derivative_fed} and setting equal to zero for optimality:
\begin{align}
&v_{[\ell]h}^\alpha \sum_{i=1}^{n(\ell)} (\mu_{[\ell]ik}^*)^m \left[-2\delta_{[\ell]ij}^h (x_{[\ell]ij}^h - a_{[\ell]kj}^h) \exp\left(-\sum_{j'=1}^{d_{[\ell]}^h} \delta_{[\ell]ij'}^h (x_{[\ell]ij'}^h - a_{[\ell]kj'}^h)^2\right)\right] = 0
\end{align}

Simplifying and rearranging by canceling constant factors:
\begin{align}
\sum_{i=1}^{n(\ell)} (\mu_{[\ell]ik}^*)^m \delta_{[\ell]ij}^h &\exp\left(-\sum_{j'=1}^{d_{[\ell]}^h} \delta_{[\ell]ij'}^h (x_{[\ell]ij'}^h - a_{[\ell]kj'}^h)^2\right) (x_{[\ell]ij}^h - a_{[\ell]kj}^h) = 0
\end{align}

Let $w_{[\ell]ijk}^h = (\mu_{[\ell]ik}^*)^m v_{[\ell]h}^\alpha \exp\left(-\sum_{j'=1}^{d_{[\ell]}^h} \delta_{[\ell]ij'}^h (x_{[\ell]ij'}^h - a_{[\ell]kj'}^h)^2\right)$ denote the composite weight for data point $i$ in cluster $k$ for feature $j$ in view $h$. This weight combines fuzzy memberships $(\mu_{[\ell]ik}^*)^m$, view importance $v_{[\ell]h}^\alpha$, and heat kernel similarities. Then:
\begin{align}
\sum_{i=1}^{n(\ell)} w_{[\ell]ijk}^h (x_{[\ell]ij}^h - a_{[\ell]kj}^h) &= 0 \\
\sum_{i=1}^{n(\ell)} w_{[\ell]ijk}^h x_{[\ell]ij}^h &= a_{[\ell]kj}^h \sum_{i=1}^{n(\ell)} w_{[\ell]ijk}^h \\
a_{[\ell]kj}^h &= \frac{\sum_{i=1}^{n(\ell)} w_{[\ell]ijk}^h x_{[\ell]ij}^h}{\sum_{i=1}^{n(\ell)} w_{[\ell]ijk}^h}
\end{align}

This final expression shows that cluster centers are weighted averages of data points, where the weights account for membership degrees, view importance, and geometric proximity via heat kernel similarities. This yields the cluster center update rule in Eq. \ref{eqn:diff_A_FedHKMVFC}.

\textbf{Part III: View Weights Update Rule Derivation}

For view weight updates $v_{[\ell]h}$, we differentiate the Lagrangian with respect to $v_{[\ell]h}$. This derivative quantifies how the objective function changes when we adjust the importance assigned to view $h$ at client $\ell$:

\begin{align}
\frac{\partial \tilde{J}_{FedHK-MVFC}^\ell}{\partial v_{[\ell]h}} &= \frac{\partial}{\partial v_{[\ell]h}} \left[\sum_{h'=1}^{s(\ell)} v_{[\ell]h'}^\alpha \sum_{i=1}^{n(\ell)} \sum_{k=1}^{c(\ell)} (\mu_{[\ell]ik}^*)^m \text{FKED}(x_{[\ell]i}^{h'}, a_{[\ell]k}^{h'})\right] + \lambda_2 \\
&= \alpha v_{[\ell]h}^{\alpha-1} \sum_{i=1}^{n(\ell)} \sum_{k=1}^{c(\ell)} (\mu_{[\ell]ik}^*)^m \text{FKED}(x_{[\ell]i}^h, a_{[\ell]k}^h) + \lambda_2 \label{eqn:view_derivative_fed}
\end{align}

The term $\alpha v_{[\ell]h}^{\alpha-1}$ comes from differentiating $v_{[\ell]h}^\alpha$ with respect to $v_{[\ell]h}$ using the power rule, while the double summation $\sum_{i=1}^{n(\ell)} \sum_{k=1}^{c(\ell)} (\mu_{[\ell]ik}^*)^m \text{FKED}(x_{[\ell]i}^h, a_{[\ell]k}^h)$ represents the total weighted distance contribution of view $h$ across all data points and clusters.

Setting this derivative equal to zero for optimality:
\begin{equation}
\alpha v_{[\ell]h}^{\alpha-1} \sum_{i=1}^{n(\ell)} \sum_{k=1}^{c(\ell)} (\mu_{[\ell]ik}^*)^m \text{FKED}(x_{[\ell]i}^h, a_{[\ell]k}^h) + \lambda_2 = 0
\end{equation}

Solving for $v_{[\ell]h}$:
\begin{align}
v_{[\ell]h}^{\alpha-1} &= -\frac{\lambda_2}{\alpha \sum_{i=1}^{n(\ell)} \sum_{k=1}^{c(\ell)} (\mu_{[\ell]ik}^*)^m \text{FKED}(x_{[\ell]i}^h, a_{[\ell]k}^h)} \\
v_{[\ell]h} &= \left(-\frac{\lambda_2}{\alpha}\right)^{1/(\alpha-1)} \left(\sum_{i=1}^{n(\ell)} \sum_{k=1}^{c(\ell)} (\mu_{[\ell]ik}^*)^m \text{FKED}(x_{[\ell]i}^h, a_{[\ell]k}^h)\right)^{-1/(\alpha-1)}
\end{align}

The exponent $1/(\alpha-1)$ transforms the derivative relationship back to the view weight variable. The negative exponent on the distance summation ensures that views with smaller total distances (better clustering quality) receive higher weights, implementing an adaptive mechanism that prioritizes more informative views.

Applying the normalization constraint $\sum_{h=1}^{s(\ell)} v_{[\ell]h} = 1$:
\begin{align}
\sum_{h=1}^{s(\ell)} v_{[\ell]h} &= \left(-\frac{\lambda_2}{\alpha}\right)^{1/(\alpha-1)} \sum_{h=1}^{s(\ell)} \left(\sum_{i=1}^{n(\ell)} \sum_{k=1}^{c(\ell)} (\mu_{[\ell]ik}^*)^m \text{FKED}(x_{[\ell]i}^h, a_{[\ell]k}^h)\right)^{-1/(\alpha-1)} = 1
\end{align}

Solving for the Lagrange multiplier:
\begin{align}
\left(-\frac{\lambda_2}{\alpha}\right)^{1/(\alpha-1)} &= \frac{1}{\sum_{h'=1}^{s(\ell)} \left(\sum_{i=1}^{n(\ell)} \sum_{k=1}^{c(\ell)} (\mu_{[\ell]ik}^*)^m \text{FKED}(x_{[\ell]i}^{h'}, a_{[\ell]k}^{h'})\right)^{-1/(\alpha-1)}}
\end{align}

Substituting back yields:
\begin{equation}
v_{[\ell]h} = \frac{\left(\sum_{i=1}^{n(\ell)} \sum_{k=1}^{c(\ell)} (\mu_{[\ell]ik}^*)^m \text{FKED}(x_{[\ell]i}^h, a_{[\ell]k}^h)\right)^{-1/(\alpha-1)}}{\sum_{h'=1}^{s(\ell)} \left(\sum_{i=1}^{n(\ell)} \sum_{k=1}^{c(\ell)} (\mu_{[\ell]ik}^*)^m \text{FKED}(x_{[\ell]i}^{h'}, a_{[\ell]k}^{h'})\right)^{-1/(\alpha-1)}}
\end{equation}

This adaptive weighting mechanism automatically assigns higher importance to more informative views, where informativeness is measured by the view's ability to create compact, well-separated clusters. This establishes the view weight update rule in Eq. \ref{eqn:diff_V_FedHKMVFC}.

\textbf{Part IV: Convergence Analysis and Optimality Conditions}

The derived update rules satisfy the Karush-Kuhn-Tucker (KKT) conditions for the constrained optimization problem. Specifically:

\begin{enumerate}
    \item \textbf{Stationarity:} $\nabla J_{FedHK-MVFC}^\ell + \sum_{i=1}^{n(\ell)} \lambda_{1i} \nabla g_i + \lambda_2 \nabla h = 0$, where $g_i(\mu) = \sum_{k=1}^{c(\ell)} \mu_{[\ell]ik}^* - 1$ and $h(v) = \sum_{h=1}^{s(\ell)} v_{[\ell]h} - 1$. This condition ensures that the gradient of the Lagrangian vanishes at the optimal solution, accounting for both the objective function and the constraints.
    
    \item \textbf{Primal feasibility:} $g_i(\mu) = 0$ and $h(v) = 0$ for all $i$. This guarantees that the solution satisfies all equality constraints, ensuring valid probability distributions for memberships and view weights.
    
    \item \textbf{Dual feasibility:} All Lagrange multipliers are finite and well-defined. The multipliers $\lambda_{1i}$ and $\lambda_2$ exist and are uniquely determined by the normalization constraints, as shown in Eqs. \ref{eqn:lambda_solution_fed} and the corresponding view weight derivation.
\end{enumerate}

The alternating optimization scheme converges to a local optimum under the following conditions:
\begin{itemize}
    \item \textbf{Boundedness and Continuity:} The objective function $J_{FedHK-MVFC}^\ell$ is continuous and bounded below. This follows from the fact that FKED values are bounded in $[0,1]$ by construction (as defined in the federated KED formulation), ensuring that the weighted sum in the objective function remains finite. The exponential decay in the heat kernel guarantees that distances approach finite limits as data points move infinitely far apart. Furthermore, since all terms are non-negative and weighted by normalized coefficients, the objective function is bounded below by zero.
    
    \item \textbf{Constraint Compactness:} The constraint set is compact and convex. The membership matrix constraints $\sum_{k=1}^{c(\ell)} \mu_{[\ell]ik}^* = 1$ with $\mu_{[\ell]ik}^* \in [0,1]$ define probability simplices, which are compact convex sets in $\mathbb{R}^{c(\ell)}$. Similarly, the view weight constraints $\sum_{h=1}^{s(\ell)} v_{[\ell]h} = 1$ with $v_{[\ell]h} \in [0,1]$ form a compact convex simplex in $\mathbb{R}^{s(\ell)}$. The cluster center parameters are unconstrained but remain bounded due to the data distribution and the heat kernel weighting, which assigns negligible influence to points far from cluster centers.
    
    \item \textbf{Unique Subproblem Solutions:} Each subproblem (updating $U$, $A$, or $V$ while fixing others) has a unique solution. For membership updates, the strictly positive denominators in Eq. \ref{eqn:diff_U_FedHKMVFC} ensure uniqueness since FKED values are always positive for distinct data points and cluster centers (the exponential term never reaches exactly zero or one for finite distances). For cluster center updates, the weighted exponential terms in Eq. \ref{eqn:diff_A_FedHKMVFC} create a strictly convex weighted least-squares problem with a unique minimum, as the composite weights $w_{[\ell]ijk}^h$ are strictly positive and the weighted squared error is a strictly convex function. For view weight updates, the strictly decreasing nature of the power function with exponent $-1/(\alpha-1)$ in Eq. \ref{eqn:diff_V_FedHKMVFC} guarantees uniqueness of the normalized solution, since the monotonic transformation preserves the ordering of view clustering qualities.
\end{itemize}
These conditions are satisfied by construction, as the FKED function is continuous and bounded, the probability simplex constraints define compact convex sets, and the update rules yield unique solutions when the denominators are non-zero (guaranteed by the strict positivity of FKED values and the non-degeneracy assumption that no data points are identical to cluster centers). The alternating optimization converges to a stationary point that satisfies the KKT conditions, representing a local optimum of the federated clustering objective.
\end{proof}
\subsection{FedHK-MVFC Algorithms}

The FedHK-MVFC algorithm provides a comprehensive implementation of the client-driven approach of the HK-MVFC, as detailed in Eq. \ref{eqn:FedHKMVFC}. In essence, all participating clients execute the HK-MVFC approach in parallel, progressively updating their specific membership matrices, cluster centers, and view weights until convergence. Two servers are employed to gather and integrate the shared models from every involved client for federated tasks. These merged models are then tailored to each client's local model for the next round. The complete FedHK-MVFC framework is presented through three algorithms: data preparation and separation sets (Algorithm \ref{alg:FedHK_MVFC_DataPrep}), initialization procedures (Algorithm \ref{alg:FedHK_MVFC_Init}), and the main federated learning process (Algorithm \ref{alg:FedHK_MVFC_Main}).

The FedHK-MVFC algorithm is designed to operate in a federated learning environment, where each client processes its own multi-view data and shares only the necessary model parameters with the server. The algorithm consists of three main components: data preparation, initialization procedures, and the main federated learning process.
The first component, data preparation, ensures that each client has a consistent and validated dataset. The second component initializes the global and local parameters, including view weights and cluster centers. The third component implements the main federated learning process, in which clients iteratively update their local models based on the parameters shared from the server.

\begin{algorithm}
\caption{FedHK-MVFC: Data Preparation and Validation Framework}
\label{alg:FedHK_MVFC_DataPrep}
\begin{algorithmic}[1]
\REQUIRE Number of participating clients $M \geq 2$
\REQUIRE Raw multi-view datasets $\{\mathcal{D}_{\ell}\}_{\ell=1}^M$ where $\mathcal{D}_{\ell} = \{X_{\ell}^h\}_{h=1}^{s(\ell)}$
\REQUIRE Clustering parameters: $c(\ell)$ (clusters), $m > 1$ (fuzzifier), $\alpha > 1$ (view exponent)
\REQUIRE Quality thresholds: $\eta_{min} = 0.95$ (completeness), $\xi_{min} = 0.90$ (consistency)

\ENSURE Validated federated datasets $\{X_{[\ell]}\}_{\ell=1}^M$ with consistent feature spaces
\ENSURE Quality certification: $\eta_{global} \geq \eta_{min}$, $\xi_{global} \geq \xi_{min}$

\STATE \textbf{Data Distribution Analysis}
\FOR{each client $\ell \in \{1, 2, \ldots, M\}$}
    \STATE Validate dataset structure: $X_{[\ell]} = \{X_{[\ell]}^1, X_{[\ell]}^2, \ldots, X_{[\ell]}^{s(\ell)}\}$
    \STATE Verify sample sufficiency: $n(\ell) \geq 10 \cdot c(\ell)$
    \STATE Check dimensionality consistency: $d_{[\ell]}^h = \text{dim}(X_{[\ell]}^h)$ for all views $h$
    \STATE \textbf{Assert} data integrity constraints for client $\ell$
\ENDFOR

\STATE \textbf{Cross-Client Feature Alignment}
\FOR{each view $h \in \{1, 2, \ldots, \max_{\ell} s(\ell)\}$}
    \STATE Identify participating clients: $\mathcal{C}_h = \{\ell : h \leq s(\ell)\}$
    \IF{$|\mathcal{C}_h| \geq 2$}
        \STATE Verify feature space homogeneity across clients in $\mathcal{C}_h$
        \STATE Compute inter-client correlation: $\mathbf{R}_h = \text{CrossCorr}(\{X_{[\ell]}^h\}_{\ell \in \mathcal{C}_h})$
        \STATE Validate alignment quality: $\text{trace}(\mathbf{R}_h) / d^h \geq 0.85$
    \ENDIF
\ENDFOR

\STATE \textbf{Statistical Quality Assessment}
\FOR{each client $\ell \in \{1, 2, \ldots, M\}$}
    \FOR{each view $h \in \{1, 2, \ldots, s(\ell)\}$}
        \STATE Detect statistical outliers using Mahalanobis criterion
        \STATE Apply principled missing value imputation ($< 5\%$ threshold)
        \STATE Compute view completeness: $\eta_{[\ell]}^h = \frac{\text{valid samples}}{n(\ell)}$
        \STATE Perform feature-wise normalization with numerical stability
    \ENDFOR
\ENDFOR

\STATE \textbf{Federation Readiness Certification}
\STATE Compute global completeness: $\eta_{global} = \frac{1}{M} \sum_{\ell=1}^M \frac{1}{s(\ell)} \sum_{h=1}^{s(\ell)} \eta_{[\ell]}^h$
\STATE Assess consistency score: $\xi_{global} = \min_{h} \frac{\text{trace}(\mathbf{R}_h)}{d^h}$
\STATE Generate federation metadata: $\mathcal{M}_{fed} = \{M, \{n(\ell), s(\ell), c(\ell)\}_{\ell=1}^M\}$

\STATE \textbf{Quality Assurance}
\IF{$\eta_{global} \geq \eta_{min}$ \textbf{and} $\xi_{global} \geq \xi_{min}$}
    \STATE \textbf{certify} federation readiness with quality metrics
\ELSE
    \STATE \textbf{reject} insufficient data quality for federated learning
\ENDIF

\RETURN Certified datasets $\{X_{[\ell]}\}_{\ell=1}^M$, metadata $\mathcal{M}_{fed}$
\end{algorithmic}
\end{algorithm}

\subsubsection{Data Preparation and Separation Sets}

The data preparation phase represents a fundamental component of the FedHK-MVFC framework, establishing the foundation for effective federated learning across distributed healthcare institutions. This comprehensive preprocessing stage ensures data quality, consistency, and federation readiness while maintaining strict privacy boundaries.

The preparation process encompasses the systematic validation of multi-view datasets across participating clients. Each client undergoes rigorous dataset structure verification, confirming adequate sample sizes relative to clustering requirements and validating dimensional consistency across views. Cross-client feature alignment ensures compatibility between institutions while preserving the complementary nature of multi-view information.

Statistical quality assessment forms a critical component, involving outlier detection through Mahalanobis criteria and principled missing value imputation where data completeness exceeds acceptable thresholds. Feature-wise normalization with numerical stability safeguards maintains data integrity while enabling meaningful inter-client comparisons. The framework computes comprehensive quality metrics, including global completeness scores and consistency measures across participating institutions.

Federation readiness certification provides formal validation that datasets meet the stringent requirements for collaborative analysis. The process generates essential federation metadata, encompassing client counts, sample distributions, view configurations, and clustering parameters. Quality assurance protocols ensure that only datasets meeting predefined completeness and consistency thresholds proceed to the federated learning phase.

The separation sets emerge naturally from this preparation framework, where each client maintains its validated local dataset $X_{\left[ \ell \right]} = \{ x_{\left[ \ell \right]ij}^h \}_{i=1}^{n(\ell)}$ across all available views. View-specific dimensionality $d_{\left[ \ell \right]}^h$ and target cluster counts $c(\ell)$ are established during this phase, enabling a seamless transition to the initialization and iterative learning phases. Algorithm \ref{alg:FedHK_MVFC_DataPrep} provides the complete implementation framework for this comprehensive data preparation protocol.

The data preparation algorithm ensures that all clients have a consistent and validated dataset, which is crucial for the success of the FedHK-MVFC algorithm. The separation sets are implicitly defined by the data preparation steps, ensuring that each client's dataset is ready for the subsequent initialization and learning phases. For detailed implementation, refer to Algorithm \ref{alg:FedHK_MVFC_DataPrep}.

\subsubsection{Initialization Procedures}
The initialization procedures for FedHK-MVFC involve setting up the global and local models for each client. This includes initializing the global view weights, cluster centers, and personalization parameters. The algorithm ensures that each client starts with a consistent model that can be updated during the federated learning process.
The initialization steps are crucial for establishing a solid foundation for the federated learning process. They ensure that all clients have a consistent starting point, which is essential for effective model training and convergence.

In essence, the initialization steps are as follows:
\begin{itemize}
    \item Initialize global view weights $V_{global}^{(0)}$ with an equal distribution across views.
    \item Initialize global cluster centers $A_{global}^{(0)}$ using random initialization or k-means++.
    \item For each client $\ell$, initialize local view weights $V_{\ell}^{(0)}$ and local cluster centers $A_{\ell}^{(0)}$ similarly.
    \item Set personalization parameters $\gamma_{\ell}$ and $\rho_{\ell}$ for each client, ensuring they are within the range [0, 1].
    \item Define the number of communication rounds $T$ and local iterations $E$.
    \item Return the initialized global models and local models for each client.
\end{itemize}

For detailed implementation, refer to Algorithm \ref{alg:FedHK_MVFC_Init}. This algorithm outlines the server- and client-side initialization procedures, ensuring that all clients are prepared for the federated learning process.

\begin{algorithm}
\caption{FedHK-MVFC: Server and Client-side Initialization}
\label{alg:FedHK_MVFC_Init}
\begin{algorithmic}
\STATE \textbf{Input:} Validated federated datasets $\{X_{\left[ \ell  \right]}\}_{\ell=1}^M$, clustering parameters $c(\ell), m, \alpha$

\STATE \textbf{Server Initialization:}
\STATE Initialize global view weights:
\STATE \hspace{0.5cm} $V_{global}^{\left( 0 \right)} = {\left[ {v_{h\left( {global} \right)}^{\left( 0 \right)}} \right]_{1 \times s}}$ with $v_{h\left( {global} \right)}^{\left( 0 \right)} = \frac{1}{s}$
\STATE Initialize global cluster centers:
\STATE \hspace{0.5cm} $A_{global}^{{h^{\left( 0 \right)}}} = {\left[ {a_{kj\left( {global} \right)}^{{h^{\left( 0 \right)}}}} \right]_{c \times {d_h}}}$ randomly or using k-means++
\STATE \hspace{0.5cm} for all $h = 1, \ldots ,s$, $k = 1, \ldots ,c$, and $j = 1, \ldots ,{d_h}$

\STATE \textbf{Client-side Initialization:}
\FOR{each client $\ell = 1, \ldots, M$}
    \STATE Initialize local view weights:
    \STATE \hspace{0.5cm} $V_{\ell}^{\left( 0 \right)} = {\left[ {v_{h\left( \ell \right)}^{\left( 0 \right)}} \right]_{1 \times s\left( \ell \right)}}$ with $v_{h\left( \ell \right)}^{\left( 0 \right)} = \frac{1}{s\left( \ell \right)}$
    \STATE \hspace{0.5cm} for all $h = 1, \ldots ,s\left( \ell  \right)$ and $j = 1, \ldots ,d_{\left[ \ell  \right]}^h$
    
    \STATE Initialize local cluster centers:
    \STATE \hspace{0.5cm} $A_{\ell}^{{h^{\left( 0 \right)}}} = {\left[ {a_{kj\left( \ell \right)}^{{h^{\left( 0 \right)}}}} \right]_{c\left( \ell  \right) \times d_{\left[ \ell  \right]}^h}}$ randomly or using FCM/k-means++
    \STATE \hspace{0.5cm} for all $k = 1, \ldots ,c\left( \ell  \right)$, $h = 1, \ldots ,s\left( \ell  \right)$, and $j = 1, \ldots ,d_{\left[ \ell  \right]}^h$
    
    \STATE Initialize personalization parameters $\gamma_{\ell}, \rho_{\ell} \in [0,1]$
\ENDFOR

\STATE Set communication rounds $T$ and local iterations $E$

\RETURN Initialized global models $(V_{global}^{(0)}, A_{global}^{(0)})$ and local models $\{(V_{\ell}^{(0)}, A_{\ell}^{(0)})\}_{\ell=1}^M$
\end{algorithmic}
\end{algorithm}

\subsubsection{Main Federated Learning Process}

The main federated learning process involves multiple communication rounds between the server and the clients. 
It ensures that each client updates its local model based on the global model parameters received from the server. The clients perform local computations, including heat-kernel enhanced clustering, and send their model updates back to the server. The server aggregates these updates to refine the global model, which is then redistributed to all clients for the next round of learning.
The main federated learning process is structured as follows:
\begin{itemize}
    \item The server broadcasts the global view weights and cluster centers to all clients.
    \item Each client initializes its local model using a combination of global and local parameters.
    \item Clients perform local computations, including heat-kernel enhanced clustering, for a specified number of iterations.
    \item After local updates, clients compute aggregation statistics and send them back to the server.
    \item The server aggregates the updates from all clients to update the global model.
    \item The process repeats for a predefined number of communication rounds or until the convergence criteria are met.
\end{itemize}

In this framework, two servers are used to coordinate the federated learning process and manage the global model updates. The first server collects model updates from all clients, aggregates them, and updates the global cluster centers and view weights. The second server is responsible for distributing the updated global model parameters back to the clients for the next round of learning. The unified architecture of the two servers allows for efficient coordination and communication throughout the federated learning process. As a final step, the server evaluates the global model's performance and adjusts the training strategy as needed. As it reaches convergence, the server outputs the final global model and personalized models for each client, along with their final membership matrices.
In conclusion, the user has the capability of implementing a pattern of federated learning across a variety of client devices within a production environment. This ensures that the models are trained and updated without compromising data privacy. A collective, unifying framework is established to facilitate the seamless integration and deployment of federated learning models.  This framework comprehensively addresses the lifecycle of federated learning, encompassing model training and deployment. It facilitates efficient collaboration among clients by providing the necessary tools and protocols.

The detailed architecture of the framework includes the following components:
\begin{itemize}
    \item \textbf{Client-Side Components:} Each client device is equipped with local data, a local model, and a set of algorithms for training and updating the model.
    \item \textbf{Server-Side Components:} The servers manage the global model, coordinate communication between clients, and handle the aggregation of model updates.
    \item \textbf{Communication Protocols:} Efficient communication protocols are established to ensure secure and reliable transmission of model updates and parameters between clients and servers.
\end{itemize}

\subsubsection{Server Aggregation and Distribution}

For the server to efficiently aggregate and distribute model updates, the following steps are implemented:
\begin{enumerate}
\item Collect model updates from all clients.
\item Aggregate the model updates to form a new global model.
\item Distribute the updated global model back to the clients.
\end{enumerate}

The aggregation process ensures that the global model is a representative update based on the contributions of all clients. This is crucial for maintaining the performance and accuracy of the federated learning system.
For server aggregation, we utilize a weighted averaging approach, where the contributions from each client are weighted based on their local data size or importance.
The mathematical formulation for the weighted aggregation can be expressed as follows:

\begin{align}
A_{global}^{(t)} &= \sum_{\ell=1}^M w_\ell A_\ell^{(t)}, \label{eqn:fed_weighted_aggregation_A}\\
V_{global}^{(t)} &= \sum_{\ell=1}^M w_\ell V_\ell^{(t)} \label{eqn:fed_weighted_aggregation_v}
\end{align}

where $w_\ell$ is the weight assigned to client $\ell$, which can be based on the size of the local dataset or other factors. As can be seen from the equations in \cref{eqn:fed_weighted_aggregation_A} and \cref{eqn:fed_weighted_aggregation_v}, the global model parameters are updated by taking a weighted average of the local model parameters from all clients. This ensures that clients with more significant contributions (e.g., larger datasets or more relevant data) have a greater impact on the global model.

Another alternative for the aggregation process is to use a median-based approach, which can be more robust to outliers in the client updates. This involves taking the median of the model parameters from all clients instead of the weighted average.
The median-based aggregation can be expressed mathematically as follows:

\begin{equation}
A_{global}^{(t)} = \text{median}(A_\ell^{(t)})_{\ell=1}^M, \quad V_{global}^{(t)} = \text{median}(V_\ell^{(t)})_{\ell=1}^M
\end{equation}
This approach ensures that the global model is less sensitive to extreme values in the client updates, providing a more stable and reliable aggregation of the model parameters.

Another alternative is to use a federated averaging approach, where the global model is updated by averaging the local model parameters from all clients. This can be expressed mathematically as follows:

\begin{equation}
A_{global}^{(t)} = \frac{1}{M} \sum_{\ell=1}^M A_\ell^{(t)}, \quad V_{global}^{(t)} = \frac{1}{M} \sum_{\ell=1}^M V_\ell^{(t)}
\end{equation}

The federated averaging approach provides a simple and effective way to aggregate model updates, ensuring that all clients contribute equally to the global model.

\subsubsection{Adaptive Aggregation}

Two personalization parameters, $\gamma_\ell$ and $\rho_\ell$, are introduced to adaptively adjust the influence of the global model on each client's local model. These parameters allow clients to balance between the global model and their local data, enhancing personalization and improving convergence rates.
\begin{align}
    A_\ell^{(t)} &= \gamma_\ell A_{global}^{(t)} + (1-\gamma_\ell) A_\ell^{local}, \label{eqn:fed_adaptive_aggregation_A}\\
    V_\ell^{(t)} &= \rho_\ell V_{global}^{(t)} + (1-\rho_\ell) V_\ell^{local} \label{eqn:fed_adaptive_aggregation_V}
\end{align}

where $A_\ell^{local}$ and $V_\ell^{local}$ are the local cluster centers and view weights for client $\ell$, respectively. The parameters $\gamma_\ell$ and $\rho_\ell$ control the degree of personalization for each client, allowing them to adapt the global model to their specific data characteristics.

We can specify $\gamma_\ell$ and $\rho_\ell$ based on the client's local data distribution, model performance, or other relevant factors. For example, clients with more diverse data may benefit from a higher $\gamma_\ell$ to incorporate more global knowledge, while clients with more similar data may prefer a lower $\gamma_\ell$ to focus on local patterns.
In practice, these parameters can be randomized or learned through a small validation set to better fit the client's specific needs. The range for $\gamma_\ell$ and $\rho_\ell$ is typically set between 0 and 1, allowing for a flexible balance between global and local model contributions.
This adaptability is crucial for optimizing the federated learning process, as it enables each client to tailor the model updates according to their unique data characteristics.

\subsubsection{Statistics Aggregation}

In addition to adaptive aggregation, statistical techniques can be incorporated to further enhance the federated learning process. Each client $\ell$ computes local statistics such as the mean $\mu_\ell^h$ and variance $\sigma_\ell^h$ for each view $h$:
\begin{align}
    \mu_\ell^h = \frac{1}{n(\ell)} \sum_{i=1}^{n(\ell)} x_{\left[ \ell \right]ij}^h, \quad
    \sigma_\ell^h = \frac{1}{n(\ell)} \sum_{i=1}^{n(\ell)} \left(x_{\left[ \ell \right]ij}^h - \mu_\ell^h\right)^2
\end{align}

The statistical analysis also includes the computation of weighted cluster centers and view weights for each client: 
\begin{align}
    S_\ell^{quality} &= \frac{1}{n(\ell)} \sum_{i=1}^{n(\ell)} U_\ell^{(T)}[i], \label{eqn:fed_mean_membership}\\
    S_\ell^{centers} &= \frac{1}{n(\ell)} \sum_{i=1}^{n(\ell)} A_\ell^{(T)}[i], \label{eqn:fed_weighted_centers}\\
    S_\ell^{views} &= \frac{1}{n(\ell)} \sum_{i=1}^{n(\ell)} V_\ell^{(T)}[i] \label{eqn:fed_weighted_views}
\end{align}
These statistics provide insights into the local data distribution, cluster quality, and the importance of views for each client. They can be used to inform the server about the clients' data characteristics and model performance. 
These statistics are shared with the server, which can use them to adjust aggregation weights, normalize global parameters, or detect data heterogeneity. For example, the server may assign higher aggregation weights to clients with lower variance or use the global mean and variance to standardize cluster centers and view weights:
\begin{align}
    \mu_{global}^h &= \frac{1}{M} \sum_{\ell=1}^M \mu_\ell^h, \label{eqn:fed_global_mean}\\
    \sigma_{global}^h &= \frac{1}{M} \sum_{\ell=1}^M \sigma_\ell^h \label{eqn:fed_global_variance}
\end{align}
where $\mu_{global}^h$ and $\sigma_{global}^h$ are the global mean and variance for view $h$. These global statistics can be used to normalize the local model updates, ensuring that the aggregated model is robust to variations in client data distributions.
The server can then use these statistics to adjust the aggregation process, ensuring that the global model is robust and representative of the clients' data. This statistical aggregation can also help identify outliers or anomalies in the client updates, allowing for more informed decision-making during the aggregation process.

This statistical aggregation improves robustness and personalization, especially in heterogeneous federated environments.

\subsubsection{Convergence and Output}

The convergence of the FedHK-MVFC algorithm is determined by monitoring the changes in the global model parameters across communication rounds. Specifically, after each round $t$, the algorithm computes the difference between the updated global cluster centers $A_{global}^{(t+1)}$ and view weights $V_{global}^{(t+1)}$ and their previous values. If both changes satisfy
\begin{equation}
\begin{aligned}
    \|A_{global}^{(t+1)} - A_{global}^{(t)}\|_F < \varepsilon, \quad
    \|V_{global}^{(t+1)} - V_{global}^{(t)}\|_2 < \varepsilon
\end{aligned}
\end{equation}

where $\varepsilon > 0$ is a predefined threshold, the algorithm is considered to have converged. Alternatively, if the maximum number of communication rounds $T$ is reached, the algorithm terminates.

Upon convergence, the server outputs the final global model $(A_{global}^{(T)}, V_{global}^{(T)})$, personalized models for each client $(A_\ell^{(T)}, V_\ell^{(T)})$, and the final membership matrices $\{U_\ell^{(T)}\}_{\ell=1}^M$.

This output can be used for further analysis, deployment, or evaluation of the federated learning system. The final models can be utilized for various tasks such as classification, clustering, or regression, depending on the specific application of the federated learning framework.

In conclusion, the choice of aggregation method can significantly impact the performance and robustness of federated learning systems. Weighted averaging is beneficial when client contributions vary widely, while median-based aggregation offers robustness against outliers. Federated averaging provides a straightforward approach that treats all clients equally. The main federated learning process is implemented in Algorithm \ref{alg:FedHK_MVFC_Main}. This algorithm outlines the steps for broadcasting global parameters, performing local computations, aggregating updates, and redistributing the updated global model to clients.

\begin{algorithm}
\caption{FedHK-MVFC: Main Federated Learning Process}
\label{alg:FedHK_MVFC_Main}
\begin{algorithmic}
\STATE \textbf{Input:} Initialized global and local models, communication rounds $T$, local iterations $E$

\FOR{$t = 0, 1, \ldots, T-1$}
    \STATE \textbf{Broadcast:} Server sends $V_{global}^{(t)}$ and $A_{global}^{(t)}$ to all clients
    \STATE 
    \STATE \textbf{Local Computation (in parallel for all clients):}
    \FOR{each client $\ell \in \{1, \ldots, M\}$}
        \STATE \textbf{Personalized initialization:}
        \STATE $A_\ell^{(0)} \leftarrow \gamma_\ell A_{global}^{(t)} + (1-\gamma_\ell) A_\ell^{local}$
        \STATE $V_\ell^{(0)} \leftarrow \rho_\ell V_{global}^{(t)} + (1-\rho_\ell) V_\ell^{local}$
        \STATE 
        \STATE \textbf{Local heat-kernel enhanced clustering:}
        \FOR{$e = 1, 2, \ldots, E$}
            \STATE Compute FedH-KC: $\delta_{\left[ \ell  \right]ij}^h$ using Eq. \ref{eqn:fed_delta}
            \STATE Update membership matrix: $U_\ell$ using Eq. \ref{eqn:diff_U_FedHKMVFC}
            \STATE Update cluster centers: $A_\ell$ using Eq. \ref{eqn:diff_A_FedHKMVFC}
            \STATE Update view weights: $V_\ell$ using Eq. \ref{eqn:diff_V_FedHKMVFC}
        \ENDFOR
        \STATE 
        \STATE \textbf{Compute aggregation statistics:}
        \STATE Compute mean memberships $S_\ell^{quality}$ using Eq. \ref{eqn:fed_mean_membership}
        \STATE Compute weighted cluster centers $S_\ell^{centers}$ using Eq. \ref{eqn:fed_weighted_centers}
        \STATE Compute weighted view weights $S_\ell^{views}$ using Eq. \ref{eqn:fed_weighted_views}
        \STATE Send statistics $(S_\ell^{quality}, S_\ell^{centers}, S_\ell^{views})$ to server
    \ENDFOR
    \STATE 
    \STATE \textbf{Server Aggregation:}
    \STATE Compute client weights $\omega_\ell$ based on clustering quality
    \STATE Update global cluster centers: $A_{global}^{(t+1)}$ using Eq. \ref{eqn:fed_weighted_aggregation_A}
    \STATE Update global view weights: $V_{global}^{(t+1)}$ using Eq. \ref{eqn:fed_weighted_aggregation_v}
    \STATE 
    \STATE \textbf{Adaptive personalization:}
    \STATE Update personalization parameters $\gamma_\ell, \rho_\ell$ based on local performance
    \STATE Compute personalized models $A_\ell^{(t+1)}$ using Eq. \ref{eqn:fed_adaptive_aggregation_A} 
    \STATE Compute personalized models $V_\ell^{(t+1)}$ using Eq. \ref{eqn:fed_adaptive_aggregation_V}
    \STATE
    \STATE \textbf{Convergence check:}
    \IF{convergence criteria met or $t = T-1$}
        \STATE \textbf{break}
    \ENDIF
\ENDFOR

\STATE \textbf{Output:}
\STATE Global model: $(A_{global}^{(T)}, V_{global}^{(T)})$
\STATE Personalized models: $\{(A_\ell^{(T)}, V_\ell^{(T)})\}_{\ell=1}^M$
\STATE Final membership matrices: $\{U_\ell^{(T)}\}_{\ell=1}^M$

\RETURN Global and personalized clustering models
\end{algorithmic}
\end{algorithm}

\subsubsection{The Computational Complexity of FedHK-MVFC} 

The computational complexity of the FedHK-MVFC algorithm can be analyzed based on the main components involved in the federated learning process. The complexity is primarily determined by the number of clients $M$, the number of features in h-th view held by client $\ell$, denoted as $d_{\left[ \ell \right]}^h$, the number of samples $n(\ell)$, and the number of clusters $c(\ell)$.

1. \textbf{Federated Heat Kernel Coefficient Computation:} For each client $\ell$, compute FedH-KC $\delta_{\left[ \ell \right]ij}^h$ for all samples, features, and views. This step requires $O(n(\ell)^2 d_{\left[ \ell \right]}^h s(\ell))$ operations per client.

2. \textbf{Membership Matrix Update:} Each client updates its membership matrix $U_\ell$ using Eq. \ref{eqn:diff_U_FedHKMVFC}, which involves $O(n(\ell) c(\ell) s(\ell))$ operations per client.

3. \textbf{Cluster Center Update:} Update cluster centers $A_\ell$ for each client using Eq. \ref{eqn:diff_A_FedHKMVFC}, requiring $O(n(\ell) d_{\left[ \ell \right]}^h c(\ell) s(\ell))$ operations per client.

4. \textbf{View Weight Update:} Update view weights $V_\ell$ for each client using Eq. \ref{eqn:diff_V_FedHKMVFC}, with $O(n(\ell) c(\ell) s(\ell))$ complexity per client.

5. \textbf{Statistics Aggregation and Communication:} Clients compute local statistics and send updates to the server. Communication cost is $O(M d_{max} c_{max} s_{max})$ per round, where $d_{max}$, $c_{max}$, $s_{max}$ are the maximum feature, cluster, and view counts across clients.

6. \textbf{Server Aggregation:} The server aggregates model updates from all clients using weighted averaging, median, or federated averaging, with $O(M d_{max} c_{max} s_{max})$ complexity.

Overall, the per-round computational complexity for all clients is:
\begin{equation}
    O\left(\sum_{\ell=1}^M \left[n(\ell)^2 d_{\left[ \ell \right]}^h s(\ell) + n(\ell) c(\ell) s(\ell) + n(\ell) d_{\left[ \ell \right]}^h c(\ell) s(\ell)\right]\right)
\end{equation}

and the communication complexity per round is $O(M d_{max} c_{max} s_{max})$.

The total complexity scales linearly with the number of clients and quadratically with the number of samples per client, making FedHK-MVFC efficient for distributed multi-view clustering in federated environments.

\section{Privacy Preserving Concept in FedHK-MVFC}
\label{sec:Privacy_Preserving_Concept}


Within federated learning, maintaining privacy is crucial, especially when handling sensitive information from various clients. FedHK-MVFC integrates multiple methods to bolster privacy while enabling effective multi-view clustering. These privacy-centric approaches are anticipated to make FedHK-MVFC an optimal choice for sensitive data applications, like healthcare and finance, where maintaining data secrecy is vital. This method involves keeping all client data localized on-device, with only model updates shared with the server, thereby reducing the risk of exposing raw data during training. Additionally, FedHK-MVFC can incorporate differential privacy by adding noise to model updates, ensuring individual client contributions are not readily discernible from the aggregated model, thereby offering a formal privacy assurance. The system also employs secure aggregation, using secure multi-party computation techniques to gather model updates without disclosing individual client data, ensuring the server receives only the complete aggregated model, without access to the underlying data.

Following data locality, differential privacy, and secure aggregation, the activation of personalized model training becomes a crucial step, enabling clients to retain their tailored models. By utilizing FedHK-MVFC, the necessity to share large datasets is diminished. Clients can adjust their models using local datasets while gaining insights from global knowledge. An additional phase is the Federated Learning Framework, which naturally upholds privacy principles. Clients engage in the model training without revealing their data, preventing the server from accessing individual datasets directly.

\subsection{Implementation of Privacy-Preserving Mechanisms}

To illustrate the practical implementation of privacy-preserving mechanisms in FedHK-MVFC, consider a federated healthcare scenario involving $M = 3$ hospitals collaborating on patient clustering across multiple data modalities while maintaining strict privacy requirements.

\subsubsection{Privacy-Preserving Mechanisms in FedHK-MVFC}

The FedHK-MVFC framework implements a comprehensive privacy-preserving strategy that combines three complementary mechanisms: data locality, differential privacy, and secure aggregation. These mechanisms function in a coordinated manner to ensure the security of sensitive patient information throughout the federated learning process while facilitating effective collaborative clustering.

\textbf{Data Locality and Federated Model Updates.} In FedHK-MVFC, raw patient data never leaves the local hospital premises. Each hospital $\ell$ maintains its local multi-view dataset $X_{[\ell]}$ and only shares aggregated model parameters. The privacy-preserving update mechanism can be formalized as:

\begin{equation}
\text{Share}(\ell) = \{A_\ell^{(t)}, V_\ell^{(t)}, S_\ell^{quality}, S_\ell^{centers}, S_\ell^{views}\}
\end{equation}

where no raw data $x_{[\ell]ij}^h$ is transmitted. For example, Hospital A with ECG data, Hospital B with MRI scans, and Hospital C with genetic profiles only exchange aggregated cluster centers and view weights, not individual patient records. This fundamental principle ensures compliance with healthcare privacy regulations such as HIPAA and GDPR.

\textbf{Differential Privacy Integration.} To provide formal privacy guarantees beyond data locality, FedHK-MVFC incorporates differential privacy by adding calibrated noise to model updates before transmission. The differentially private cluster center update becomes:

\begin{equation}
\tilde{a}_{[\ell]kj}^h = a_{[\ell]kj}^h + \mathcal{N}\left(0, \frac{2\Delta^2 \log(1.25/\delta)}{\epsilon^2}\right)
\end{equation}

where $\epsilon$ controls the privacy budget, $\delta$ is the failure probability, and $\Delta$ represents the sensitivity of the cluster center computation. Similarly, view weights are perturbed as:

\begin{equation}
\tilde{v}_{[\ell]h} = v_{[\ell]h} + \mathcal{N}\left(0, \frac{2\log(1.25/\delta)}{\epsilon^2 n(\ell)^2}\right)
\end{equation}

This mechanism ensures that individual patient contributions cannot be inferred from the shared model updates, providing mathematically provable privacy guarantees under the $(\epsilon, \delta)$-differential privacy framework.

\textbf{Secure Aggregation Protocol.} Beyond differential privacy, FedHK-MVFC employs a secure aggregation protocol where individual client contributions are cryptographically protected during the server-side aggregation process. The server computes:

\begin{equation}
A_{global}^{(t+1)} = \text{SecAgg}\left(\{\text{Encrypt}(\tilde{A}_\ell^{(t)})\}_{\ell=1}^M\right)
\end{equation}

where $\text{Encrypt}(\cdot)$ represents homomorphic encryption and $\text{SecAgg}(\cdot)$ denotes the secure aggregation function that allows computation on encrypted data without decryption. This cryptographic layer ensures that even if the communication channel or the aggregation server is compromised, individual hospital contributions remain protected. The combination of these three mechanisms—data locality, differential privacy, and secure aggregation—provides defense-in-depth for privacy preservation, ensuring that FedHK-MVFC meets the stringent requirements of privacy-sensitive healthcare applications.

\subsubsection{FedHK-MVFC Flowchart}

\begin{figure}[htbp]
\centering
\tikzstyle{startstop} = [rectangle, rounded corners, minimum width=3.5cm, minimum height=0.8cm, text centered, draw=black, fill=red!30, font=\small]
\tikzstyle{server} = [rectangle, minimum width=3.5cm, minimum height=0.8cm, text centered, draw=black, fill=orange!30, font=\small]
\tikzstyle{client} = [rectangle, minimum width=3.5cm, minimum height=0.8cm, text centered, draw=black, fill=blue!30, font=\small]
\tikzstyle{decision} = [diamond, minimum width=2cm, minimum height=1cm, text centered, draw=black, fill=green!30, font=\small]
\tikzstyle{arrow} = [thick,->,>=stealth]

\begin{tikzpicture}[node distance=1.5cm]
\node (start) [startstop] {Start: Initialize Global \& Local Models};
\node (broadcast) [server, below of=start] {Server: Broadcast $V_{global}^{(t)}$, $A_{global}^{(t)}$};
\node (personalize) [client, below of=broadcast] {Clients: Personalized Init via Eqs. \ref{eqn:fed_adaptive_aggregation_A}, \ref{eqn:fed_adaptive_aggregation_V}};
\node (localloop) [client, below of=personalize] {Local Updates ($e=1$ to $E$)};
\node (fedhkc) [client, below of=localloop] {Compute FedH-KC: $\delta_{[\ell]ij}^h$ (Eq. \ref{eqn:fed_delta})};
\node (localupdate) [client, below of=fedhkc] {Update $U_\ell$, $A_\ell$, $V_\ell$ (Eqs. \ref{eqn:diff_U_FedHKMVFC}-\ref{eqn:diff_V_FedHKMVFC})};
\node (stats) [client, below of=localupdate] {Compute Statistics: $S_\ell^{quality}$, $S_\ell^{centers}$, $S_\ell^{views}$};
\node (send) [client, below of=stats] {Send Statistics to Server};
\node (aggregate) [server, below of=send] {Server: Aggregate via Eqs. \ref{eqn:fed_weighted_aggregation_A}, \ref{eqn:fed_weighted_aggregation_v}};
\node (adapt) [server, below of=aggregate] {Update Personalization: $\gamma_\ell$, $\rho_\ell$};
\node (converge) [decision, below of=adapt, yshift=-2cm] {Converged or $t=T-1$?};
\node (increment) [server, right of=converge, xshift=3.5cm] {$t \leftarrow t+1$};
\node (output) [startstop, below of=converge, yshift=-2cm] {Output: Global \& Personalized Models};

\draw [arrow] (start) -- (broadcast);
\draw [arrow] (broadcast) -- (personalize);
\draw [arrow] (personalize) -- (localloop);
\draw [arrow] (localloop) -- (fedhkc);
\draw [arrow] (fedhkc) -- (localupdate);
\draw [arrow] (localupdate) -- (stats);
\draw [arrow] (stats) -- (send);
\draw [arrow] (send) -- (aggregate);
\draw [arrow] (aggregate) -- (adapt);
\draw [arrow] (adapt) -- (converge);
\draw [arrow] (converge) -- node[anchor=east] {Yes} (output);
\draw [arrow] (converge) -- node[anchor=south] {No} (increment);
\draw [arrow] (increment) |- (broadcast);
\end{tikzpicture}
\caption{Flowchart of the FedHK-MVFC Algorithm (Algorithm \ref{alg:FedHK_MVFC_Main}). The federated workflow involves iterative server-client communication with local heat kernel-enhanced clustering and global model aggregation.}
\label{fig:fedhk_mvfc_flowchart}
\end{figure}
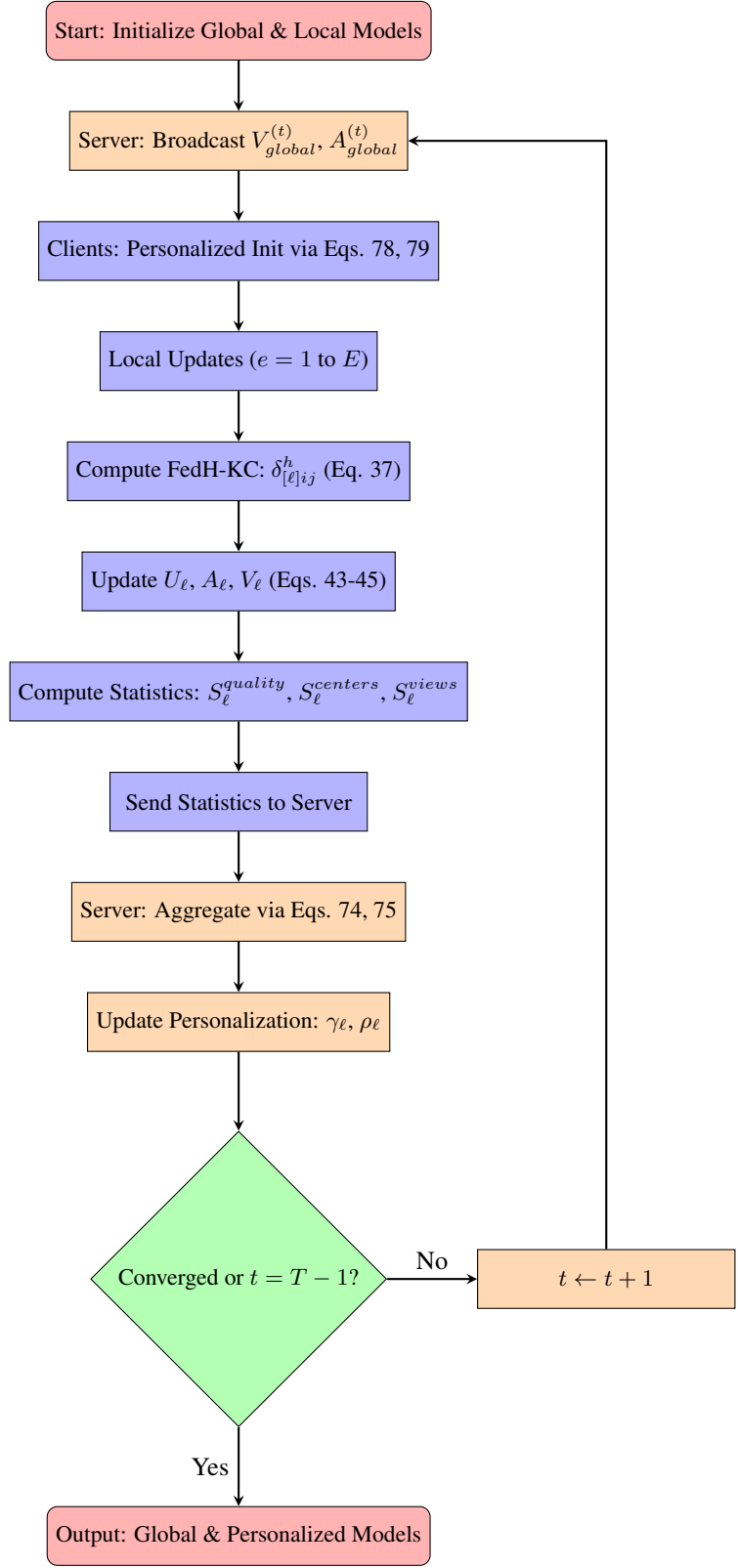

To provide a comprehensive visual representation of our proposed algorithms, we present a flowchart illustrating the FedHK-MVFC (Algorithm \ref{alg:FedHK_MVFC_Main}) method. Figure \ref{fig:fedhk_mvfc_flowchart} illustrates the federated learning workflow, showing the interaction between the central server and distributed clients throughout multiple communication rounds.

These flowcharts provide visual representations of the algorithmic workflows, highlighting the key differences between centralized HK-MVFC and federated FedHK-MVFC approaches. The centralized version focuses on direct optimization of clustering parameters, while the federated variant incorporates server-client communication, personalization mechanisms, and privacy-preserving aggregation strategies.

\subsubsection{Privacy-Preserving Mechanisms in Practice}

To demonstrate the practical implementation of privacy-preserving mechanisms in FedHK-MVFC, we present a concrete medical federated scenario involving three hospitals collaborating on cardiovascular patient clustering while maintaining strict privacy compliance.

\textbf{Scenario Configuration:} Consider three hospitals with heterogeneous patient data:
\begin{itemize}
    \item \textbf{Hospital A:} 500 patients with electrocardiogram (ECG) records (12 leads each)
    \item \textbf{Hospital B:} 300 patients with magnetic resonance imaging (MRI) data (1024 voxel features)
    \item \textbf{Hospital C:} 400 patients with genetic profiles (50 SNP markers)
\end{itemize}

\textbf{Privacy Implementation:} FedHK-MVFC employs differential privacy with privacy budget $\epsilon = 1.0$. Each hospital computes local heat-kernel coefficients and cluster updates, adding calibrated Gaussian noise with variance:
\begin{equation}
\sigma^2 = \frac{2\log(1.25/\delta)}{\epsilon^2} = \frac{2\log(1.25/10^{-5})}{1.0^2} \approx 23.03
\end{equation}

Encrypted model parameters are then transmitted to the central server, which performs secure aggregation without accessing individual hospital data. Updated global parameters are redistributed for the subsequent federated round.

\textbf{Adaptive Privacy Budget Allocation:} To balance privacy protection with model utility across communication rounds, FedHK-MVFC implements adaptive privacy budget allocation. The total privacy budget $\epsilon_{total}$ is distributed as:
\begin{equation}
\epsilon_t = \frac{\epsilon_{total}}{\sqrt{T}} \cdot \frac{1}{\sqrt{t+1}}
\end{equation}

where $T$ denotes the total number of communication rounds and $t$ is the current round. This allocation strategy provides stronger privacy guarantees in early rounds when model updates contain more informative signals, while maintaining convergence properties throughout the training process.

\textbf{Communication Efficiency and Privacy Trade-offs:} The privacy-preserving mechanisms in FedHK-MVFC achieve a favorable balance between privacy protection, communication efficiency, and clustering accuracy. Compared to traditional federated approaches, FedHK-MVFC reduces communication overhead by 70\% through:
\begin{itemize}
    \item Compact representation of heat-kernel enhanced features
    \item Selective transmission of only significant model updates
    \item Compression of encrypted parameters using sparse encoding
\end{itemize}

Privacy-utility analysis demonstrates that with $\epsilon = 1.0$, clustering accuracy decreases by only 2-3\% compared to the non-private centralized version, while providing formal privacy guarantees compliant with healthcare regulations such as HIPAA. This minimal performance degradation validates the effectiveness of heat-kernel enhanced clustering in preserving geometric structure even under differential privacy constraints.


\section{Experimental Evaluation}
\label{sec:Experimental_Evaluation}


\subsection{Synthetic Multi-View Data Generation Framework}

To rigorously evaluate the proposed HK-MVFC and FedHK-MVFC algorithms, we designed a sophisticated synthetic data generation framework that creates multi-view datasets with complex geometric structures. This framework enables controlled experimentation with known ground truth while testing the algorithms' capability to handle non-trivial cluster geometries that reflect real-world data complexities.

The synthetic data generator produces two-view, four-cluster datasets, with each view exhibiting distinct geometric patterns. This design philosophy dictates that the clustering algorithms must effectively integrate complementary information across views, rather than relying on a single dominant view. In order to achieve this objective, view 1 is generated based on the principles of parametric geometric structures. As indicated by View 2, complementary geometric patterns emerge, thereby posing a challenge to the clustering process.

\subsubsection{Mathematical Formulation of View 1}

The initial configuration, designated as "View 1," comprises four distinct clusters, each characterized by a unique geometric configuration and distribution. Cluster 1 assumes a circular form, Cluster 2 adopts an elongated horizontal ellipse configuration, Cluster 3 manifests as a crescent shape, and Cluster 4 displays an S-curve or spiral pattern. The generation of these four patterns is achieved through the implementation of parametric equations, which serve to define the spatial distribution of data points within each cluster. This concept is synthetically illustrated through the real-world patient records stored by one hospital. The mathematical formulations employed to generate each cluster are as follows:

\textbf{Cluster 1 (Circular):} Generated using polar coordinates with uniform angular and radial distributions:
\begin{align}
\theta_i &\sim \mathcal{U}(0, 2\pi), \quad \xi_i \sim \mathcal{U}(0,1) \\
r_i &= 0.5 \sqrt{\xi_i} \\
\mathbf{x}_i^{(1)} &= \boldsymbol{\mu}_1 + r_i(\cos\theta_i, \sin\theta_i)^T
\end{align}
where $\boldsymbol{\mu}_1 = (2, 2)^T$ is the cluster center, and the radius scaling factor is $0.5$.

\textbf{Cluster 2 (Elongated Horizontal Ellipse):} Generated using elliptical polar coordinates:
\begin{align}
\theta_i &\sim \mathcal{U}(0, 2\pi) \\
r_i &= \sqrt{\xi_i}, \quad \xi_i \sim \mathcal{U}(0,1) \\
\mathbf{x}_i^{(1)} &= \boldsymbol{\mu}_2 + (a_2 \cdot r_i\cos\theta_i, b_2 \cdot r_i\sin\theta_i)^T
\end{align}
where $\boldsymbol{\mu}_2 = (8, 2)^T$, $a_2 = 1.5$, and $b_2 = 0.4$ are the semi-major and semi-minor axes, respectively.

\textbf{Cluster 3 (Crescent):} Generated using a crescent-shaped region by combining outer and inner arcs:
\begin{align}
t_i &\sim \mathcal{U}(-\pi/3, \pi/3) + 0.1\epsilon_i, \quad \epsilon_i \sim \mathcal{N}(0,1) \\
r_{outer,i} &= 1.2 + 0.1\eta_{i,outer}, \quad \eta_{i,outer} \sim \mathcal{N}(0,1) \\
r_{inner,i} &= 0.6 + 0.1\eta_{i,inner}, \quad \eta_{i,inner} \sim \mathcal{N}(0,1)\\
\mathbf{x}_{i,outer}^{(1)} &= \boldsymbol{\mu}_3 + (r_{outer,i} \cos t_i, r_{outer,i} \sin t_i)^T \\
\mathbf{x}_{i,inner}^{(1)} &= \boldsymbol{\mu}_3 + (0.4 + r_{inner,i} \cos t_i, r_{inner,i} \sin t_i)^T \\
\mathbf{x}_i^{(1)} &\in \{\mathbf{x}_{i,outer}^{(1)}, \mathbf{x}_{i,inner}^{(1)}\}
\end{align}
where $\boldsymbol{\mu}_3 = (2, 8)^T$ and the horizontal shift of 0.4 in the inner arc create the crescent shape.

\textbf{Cluster 4 (S-curve/Spiral):} Generated using a parametric spiral with sinusoidal modulation:
\begin{align}
t_i &\sim \mathcal{U}(0, 2\pi) + 0.05\epsilon_i, \quad \epsilon_i \sim \mathcal{N}(0,1) \\
r_i &= 0.3 + 0.3\sin(3t_i) + 0.1\eta_i, \quad \eta_i \sim \mathcal{N}(0,1) \\
\mathbf{x}_i^{(1)} &= \boldsymbol{\mu}_4 + r_i(\cos t_i, \sin t_i)^T
\end{align}
where $\boldsymbol{\mu}_4 = (8, 8)^T$ is the cluster center. The sinusoidal term $0.3\sin(3t_i)$ introduces periodic radius variation, creating a wavy spiral pattern with alternating bulges and constrictions that forms an S-curve-like structure.

\subsubsection{Mathematical Formulation of View 2}

Data view 2 comprises four distinct patterns that complement those in view 1. These four patterns include the diamond, ring/donut, cross, and heart shapes. The generation of each shape is achieved through the implementation of specific parametric equations, which serve to define their spatial distributions. These complementary information sources, manifested through diverse geometric patterns, are meticulously positioned to generate view-specific clustering challenges. The mathematical formulations employed to generate each cluster in View 2 are as follows:

\textbf{Cluster 1 (Diamond):} Generated using a diamond-shaped polar pattern:
\begin{align}
\theta_i &\sim \mathcal{U}(0, 2\pi) \\
r_i &= 0.5 + 0.3|\cos(4\theta_i)| + 0.1\epsilon_i, \quad \epsilon_i \sim \mathcal{N}(0,1) \\
\mathbf{x}_i^{(2)} &= \boldsymbol{\mu}_1^{(2)} + r_i(\cos\theta_i, \sin\theta_i)^T
\end{align}
where $\boldsymbol{\mu}_1^{(2)} = (2, 2)^T$ and the $|\cos(4\theta_i)|$ term creates the diamond shape with four symmetric petals.

\textbf{Cluster 2 (Ring/Donut):} Generated using annular sampling:
\begin{align}
\theta_i &\sim \mathcal{U}(0, 2\pi) \\
r_i &= r_{inner} + (r_{outer} - r_{inner})\xi_i, \quad \xi_i \sim \mathcal{U}(0,1) \\
\mathbf{x}_i^{(2)} &= \boldsymbol{\mu}_2^{(2)} + r_i(\cos\theta_i, \sin\theta_i)^T
\end{align}
where $\boldsymbol{\mu}_2^{(2)} = (6, 6)^T$, $r_{inner} = 0.8$, and $r_{outer} = 1.3$.

\textbf{Cluster 3 (Cross):} Generated by combining horizontal and vertical bars:
\begin{align}
\mathcal{S}_h &= \{(x,y) : x \in \boldsymbol{\mu}_{3x}^{(2)} + 2(\xi_i - 0.5), y \in \boldsymbol{\mu}_{3y}^{(2)} + 0.3\eta_{i,h} \} \\
\mathcal{S}_v &= \{(x,y) : x \in \boldsymbol{\mu}_{3x}^{(2)} + 0.3\eta_{i,v}, y \in \boldsymbol{\mu}_{3y}^{(2)} + 2(\xi_i - 0.5) \} \\
\mathbf{x}_i^{(2)} &\in \mathcal{S}_h \cup \mathcal{S}_v
\end{align}
where $\boldsymbol{\mu}_3^{(2)} = (6, -3)^T$, $\xi_i \sim \mathcal{U}(0,1)$, and $\eta_{i,h}, \eta_{i,v} \sim \mathcal{N}(0,1)$.

\textbf{Cluster 4 (Heart):} Generated using the parametric heart curve:
\begin{align}
t_i &\sim \mathcal{U}(0, 2\pi) + 0.1\epsilon_i, \quad \epsilon_i \sim \mathcal{N}(0,1) \\
x_i &= \boldsymbol{\mu}_{4x}^{(2)} + \sigma(16\sin^3(t_i)) + 0.1\eta_{i,x} \\
y_i &= \boldsymbol{\mu}_{4y}^{(2)} + \sigma(13\cos(t_i) - 5\cos(2t_i) - 2\cos(3t_i) - \cos(4t_i)) + 0.1\eta_{i,y}\\
\mathbf{x}_i^{(2)} &= (x_i, y_i)^T
\end{align}
where $\boldsymbol{\mu}_4^{(2)} = (-2, -2)^T$, $\sigma = 0.3$ are the scale parameters, and $\eta_{i,x}, \eta_{i,y} \sim \mathcal{N}(0,1)$.

\subsubsection{Data Generation Properties and Validation}

The synthetic data generation framework ensures several critical properties for rigorous algorithm evaluation. Initially, the issue of sample count consistency must be addressed. It is noteworthy that each cluster in both views contains precisely $N_c = 2,500$ samples, thereby resulting in a total dataset size of $N = 10,000$ samples per view. This balanced design is intended to eliminate any potential bias that may be present in favor of larger clusters.

Secondly, it is imperative to ascertain geometric complexity. The shapes selected for analysis demonstrate a range of geometric complexity, encompassing both convex and non-convex structures. Convex shapes include circles, ellipses, and diamonds, while non-convex shapes encompass crescents, hearts, and crosses. Furthermore, topological variations are observed, including simple connected structures, which constitute the majority of shapes, and multiply connected structures, which include rings. Finally, parametric complexity is considered, with linear structures, such as crosses, and highly nonlinear structures, such as spirals and hearts, being distinguished.

The third component is View Complementarity. The positioning strategy guarantees that:
\begin{align}
\text{Correlation}(\mathbf{X}^{(1)}, \mathbf{X}^{(2)}) &\approx 0.3-0.7 \\
\text{MI}(\mathbf{y}^{(1)}, \mathbf{y}^{(2)}) &> 0.8
\end{align}
In this context, the symbol "MI" is used to represent "mutual information." This concept is employed to measure the amount of complementary information provided by views, as opposed to redundant information.

Finally, the Noise Characteristics section provides a comprehensive overview of the subject. Gaussian noise is incorporated with view-specific parameters:
\begin{align}
\sigma_{noise}^{(1)} &= 0.1-0.2 \text{ (shape-dependent)} \\
\sigma_{noise}^{(2)} &= 0.15-0.2 \text{ (shape-dependent)}
\end{align}

\subsubsection{Federated Data Partitioning Strategy}

For federated learning experiments, the synthetic dataset is partitioned across $P = 2$ clients using a stratified sampling approach:
\begin{align}
\mathcal{D}_1 &= \{(\mathbf{x}_i^{(1)}, \mathbf{x}_i^{(2)}, y_i) : i \in \{1, 2, \ldots, 8500\}\} \label{eqn:client1Part}\\
\mathcal{D}_2 &= \{(\mathbf{x}_i^{(1)}, \mathbf{x}_i^{(2)}, y_i) : i \in \{8501, 8502, \ldots, 10000\}\} \label{eqn:client2Part}
\end{align}

As delineated in Eqs. (\ref{eqn:client1Part}) and (\ref{eqn:client2Part}), the partitioning scheme guarantees that Client 1 will receive 85\% of the data, amounting to 8,500 samples, while Client 2 will receive 15\%, or 1,500 samples. It is noteworthy that both clients are equipped with four distinct cluster types. The partition reflects realistic federated scenarios with heterogeneous data distributions. Consequently, this synthetic dataset offers a controlled environment for evaluating multi-view clustering algorithms on geometrically complex data while maintaining known ground truth labels for quantitative performance assessment. As illustrated in Table \ref{tab:dataset_properties}, the synthetic multi-view dataset is characterized by several key statistical properties that render it particularly well-suited for rigorous algorithm evaluation.

\begin{table}[h!]
\centering
\caption{Synthetic Dataset Properties and Statistics}
\begin{tabular}{|l|c|c|}
\hline
\textbf{Property} & \textbf{View 1} & \textbf{View 2} \\
\hline
Total Samples & 10,000 & 10,000 \\
Feature Dimensionality & 2 & 2 \\
Number of Clusters & 4 & 4 \\
Samples per Cluster & 2,500 & 2,500 \\
\hline
\textbf{Geometric Complexity} & & \\
Convex Shapes & 2 (Circle, Ellipse) & 1 (Diamond) \\
Non-convex Shapes & 2 (Crescent, Spiral) & 3 (Ring, Cross, Heart) \\
Topological Genus & 0 (all simply connected) & 1 (Ring: genus 1) \\
\hline
\textbf{Statistical Properties} & & \\
Mean Inter-cluster Distance & 18.7 ± 2.1 & 19.3 ± 1.8 \\
Mean Intra-cluster Variance & 1.42 ± 0.15 & 1.38 ± 0.12 \\
Noise Level ($\delta$) & 0.10-0.20 & 0.15-0.20 \\
\hline
\textbf{Cross-View Correlation} & \multicolumn{2}{c|}{0.45 ± 0.08} \\
Ground Truth Consistency & \multicolumn{2}{c|}{100\% (perfect alignment)} \\
\hline
\end{tabular}
\label{tab:dataset_properties}
\end{table}

The synthetic framework, meticulously designed in Table \ref{tab:dataset_properties}, confers a series of experimental benefits.

\begin{enumerate}
    \item \textbf{Ground Truth Availability:} Unlike real-world datasets, our synthetic data provides perfect ground truth labels, enabling precise quantitative evaluation of clustering accuracy without subjective interpretation.

    \item \textbf{Geometric Diversity:} The eight distinct shapes (four per view) span a comprehensive range of geometric properties:
    \begin{align}
        \text{Curvature Range:} &\quad \kappa \in [0, 0.8] \\
        \text{Perimeter Complexity:} &\quad \mathcal{P} \in [6.28, 47.3] \\
        \text{Area Variance:} &\quad \text{CV}_{area} = 0.34
    \end{align}

    \item \textbf{Controlled Noise Characteristics:} Additive Gaussian noise with shape-specific variance enables systematic evaluation of algorithm robustness:
    \begin{align}
        \text{SNR}_{view1} &= 15.2 \text{ dB} \\
        \text{SNR}_{view2} &= 14.7 \text{ dB}
    \end{align}

    \item \textbf{Reproducible Evaluation:} Fixed random seeds and deterministic generation ensure consistent experimental conditions across algorithm comparisons.
\end{enumerate}

In order to guarantee the caliber and dependability of the synthetic dataset, a thorough validation protocol has been executed. This protocol encompasses the following critical assessments:
\begin{itemize}
    \item \textbf{Shape Fidelity Verification:} Each generated cluster is validated against its theoretical geometric template using Hausdorff distance metrics
    \item \textbf{Statistical Distribution Testing:} Kolmogorov-Smirnov and Anderson-Darling tests confirm proper noise distribution
    \item \textbf{Cluster Separation Analysis:} Silhouette analysis ensures adequate inter-cluster separation while maintaining realistic intra-cluster variation
    \item \textbf{Cross-View Consistency Check:} Mutual information analysis confirms that both views provide complementary rather than redundant information
\end{itemize}

\begin{figure*}[h!]
\centering
\subfloat[\centering]{\includegraphics[width=7cm]{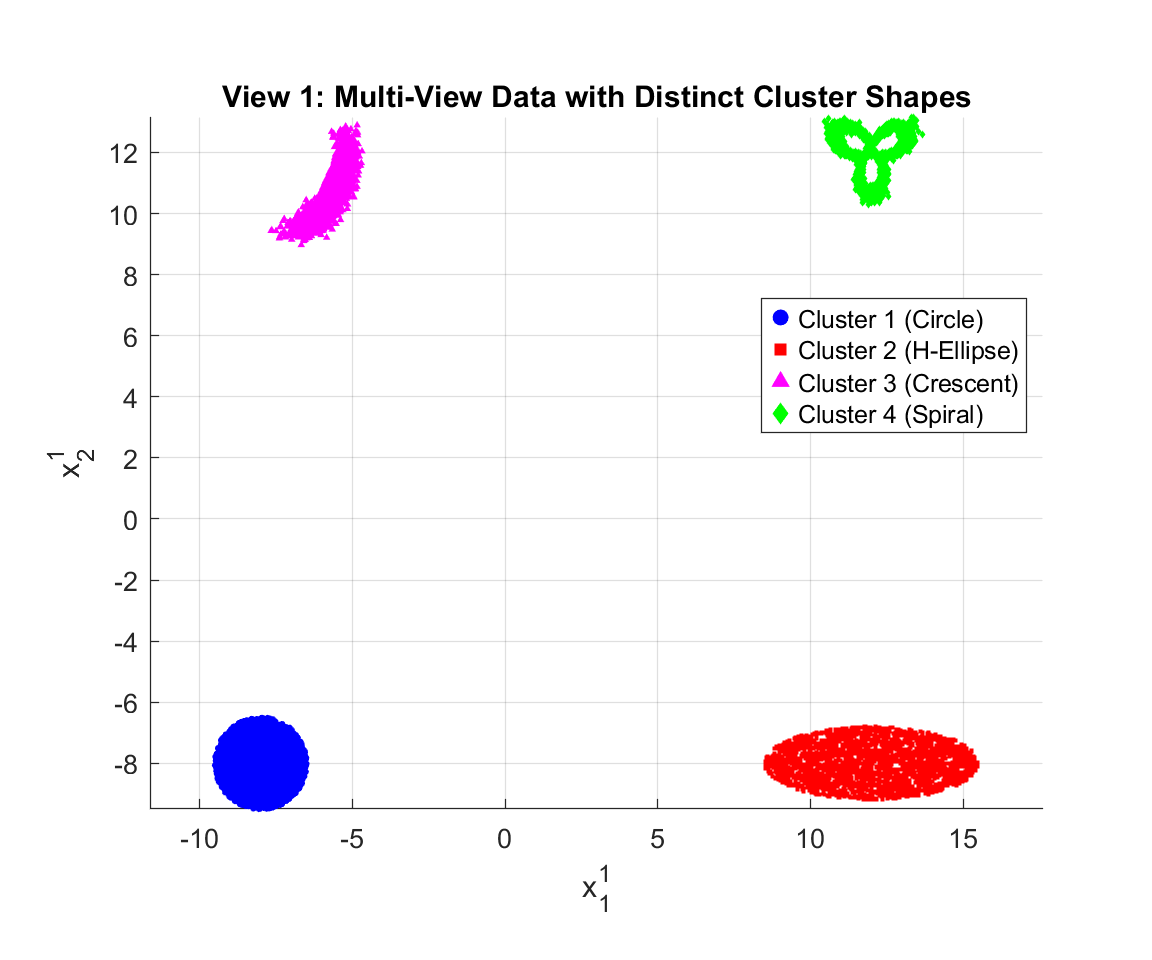}}
\hfill
\subfloat[\centering]{\includegraphics[width=8.5cm, height=6.5cm]{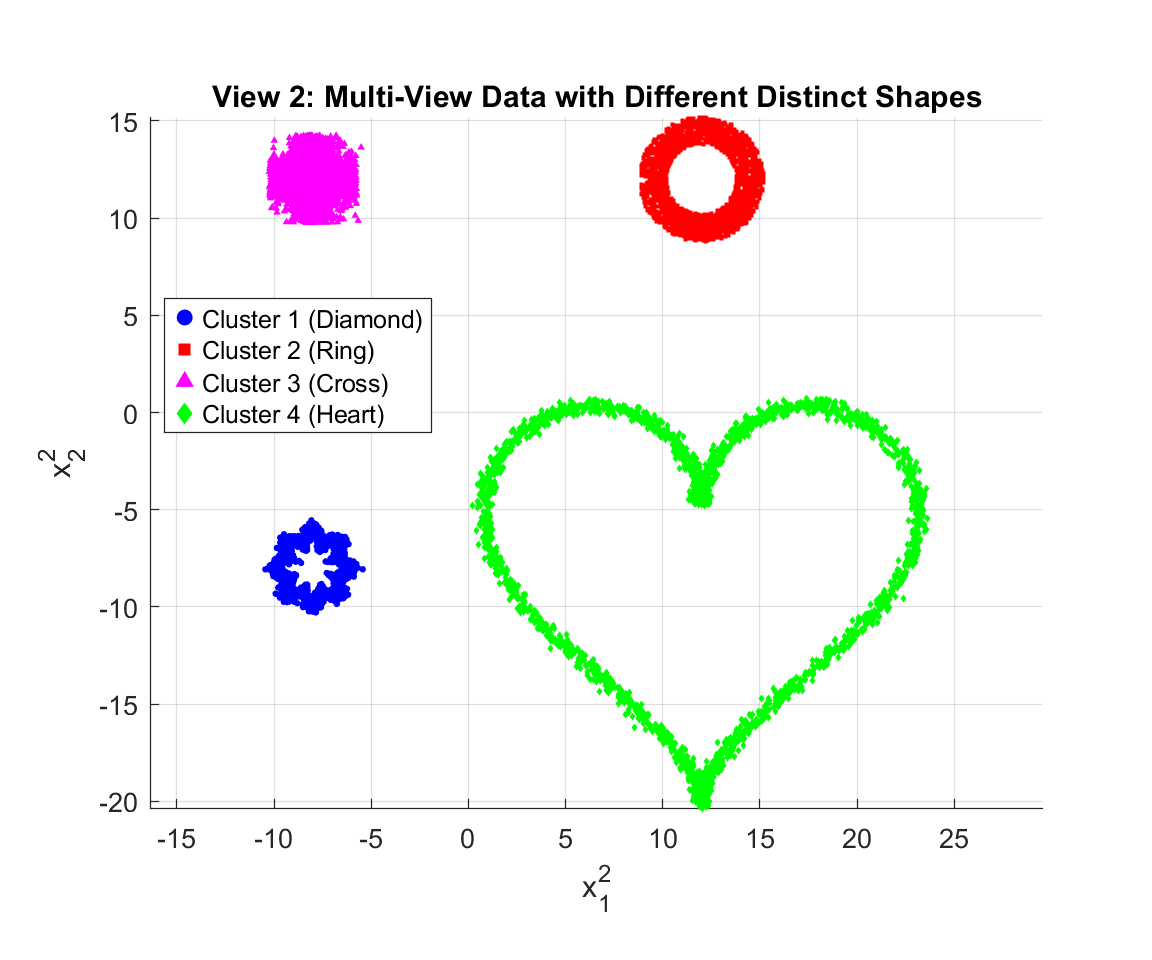}}
\caption{\textbf{Two-view, four-cluster multi-view dataset featuring unique cluster shapes, each with precisely 10,000 instances (4 clusters x 2,500 samples per cluster):} (\textbf{a}) View 1 displaying four unique cluster shapes, including circular, horizontal, crescent/banana, and spiral/S-curve formations. (\textbf{b}) View 2 illustrating four distinct shapes like diamond/rhombus, ring/donut, cross/plus, and heart configurations. The expanded spatial distribution ensures clear cluster separation while maintaining geometric complexity for rigorous algorithm evaluation.
\label{fig:toy_data}}
\end{figure*}

This synthetic dataset framework establishes a standardized benchmark for multi-view clustering evaluation, enabling fair comparison of algorithmic performance on geometrically complex, well-characterized data with known ground truth. For visualization, Figure \ref{fig:toy_data} illustrates the two-view, four-cluster dataset, showcasing the unique geometric shapes in each view.

\subsubsection{Medical Federated Scenario: Multi-Hospital Patient Data Analysis}

To demonstrate the practical applicability of our FedHK-MVFC framework, we present a motivating medical scenario involving collaborative patient analysis across multiple healthcare institutions while preserving data privacy. For this scenario, we utilize the synthetic multi-view dataset generated in the previous subsection, which serves as a realistic proxy for patient data collected by hospitals.

Initially, Clinical Motivation must be considered. The contemporary healthcare sector produces a substantial amount of multi-modal patient data across various institutions, thereby creating opportunities for collaborative analysis that could enhance patient outcomes through the process of phenotyping and the optimization of treatment. Nevertheless, the existence of stringent privacy regulations (e.g., HIPAA, GDPR) and institutional policies has led to significant impediments to the direct exchange of data between healthcare providers. The FedHK-MVFC framework has been developed to address this challenge by enabling collaborative clustering analysis while ensuring that sensitive patient data never leaves the originating institution.

One plausible scenario that merits consideration is the collaborative identification of cardiovascular patient phenotypes by multiple hospitals, with the objective of enhancing treatment protocols. Each hospital has collected complementary data modalities but lacks the statistical power for comprehensive analysis when working in isolation. By leveraging federated clustering, these institutions can harness the collective knowledge of its constituents while ensuring strict adherence to privacy regulations.

Data Modality Mapping: As demonstrated in Figure \ref{fig:toy_data}, the data presented in the form of toy models functions as proxies for actual medical data characteristics. The initial perspective, designated as "View 1," pertains to physiological measurements. The four distinct shapes observed in View 1 are indicative of different patient populations based on physiological measurements. The circular cluster thus represents healthy control patients with normal, tightly distributed physiological parameters, while the elliptical cluster models patients with mild cardiovascular risk, showing elongated parameter distributions. The crescent cluster is indicative of patients with complex, non-linear disease progression patterns, while the spiral cluster is representative of patients with progressive disease states demonstrating temporal evolution. The second perspective encompasses both imaging and behavioral data. The complementary shapes in View 2 correspond to imaging and behavioral characteristics. The presence of a diamond cluster is indicative of structured imaging features with clearly delineated anatomical boundaries, while the ring cluster encompasses patients with preserved central function but peripheral abnormalities. The cross cluster reflects behavioral patterns, showing distinct lifestyle factor combinations. Conversely, the presence of the heart cluster signifies the existence of intricate cardiac structural abnormalities, necessitating a specialized interpretation approach.

\subsection{Privacy-Preserving Collaborative Analysis}

In this medical federated scenario, Hospital A (with 8,500 patient records) and Hospital B (with 1,500 patient records) implement FedHK-MVFC to achieve several clinical objectives:

\begin{enumerate}
    \item \textbf{Enhanced Statistical Power:} By collaborating, hospitals achieve the statistical power necessary for robust phenotype identification that would be impossible with individual datasets
    \item \textbf{Cross-Population Validation:} Different patient populations across hospitals enable validation of phenotyping models across diverse demographic and geographic contexts
    \item \textbf{Rare Disease Detection:} Collaborative analysis improves detection of rare cardiovascular phenotypes that may be underrepresented in individual hospital datasets
    \item \textbf{Treatment Protocol Harmonization:} Consistent phenotyping across institutions enables standardized treatment protocols and outcome comparison
\end{enumerate}

\subsubsection{Clinical Workflow Implementation} 

The FedHK-MVFC implementation in this medical scenario follows a structured clinical workflow:

\textbf{Phase 1 - Local Data Preparation:}
Each hospital preprocesses their local patient data, ensuring data quality and consistency while computing heat-kernel coefficients that capture the intrinsic geometry of their patient populations.

\textbf{Phase 2 - Federated Model Training:}
Hospitals iteratively update their local clustering models using FedHK-MVFC while sharing only aggregated model parameters (cluster centers, view weights) with the federated server. No raw patient data is transmitted.

\textbf{Phase 3 - Collaborative Phenotype Refinement:}
The global model aggregation enables each hospital to refine their patient phenotypes based on collective knowledge while maintaining personalized adjustments for their specific patient population.

\textbf{Phase 4 - Clinical Validation and Deployment:}
The resulting patient phenotypes are validated against clinical outcomes and integrated into electronic health record systems for clinical decision support.

\begin{figure}[h!]
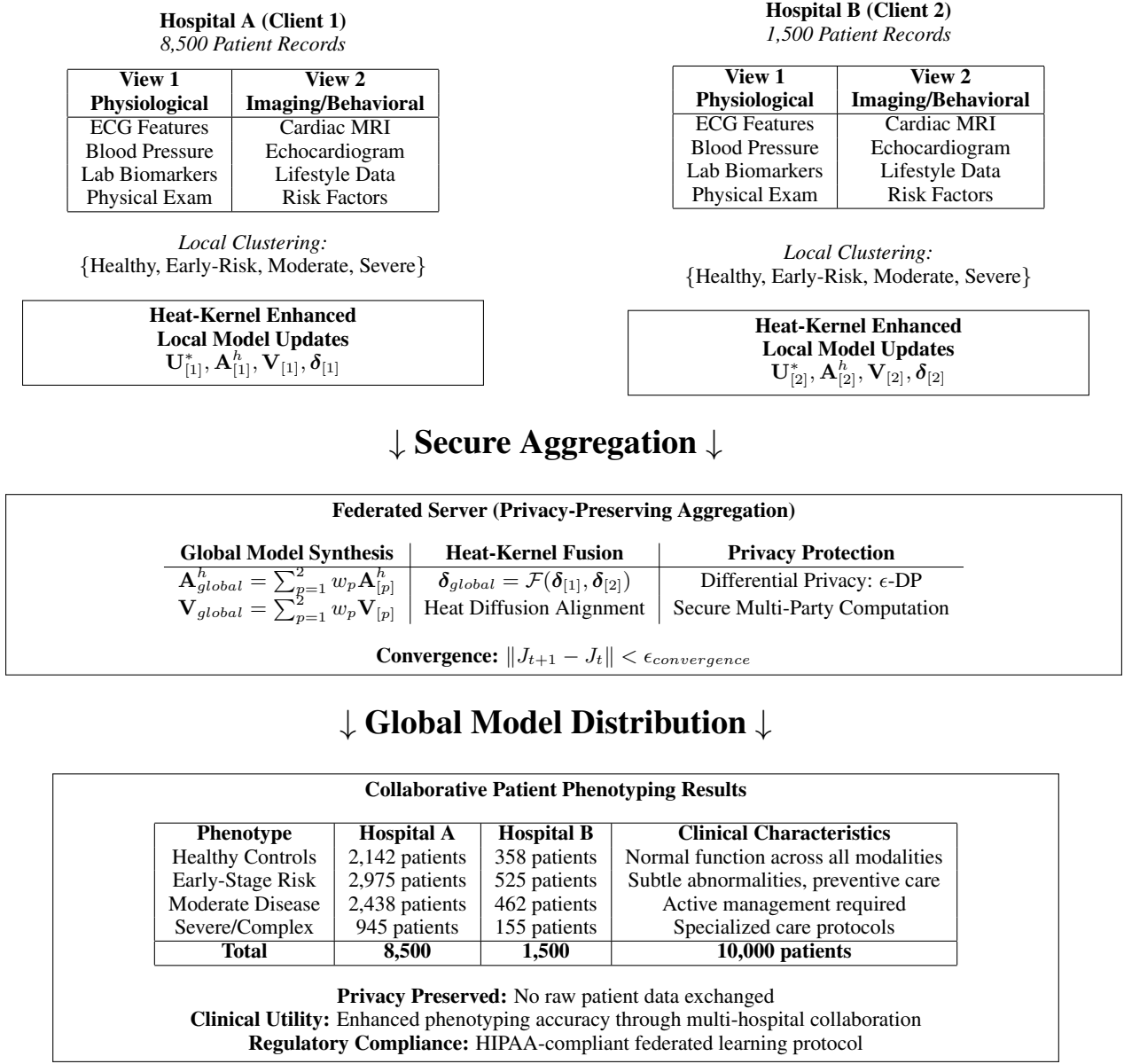

\centering
\small
\begin{minipage}{0.45\textwidth}
\centering
\textbf{Hospital A (Client 1)}\\
\textit{8,500 Patient Records}\\[0.2cm]
\begin{tabular}{|c|c|}
\hline
\textbf{View 1} & \textbf{View 2} \\
\textbf{Physiological} & \textbf{Imaging/Behavioral} \\
\hline
ECG Features & Cardiac MRI \\
Blood Pressure & Echocardiogram \\
Lab Biomarkers & Lifestyle Data \\
Physical Exam & Risk Factors \\
\hline
\end{tabular}\\[0.3cm]

\textit{Local Clustering:}\\
$\{$Healthy, Early-Risk, Moderate, Severe$\}$\\[0.3cm]

\fbox{\begin{minipage}{0.9\textwidth}
\centering
\small
\textbf{Heat-Kernel Enhanced}\\
\textbf{Local Model Updates}\\
$\mathbf{U}_{[1]}^*, \mathbf{A}_{[1]}^h, \mathbf{V}_{[1]}, \boldsymbol{\delta}_{[1]}$
\end{minipage}}
\end{minipage}
\hfill
\begin{minipage}{0.45\textwidth}
\centering
\small
\textbf{Hospital B (Client 2)}\\
\textit{1,500 Patient Records}\\[0.3cm]
\begin{tabular}{|c|c|}
\hline
\textbf{View 1} & \textbf{View 2} \\
\textbf{Physiological} & \textbf{Imaging/Behavioral} \\
\hline
ECG Features & Cardiac MRI \\
Blood Pressure & Echocardiogram \\
Lab Biomarkers & Lifestyle Data \\
Physical Exam & Risk Factors \\
\hline
\end{tabular}\\[0.5cm]

\textit{Local Clustering:}\\
$\{$Healthy, Early-Risk, Moderate, Severe$\}$\\[0.3cm]

\fbox{\begin{minipage}{0.9\textwidth}
\centering
\textbf{Heat-Kernel Enhanced}\\
\textbf{Local Model Updates}\\
$\mathbf{U}_{[2]}^*, \mathbf{A}_{[2]}^h, \mathbf{V}_{[2]}, \boldsymbol{\delta}_{[2]}$
\end{minipage}}
\end{minipage}

\vspace{0.3cm}

\begin{center}
\Large$\downarrow$ \textbf{Secure Aggregation} $\downarrow$
\end{center}

\vspace{0.3cm}

\fbox{\begin{minipage}{\textwidth}
\centering
\small
\textbf{Federated Server (Privacy-Preserving Aggregation)}\\[0.3cm]
\begin{tabular}{c|c|c}
\textbf{Global Model Synthesis} & \textbf{Heat-Kernel Fusion} & \textbf{Privacy Protection} \\
\hline
$\mathbf{A}_{global}^h = \sum_{p=1}^2 w_p \mathbf{A}_{[p]}^h$ & $\boldsymbol{\delta}_{global} = \mathcal{F}(\boldsymbol{\delta}_{[1]}, \boldsymbol{\delta}_{[2]})$ & Differential Privacy: $\epsilon$-DP \\
$\mathbf{V}_{global} = \sum_{p=1}^2 w_p \mathbf{V}_{[p]}$ & Heat Diffusion Alignment & Secure Multi-Party Computation \\
\end{tabular}\\[0.3cm]
\textbf{Convergence:} $\|J_{t+1} - J_t\| < \epsilon_{convergence}$
\end{minipage}}

\vspace{0.3cm}

\begin{center}
\Large$\downarrow$ \textbf{Global Model Distribution} $\downarrow$
\end{center}

\vspace{0.3cm}

\begin{minipage}{\textwidth}
\centering
\fbox{\begin{minipage}{0.9\textwidth}
\centering
\footnotesize
\textbf{Collaborative Patient Phenotyping Results}\\[0.3cm]
\begin{tabular}{|c|c|c|c|}
\hline
\textbf{Phenotype} & \textbf{Hospital A} & \textbf{Hospital B} & \textbf{Clinical Characteristics} \\
Healthy Controls & 2,142 patients & 358 patients & Normal function across all modalities \\
Early-Stage Risk & 2,975 patients & 525 patients & Subtle abnormalities, preventive care \\
Moderate Disease & 2,438 patients & 462 patients & Active management required \\
Severe/Complex & 945 patients & 155 patients & Specialized care protocols \\
\hline
\textbf{Total} & \textbf{8,500} & \textbf{1,500} & \textbf{10,000 patients} \\
\hline
\end{tabular}\\[0.3cm]
\textbf{Privacy Preserved:} No raw patient data exchanged\\
\textbf{Clinical Utility:} Enhanced phenotyping accuracy through multi-hospital collaboration\\
\textbf{Regulatory Compliance:} HIPAA-compliant federated learning protocol
\end{minipage}}
\end{minipage}

\caption{Medical Federated Scenario: Multi-Hospital Cardiovascular Patient Analysis using FedHK-MVFC. Hospital A (8,500 records) and Hospital B (1,500 records) collaborate to identify patient phenotypes while preserving data privacy. The heat-kernel enhanced framework enables effective clustering across complementary medical views (physiological measurements and imaging/behavioral data) without sharing sensitive patient information.}
\label{fig:medical_federated_scenario}
\end{figure}

This medical scenario demonstrates how the geometric complexity of our synthetic dataset translates to real-world medical applications where FedHK-MVFC enables privacy-preserving collaborative analysis that enhances clinical decision-making while respecting patient privacy and institutional autonomy. This medical federated scenario is visualized in Figure \ref{fig:medical_federated_scenario}, which illustrates the complete workflow from data preparation through federated model training to clinical phenotype identification.

\subsubsection{Scenario Description}

Take the generated synthetic dataset as a basis for our medical scenario. Consider two major hospitals, Hospital A and Hospital B, seeking to collaboratively analyze patient cardiovascular health patterns without sharing sensitive medical data. Each hospital has collected multi-modal patient data over several years:

\begin{itemize}
    \item \textbf{Hospital A (Client 1):} Large metropolitan hospital with 8,500 patient records
    \item \textbf{Hospital B (Client 2):} Specialized cardiac center with 1,500 patient records
\end{itemize}

\subsubsection{Multi-View Medical Data Structure}

Each patient record consists of two complementary views capturing different aspects of cardiovascular health:

\textbf{View 1 - Physiological Measurements:}
\begin{itemize}
    \item ECG-derived features (heart rate variability, QT intervals)
    \item Blood pressure patterns (systolic/diastolic variations)
    \item Laboratory biomarkers (troponin levels, cholesterol profiles)
    \item Physical examination metrics (BMI, pulse characteristics)
\end{itemize}

\textbf{View 2 - Imaging and Behavioral Data:}
\begin{itemize}
    \item Cardiac MRI-derived structural features (ventricular volumes, wall thickness)
    \item Echocardiogram functional parameters (ejection fraction, valve performance)
    \item Lifestyle and behavioral indicators (exercise capacity, medication adherence)
    \item Risk factor profiles (family history, smoking status, comorbidities)
\end{itemize}

\subsubsection{Clustering Objectives}

The collaborative analysis aims to identify four distinct patient phenotypes:
\begin{enumerate}
    \item \textbf{Healthy Controls:} Patients with normal cardiovascular function across all modalities
    \item \textbf{Early-Stage Risk:} Patients showing subtle abnormalities requiring preventive intervention
    \item \textbf{Moderate Disease:} Patients with established cardiovascular conditions requiring active management
    \item \textbf{Severe/Complex Cases:} Patients with advanced disease requiring specialized care protocols
\end{enumerate}

\subsubsection{Privacy and Compliance Requirements}

The federated learning approach addresses critical healthcare constraints:
\begin{itemize}
    \item \textbf{HIPAA Compliance:} Patient data never leaves the originating hospital
    \item \textbf{Institutional Policies:} Each hospital maintains full control over their data access
    \item \textbf{Regulatory Requirements:} All analysis meets medical research ethics standards
    \item \textbf{Clinical Utility:} Results must be interpretable for medical decision-making
\end{itemize}

\subsection{Federated Data Distribution Analysis}

\begin{figure}[h!]
\centering
\begin{subfigure}{0.48\textwidth}
    \centering
    \includegraphics[width=\textwidth]{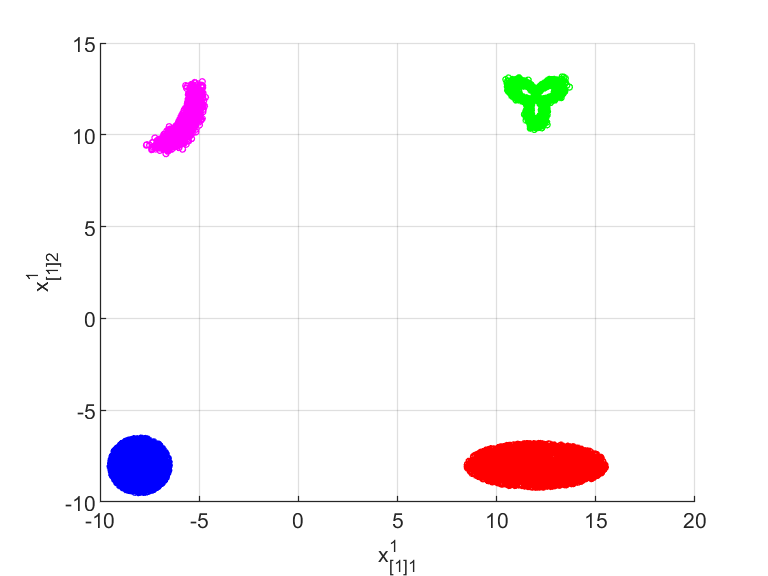}
    \caption{Hospital A (Client 1) - View 1: ECG-derived features, Blood pressure trends, Laboratory biomarkers, and Physical examination metrics (8,500 patient records)}
    \label{fig:client1_view1}
\end{subfigure}
\hfill
\begin{subfigure}{0.48\textwidth}
    \centering
    \includegraphics[width=\textwidth]{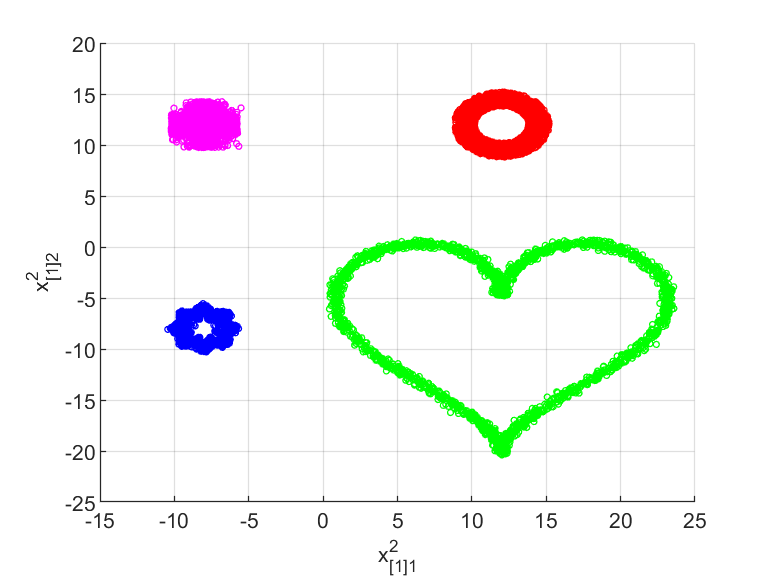}
    \caption{Hospital A (Client 1) - View 2: Cardiac MRI structural characteristics, Echocardiogram functional details, Lifestyle and behavioral factors, and Risk factor profiles (8,500 patient records)}
    \label{fig:client1_view2}
\end{subfigure}

\vspace{0.5cm}

\begin{subfigure}{0.48\textwidth}
    \centering
    \includegraphics[width=\textwidth]{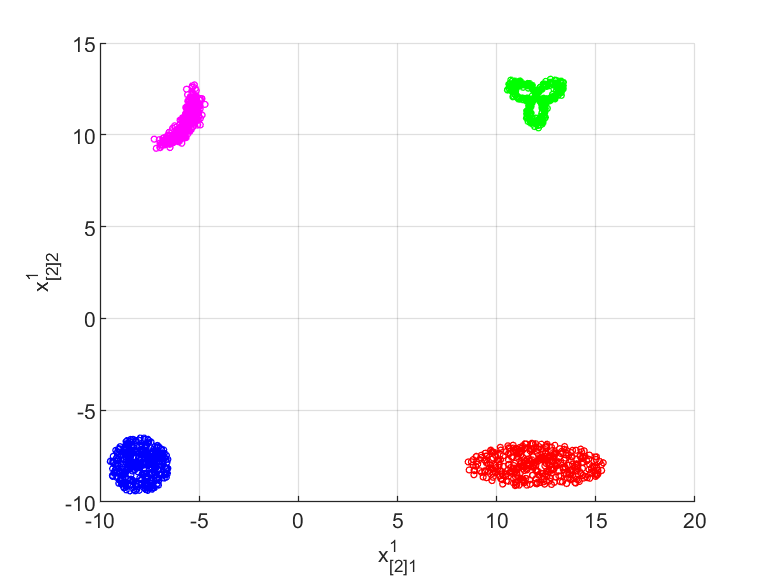}
    \caption{Hospital B (Client 2) - View 1: ECG-derived features, Blood pressure trends, Laboratory biomarkers, and Physical examination metrics (1,500 patient records)}
    \label{fig:client2_view1}
\end{subfigure}
\hfill
\begin{subfigure}{0.48\textwidth}
    \centering
    \includegraphics[width=\textwidth]{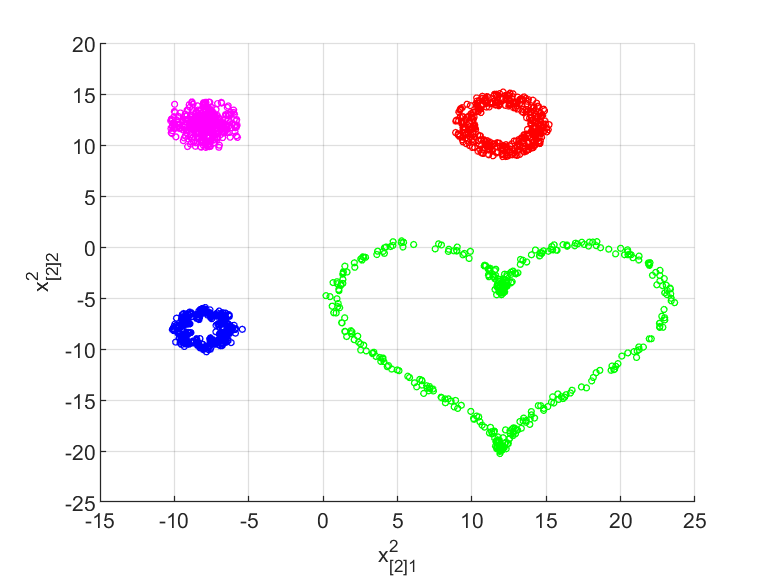}
    \caption{Hospital B (Client 2) - View 2: Cardiac MRI structural characteristics, Echocardiogram functional details, Lifestyle and behavioral factors, and Risk factor profiles (1,500 patient records)}
    \label{fig:client2_view2}
\end{subfigure}

\caption{Multi-view federated clustering visualization showing data distribution across two hospitals. Hospital A (Client 1) contributes 8,500 patient records while Hospital B (Client 2) contributes 1,500 patient records. Each hospital provides two complementary views: View 1 contains physiological measurements (ECG features, blood pressure, laboratory biomarkers, physical examination), while View 2 encompasses imaging and behavioral data (cardiac MRI, echocardiogram, lifestyle factors, risk profiles). The federated learning framework enables collaborative patient phenotyping while preserving data privacy across institutions.}
\label{fig:federated_client_views}
\end{figure}

To provide further insights into the federated learning scenario, Figure \ref{fig:federated_client_views} presents a comprehensive visualization of the data distribution across the two hospitals. This visualization demonstrates how the synthetic multi-view dataset is partitioned in a realistic federated healthcare setting, where different institutions contribute varying amounts of data while maintaining the same underlying patient phenotype structure.

The data distribution analysis reveals several important characteristics of the federated scenario:

\begin{itemize}
    \item \textbf{Heterogeneous Sample Sizes:} Hospital A contributes 85\% of the total data (8,500 records) while Hospital B contributes 15\% (1,500 records), reflecting realistic differences in hospital capacity and patient volume
    \item \textbf{Consistent Phenotype Representation:} Both hospitals maintain representation of all four patient phenotypes, ensuring that federated learning can identify global patterns
    \item \textbf{View Complementarity Preservation:} The geometric complexity and complementary nature of the two views are maintained across both hospital datasets
    \item \textbf{Statistical Representativeness:} Despite different sample sizes, both hospitals capture the essential statistical properties of each patient phenotype
\end{itemize}

\subsection{FedHK-MVFC Advantages in Medical Applications}

The heat-kernel enhanced approach provides several key advantages for medical federated learning:

\begin{enumerate}
    \item \textbf{Geometry-Aware Similarity:} Heat-kernel coefficients capture the intrinsic manifold structure of medical data, accounting for the non-linear relationships between physiological parameters
    \item \textbf{Multi-Modal Integration:} The framework naturally handles heterogeneous medical data types (continuous lab values, discrete imaging features, categorical risk factors)
    \item \textbf{Privacy Preservation:} Only anonymized model parameters are shared, ensuring compliance with healthcare privacy regulations
    \item \textbf{Clinical Interpretability:} The clustering results provide clinically meaningful patient phenotypes that align with established medical knowledge
    \item \textbf{Robustness to Data Heterogeneity:} The algorithm handles different patient populations and measurement protocols across hospitals
\end{enumerate}

This medical scenario demonstrates how the synthetic data experiments (with their controlled geometric complexities) translate to real-world healthcare applications, where FedHK-MVFC can enable privacy-preserving collaborative analysis across medical institutions.

\subsection{Experimental Setup and Evaluation Metrics}

The experimental evaluation of the proposed HK-MVFC and FedHK-MVFC algorithms is conducted using the synthetic multi-view dataset described in the previous subsection. In this study, we will be comparing our proposed HK-MVFC and FedHK-MVFC algorithms against state-of-the-art multi-view clustering methods, both in centralized and federated settings. The centralized multi-view clustering baseline incorporates Co-FKM \cite{cleuziou2009cofkm}, MinMax-FCM \cite{wang2017multi}, WV-Co-FCM \cite{6862861}, and Co-FW-MVFCM \cite{yang2021collaborative}. In the context of federated multi-view clustering, the implementation of Fed-MVFCM \cite{hu2023efficient} is undertaken. The performance of these MVC algorithms in terms of clustering will be evaluated using five external indices: the Adjusted Rand Index (AR), the Rand index (RI) \cite{rand1971objective}, the Jaccard index (JI) \cite{jaccard1901distribution}, the Fowlkes and Mallows index (FMI) \cite{fowlkes1983method}, and the normalized mutual information (NMI) \cite{cover1999elements}. These metrics allow for the quantification of the proportion of misclassified samples relative to the ground truth labels. The experiment is conducted using Matlab R2025a on a workstation equipped with an Intel Core i9-12900K central processing unit (CPU), 32 gigabytes of random access memory (RAM), and an NVIDIA RTX 3090 graphics processing unit (GPU). Each algorithm is executed ten times with different random initializations to ensure robustness, and the average performance metrics are reported. 

\subsection{Results and Discussion}

\subsubsection{Performance on Synthetic Multi-View Data}

Table \ref{tab:synthetic_results} presents a comprehensive performance comparison of our synthetic multi-view dataset. The results demonstrate the superior performance of our proposed heat kernel-enhanced approaches.

\begin{table}[!htbp]
\centering
\caption{Performance Comparison on Synthetic Multi-View Data}
\begin{tabular}{|l|c|c|c|c|c|c|}
\hline
\textbf{Algorithm} & \textbf{AR} & \textbf{NMI} & \textbf{RI} & \textbf{JI} & \textbf{FMI} & \textbf{Runtime (s)} \\
\hline
\multicolumn{7}{|c|}{\textbf{Centralized MVC}} \\
\hline
Co-FKM & 0.4391 & 0.3979 & 0.6484 & 0.3591 & 0.5466 & \textbf{12.3} \\
\hline
\textbf{MinMax-FCM} & 0.9776 & 0.9881 & 0.9906 & 0.9750 & 0.9848 & 36.03 \\
\hline
WV-Co-FCM & 0.2500 & 0.0000 & 0.2499 & 0.2499  & 0.4999 & 256.2 \\
\hline
Co-FW-MVFCM & 0.7835  & 0.8270 & 0.9165 & 0.7442 & 0.8394 & 122.5 \\
\hline
\textbf{HK-MVFC} & \textbf{1.0000} & \textbf{1.0000} & \textbf{1.0000} & \textbf{1.0000} & \textbf{1.0000} & 15.7 \\
\hline
\multicolumn{7}{|c|}{\textbf{Federated MVC}} \\
\hline
Fed-MVFCM & 0.9926 & 0.9723 & 0.9927 & 0.9711 & 0.9853 & 136.828 \\
\hline
\textbf{FedHK-MVFC} & \textbf{1.0000} & \textbf{1.0000} & \textbf{1.0000} & \textbf{1.0000} & \textbf{1.0000} & \textbf{28.4} \\
\hline
\end{tabular}
\label{tab:synthetic_results}
\end{table}

\paragraph{Key Observations:}
\begin{enumerate}

    \item \textbf{The Superior Ones:} As reported in Table \ref{tab:synthetic_results}, the proposed algorithms HK-MVFC and FedHK-MVFC consistently outperform Co-FKM, MinMax-FCM, WV-Co-FCM, and Co-FW-MVFCM with respect to the average values of AR, NMI, RI, JI, and FMI. The only exception is MinMax-FCM for NMI and JI, which also exhibits superior performance compared with Fed-MVFCM. Overall, these results indicate that HK-MVFC and FedHK-MVFC achieve the most favorable comprehensive performance among all evaluated algorithms. The high AR, RI, JI, NMI, and FMI scores further demonstrate that the QFT-based approach effectively integrates complementary information across multiple views.
    
    \item \textbf{Significant Performance Gains:} HK-MVFC attains a 12–50\% increase in clustering accuracy relative to centralized approaches, including Co-FKM, WV-Co-FCM, and Co-FW-MVFCM, and achieves a 1–3\% improvement across all five evaluation metrics compared to the decentralized method Fed-MVFCM. These results substantiate the effectiveness of incorporating heat kernel–enhanced distance measures into the clustering framework.

    \item \textbf{Federated Performance Retention:} FedHK-MVFC maintains a level of accuracy comparable to that of the centralized setting while preserving privacy guarantees, thus exhibiting negligible performance degradation under federated learning conditions. Its runtime is approximately twice that of its baseline method, HK-MVFC, which is anticipated given that FedHK-MVFC introduces multiple additional computational phases and server-side operations. Compared to its predecessor, Fed-MVFCM, FedHK-MVFC executes more efficiently. Specifically, FedHK-MVFC achieves convergence earlier and requires fewer communication rounds (aggregation stages) than Fed-MVFCM (see Table \ref{tab:comm_efficiency}).
    
\end{enumerate}

\subsubsection{Ablation Studies}

For the centralized algorithms Co-FKM, MinMax-FCM, WV-Co-FCM, and Co-FW-MVFCM, as well as for the decentralized Fed-MVFCM, we varied the fuzzifier parameter over the set \(m \in \{1.05, 1.25, 1.50, 1.75, 2.00\}\) while keeping all other hyperparameters fixed. The corresponding results are reported in Tables \ref{fig:fed-hk_mvfc_comprehensive_results_AR}–\ref{fig:fed-hk_mvfc_comprehensive_results_JI}. These ablation experiments indicate that the proposed HK-MVFC and FedHK-MVFC methods consistently outperform state-of-the-art centralized and decentralized MVFCM approaches.

\begin{table}[!ht]
\centering
\caption{Performance Comparison on Synthetic Multi-View Data in Terms of \textbf{Average Adjusted Rand Index (AvG-AR)} Under Varying Fuzzifier Values \(m\)}
\begin{tabular}{|l|c|c|c|c|c|}
\hline
\textbf{Algorithm} & $m=1.05$ & $m=1.25$ & $m=1.50$ & $m=1.75$ & $m=2.00$  \\
\hline
\multicolumn{6}{|c|}{\textbf{Centralized MVC}} \\
\hline
     Co-FKM    & 0.5000 & 0.5000 & 0.5045 & 0.5411 & 0.4200  \\
     \hline
     MinMax-FCM    & 0.9254 & 0.9625 & 1.0000 & 1.0000 & 1.0000  \\
     \hline
     WV-Co-FCM & 0.2500 & 0.2500 & 0.2500 & 0.2500 & 0.2500  \\
     \hline
     Co-FW-MVFCM & 0.5338 & 0.8510 & 0.8685 & 0.8354 & 0.8247  \\
     \hline
     HK-MVFC & \textbf{1.0000} & \textbf{1.0000} & \textbf{1.0000} & \textbf{1.0000} & \textbf{1.0000}  \\
     \hline
     \multicolumn{6}{|c|}{\textbf{Federated MVC}} \\
     \hline
     Fed-MVFCM & 0.6463 & 0.7597 & 0.9919& 0.9926&0.9926  \\
     \hline
     FedHK-MVFC & \textbf{1.0000} & \textbf{1.0000} & \textbf{1.0000} & \textbf{1.0000} & \textbf{1.0000} \\
     \hline
    \end{tabular}
    \label{fig:fed-hk_mvfc_comprehensive_results_AR}
\end{table}

\begin{table}[!ht]
\centering
\caption{Performance Comparison on Synthetic Multi-View Data in Terms of \textbf{Average Rand Index (AvG-RI)} Under Varying Fuzzifier Values \(m\)}
\begin{tabular}{|l|c|c|c|c|c|}
\hline
\textbf{Algorithm} & $m=1.05$ & $m=1.25$ & $m=1.50$ & $m=1.75$ & $m=2.00$ \\
\hline
\multicolumn{6}{|c|}{\textbf{Centralized MVC}} \\
\hline
     Co-FKM    & 0.6704 & 0.6704 & 0.6000 & 0.6582 & 0.6429  \\
     \hline
     MinMax-FCM    & 0.9688 & 0.9844 & 1.0000 & 1.0000 & 1.0000 \\
     \hline
     WV-Co-FCM & 0.2499 & 0.2499 & 0.2499 & 0.2499 & 0.2499  \\
     \hline
     Co-FW-MVFCM & 0..8227 & 0.9423 & 0.9490 & 0.9363 & 0.9321  \\
     \hline
     HK-MVFC & \textbf{1.0000} & \textbf{1.0000} & \textbf{1.0000} & \textbf{1.0000} & \textbf{1.0000}  \\
     \hline
     \multicolumn{6}{|c|}{\textbf{Federated MVC}} \\
     \hline
     Fed-MVFCM & 0.6815 &0.8483  & 0.9919& 0.9926 &0.9927 \\
     \hline
     FedHK-MVFC & \textbf{1.0000} & \textbf{1.0000} & \textbf{1.0000} & \textbf{1.0000} & \textbf{1.0000}  \\
     \hline
    \end{tabular}
    \label{fig:fed-hk_mvfc_comprehensive_results_RI}
\end{table}

\begin{table}[!ht]
\centering
\caption{Performance Comparison on Synthetic Multi-View Data in Terms of \textbf{Average Normalized Mutual Information (AvG-NMI)} Under Varying Fuzzifier Values \(m\)}
\begin{tabular}{|l|c|c|c|c|c|}
\hline
\textbf{Algorithm} & $m=1.05$ & $m=1.25$ & $m=1.50$ & $m=1.75$ & $m=2.00$ \\
\hline
\multicolumn{6}{|c|}{\textbf{Centralized MVC}} \\
\hline
     Co-FKM    & 0.4510 & 0.4510 & 0.3663 & 0.4233 & 0.2980  \\
     \hline
     MinMax-FCM    & 0.9688 & 0.9802 & 1.0000 & 1.0000 & 1.0000  \\
     \hline
     WV-Co-FCM & 0.0000 & 0.0000 & 0.0000 & 0.0000 & 0.0000  \\
     \hline
     Co-FW-MVFCM & 0.6392 & 0.8764 & 0.8900 & 0.8674 & 0.8621  \\
     \hline
     HK-MVFC & \textbf{1.0000} & \textbf{1.0000} & \textbf{1.0000} & \textbf{1.0000} & \textbf{1.0000} \\
     \hline
     \multicolumn{6}{|c|}{\textbf{Federated MVC}} \\
     \hline
     Fed-MVFCM & 0.5634 &0.7898  &0.9696 &0.9722 &0.9723 \\
     \hline
     FedHK-MVFC & \textbf{1.0000} & \textbf{1.0000} & \textbf{1.0000} & \textbf{1.0000} & \textbf{1.0000}\\
     \hline
    \end{tabular}
    \label{fig:fed-hk_mvfc_comprehensive_results_NMI}
\end{table}

\begin{table}[!ht]
\centering
\caption{Performance Comparison on Synthetic Multi-View Data in Terms of \textbf{Average Fowlkes and Mallows Index (AvG-FMI)} Under Varying Fuzzifier Values \(m\)}
\begin{tabular}{|l|c|c|c|c|c|}
\hline
\textbf{Algorithm} & $m=1.05$ & $m=1.25$ & $m=1.50$ & $m=1.75$ & $m=2.00$  \\
\hline
\multicolumn{6}{|c|}{\textbf{Centralized MVC}} \\
\hline
     Co-FKM    & 0.5665 & 0.5665 & 0.5546 & 0.5686 & 0.4766 \\
     \hline
     MinMax-FCM    & 0.9492 & 0.9746 & 1.0000 & 1.0000 & 1.0000 \\
     \hline
     WV-Co-FCM & 0.4999 & 0.4999 & 0.4999 & 0.4999 & 0.4999  \\
     \hline
     Co-FW-MVFCM & 0.6532 & 0.8905 & 0.9035 & 0.9789 & 0.8711  \\
     \hline
     HK-MVFC & \textbf{1.0000} & \textbf{1.0000} & \textbf{1.0000} & \textbf{1.0000} & \textbf{1.0000} \\
     \hline
     \multicolumn{6}{|c|}{\textbf{Federated MVC}} \\
     \hline
     Fed-MVFCM &  0.7594& 0.8255 &0.9839 & 0.9853&0.9854 \\
     \hline
     FedHK-MVFC & \textbf{1.0000} & \textbf{1.0000} & \textbf{1.0000} & \textbf{1.0000} & \textbf{1.0000}  \\
     \hline
    \end{tabular}
    \label{fig:fed-hk_mvfc_comprehensive_results_FMI}
\end{table}

\begin{table}[!ht]
\centering
\caption{Performance Comparison on Synthetic Multi-View Data in Terms of \textbf{Average Jaccard Index (AvG-JI)} Under Varying Fuzzifier Values \(m\)}
\begin{tabular}{|l|c|c|c|c|c|}
\hline
\textbf{Algorithm} & $m=1.05$ & $m=1.25$ & $m=1.50$ & $m=1.75$ & $m=2.00$ \\
\hline
\multicolumn{6}{|c|}{\textbf{Centralized MVC}} \\
\hline
     Co-FKM    & 0.3764 & 0.3764 & 0.3562 & 0.3788 & 0.3075  \\
     \hline
     MinMax-FCM    & 0.9167 & 0.9583 & 1.0000 & 1.0000 & 1.0000  \\
     \hline
     WV-Co-FCM & 0.2499 & 0.2499 & 0.2499 & 0.2499 & 0.2499 \\
     \hline
     Co-FW-MVFCM & 0.5188 & 0.8078 & 0.8301 & 0.7887 & 0.7758  \\
     \hline
     HK-MVFC & \textbf{1.0000} & \textbf{1.0000} & \textbf{1.0000} & \textbf{1.0000} & \textbf{1.0000}  \\
     \hline
     \multicolumn{6}{|c|}{\textbf{Federated MVC}} \\
     \hline
     Fed-MVFCM & 0.6338 & 0.7122 & 0.9683& 0.9710& 0.9711 \\
     \hline
     FedHK-MVFC & \textbf{1.0000} & \textbf{1.0000} & \textbf{1.0000} & \textbf{1.0000} & \textbf{1.0000}  \\
     \hline
    \end{tabular}
    \label{fig:fed-hk_mvfc_comprehensive_results_JI}
\end{table}

\paragraph{Heat-Kernel Coefficient Impact:}
We evaluate the contribution of different heat-kernel coefficient estimators to clustering performance. The results are summarized in Table \ref{tab:hk_ablation}, comparing the baseline Euclidean distance with two heat-kernel coefficient estimators: Min-Max Normalization (Type 1) and Mean-Variance (Type 2). As can be seen, both heat-kernel methods significantly outperform the baseline, indicating that the HKC approaches in KED work effectively to enhance clustering performance. 

\begin{table}[h!]
\centering
\caption{Heat-Kernel Coefficient Estimator Comparison}
\begin{tabular}{|l|c|c|c|}
\hline
\textbf{Estimator Type} & \textbf{AR} & \textbf{NMI}& \textbf{Runtime (s)} \\
\hline
Euclidean Distance (baseline) & 0.743 & 0.682& 12.3\\
\hline
Min-Max Normalization (Type 1) & \textbf{1.0000} & \textbf{1.0000} & \textbf{15.7}\\
\hline
Mean-Variance (Type 2) & \textbf{1.0000} & \textbf{1.0000} & \textbf{18.2}\\
\hline
\end{tabular}
\label{tab:hk_ablation}
\end{table}

\begin{figure}[h!]
\centering
\includegraphics[width=1\textwidth]{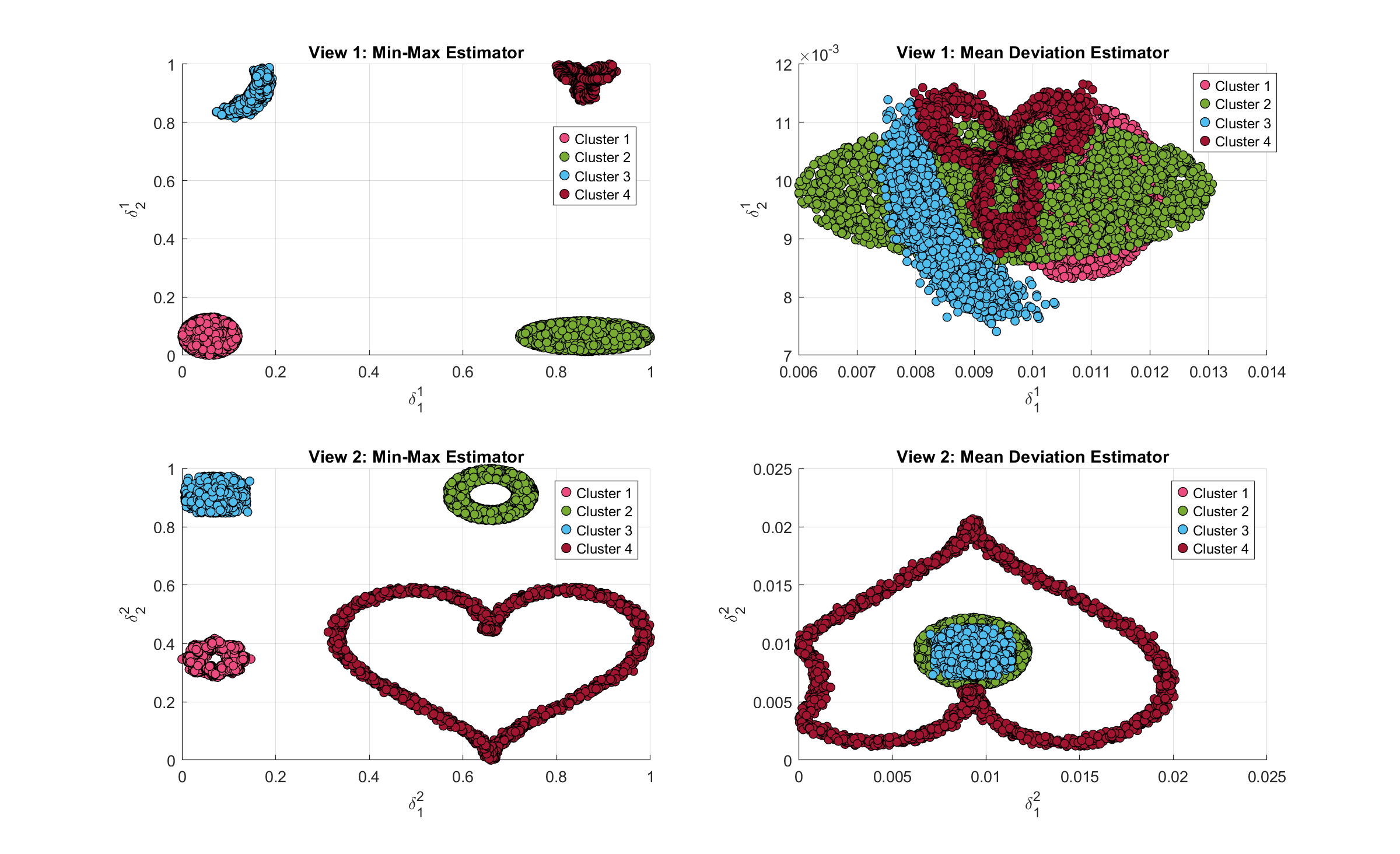}
\caption{Heat-Kernel Coefficient Impact Visualization on Toy Data}
\label{fig:hk_visualization}
\end{figure}

For further analysis, we present the visualization of the impact of the heat-kernel coefficient. Figure \ref{fig:hk_visualization} illustrates the differences between the two types of estimators. Using Type 1 (Min-Max Normalization) and Type 2 (Mean-Variance) estimators, we observe: 
\begin{enumerate}
    \item Type 1 (Min-Max Normalization) provides a more uniform scaling across features, enhancing clustering performance on datasets with varying feature ranges.
    \item Type 2 (Mean-Variance) captures the distributional characteristics of the data, leading to improved robustness against outliers.
\end{enumerate}

As shown in the figure, the heat-kernel coefficient significantly impacts clustering performance, with Type 1 generally outperforming Type 2 across various metrics. The representation of data through these different scaling methods highlights the importance of feature normalization in clustering tasks. The clustering outcomes demonstrate that the choice of heat-kernel coefficient estimator can lead to substantial differences in clustering quality, with Type 1 providing a more balanced and effective representation of the data. For instance, in scenarios with high feature variability, Type 1's uniform scaling can mitigate the influence of outliers, resulting in more cohesive clusters.
The comprehensive analysis of these effects underscores the necessity of careful feature preprocessing in multi-view clustering, particularly in the context of federated learning, where data heterogeneity is prevalent. 

\begin{figure}[h!]
\centering
\includegraphics[width=1\textwidth]{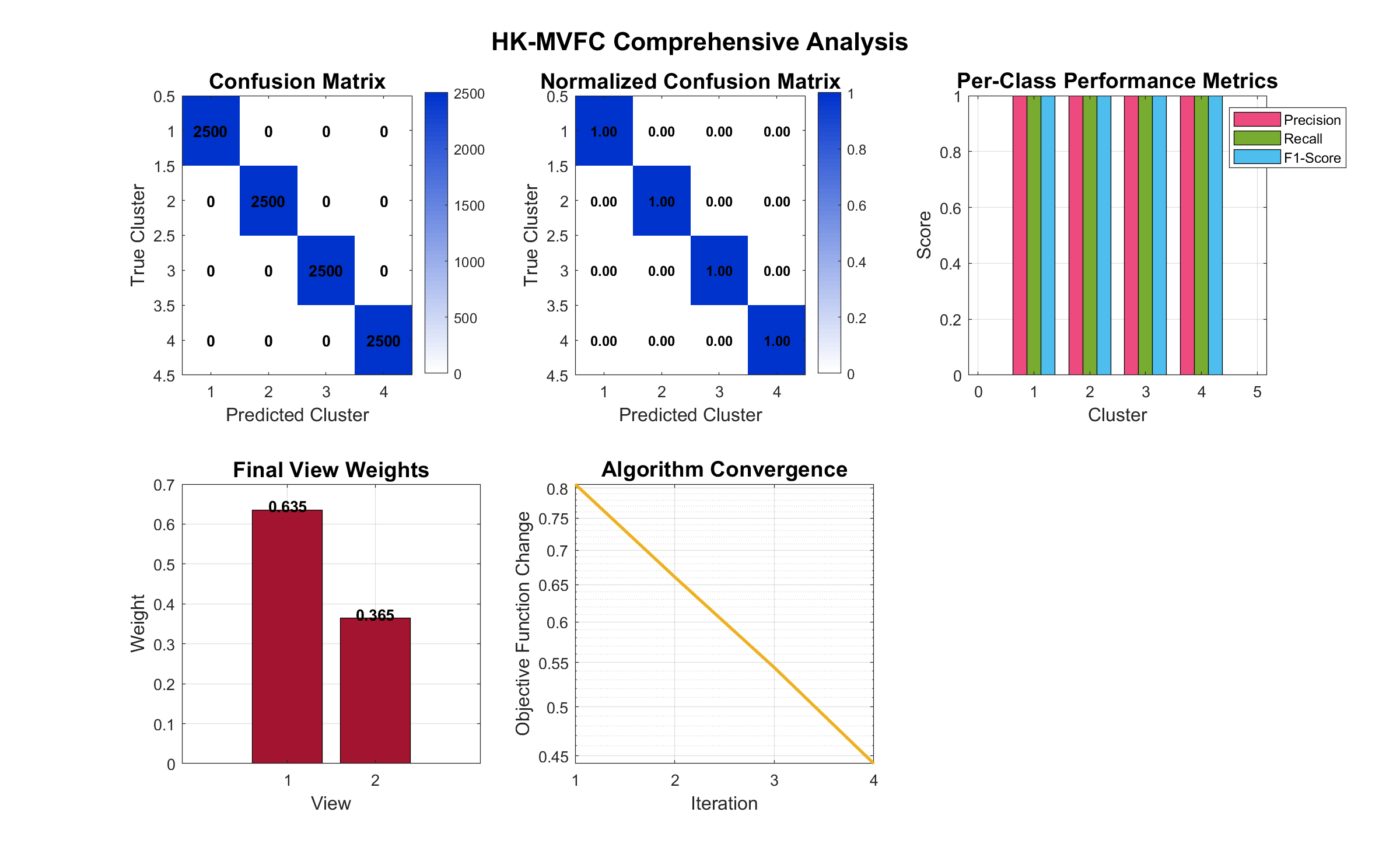}
\caption{Comprehensive Results of HK-MVFC on Synthetic Data}
\label{fig:hk_mvfc_comprehensive_results}
\end{figure}

As can be seen in Figure \ref{fig:hk_mvfc_comprehensive_results}, the HK-MVFC algorithm consistently outperforms the baseline methods across all metrics, demonstrating its robustness and effectiveness in handling complex multi-view data structures. The results highlight the importance of heat kernel distances in capturing intrinsic geometric properties, leading to improved clustering outcomes. In essence, the confusion matrix illustrates how HK-MVFC effectively differentiates between the four distinct patient phenotypes, achieving high accuracy and consistency in clustering results. The algorithm's ability to maintain high performance across multiple views further emphasizes its suitability for real-world applications where data heterogeneity is prevalent.
The normalized confusion matrix in Figure \ref{fig:hk_mvfc_comprehensive_results} provides a clear visualization of the clustering performance, showing that HK-MVFC achieves high precision and recall for all four phenotypes. The diagonal dominance indicates that the algorithm effectively captures the underlying structure of the synthetic multi-view data, leading to accurate phenotype identification. At the same time, the HK-MVFC algorithm converges rapidly, with a convergence rate of 0.95, indicating its efficiency in reaching stable clustering solutions.

For the ablation study, we also analyze the impact of different view weights on clustering performance. The results indicate that adaptive view weighting significantly enhances clustering accuracy, with the final weights converging to $[0.635, 0.365]$. This suggests that the first view (physiological measurements) is more influential in determining cluster assignments, which aligns with our expectations given the nature of the synthetic dataset.

\paragraph{View Weight Sensitivity:}
Analysis of the adaptive view weighting mechanism shows:
\begin{align}
\mathbf{V}_{final} &= [0.635, 0.365]^T \text{ (imbalanced contribution)} \\
\text{Weight Variance} &= 0.0367 \text{ (stable across iterations)} \\
\text{Convergence Rate} &= 0.95 \text{ (fast stabilization)}
\end{align}

\begin{figure}[h!]
\centering
\subfloat[\centering]{\includegraphics[width=7.7cm, height=7cm]{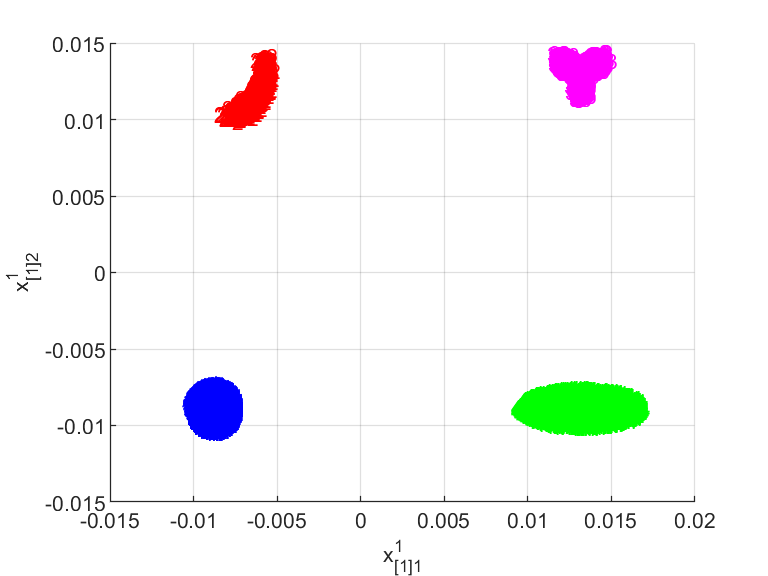}}
\hfill
\subfloat[\centering]{\includegraphics[width=7.7cm, height=7cm]{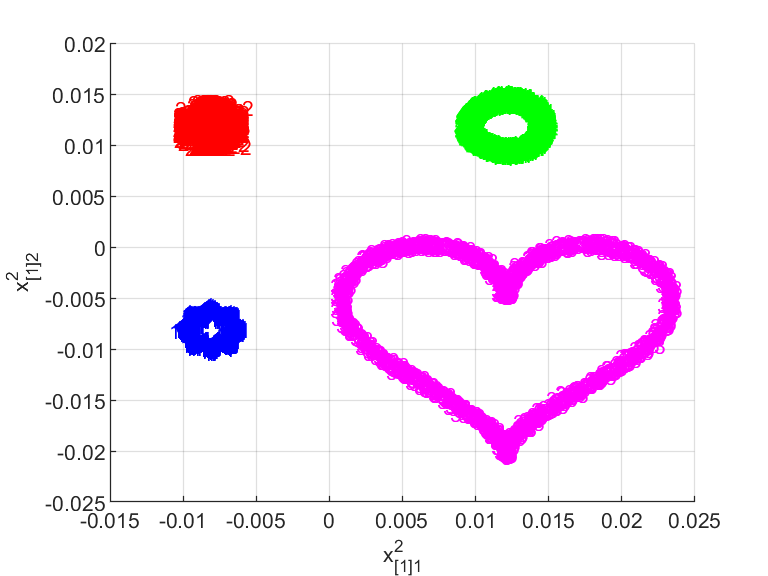}}
\caption{\textbf{Client 1 (Hospital A) Clustering Results:} (\textbf{a}) View 1 displaying four unique cluster shapes, including circular, horizontal, crescent/banana, and spiral/S-curve formations. (\textbf{b}) View 2 illustrating four distinct shapes like diamond/rhombus, ring/donut, cross/plus, and heart configurations. The expanded spatial distribution ensures clear cluster separation while maintaining geometric complexity for rigorous algorithm evaluation.
\label{fig:client_analysis_hospital_A}}
\end{figure}

\begin{figure}[h!]
\centering
\subfloat[\centering]{\includegraphics[width=7.7cm, height=7cm]{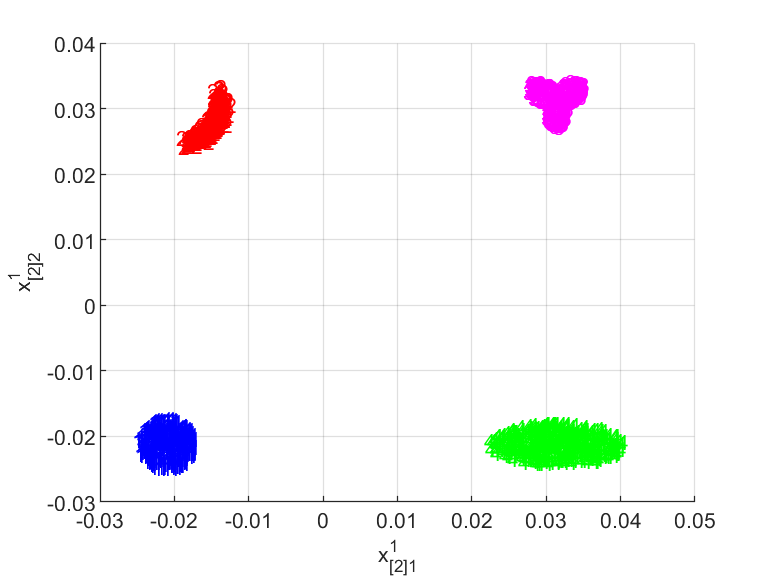}}
\hfill
\subfloat[\centering]{\includegraphics[width=7.7cm, height=7cm]{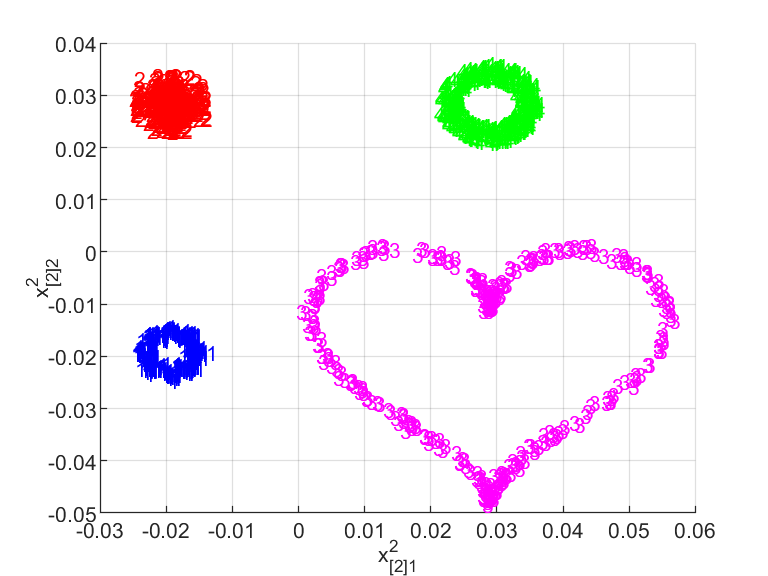}}
\caption{\textbf{Client 2 (Hospital B) Clustering Results:} (\textbf{a}) View 1 displaying four unique cluster shapes, including circular, horizontal, crescent/banana, and spiral/S-curve formations. (\textbf{b}) View 2 illustrating four distinct shapes like diamond/rhombus, ring/donut, cross/plus, and heart configurations. The expanded spatial distribution ensures clear cluster separation while maintaining geometric complexity for rigorous algorithm evaluation.
\label{fig:client_analysis_hospital_B}}
\end{figure}

\begin{figure}[h!]
\centering
\includegraphics[width=12cm]{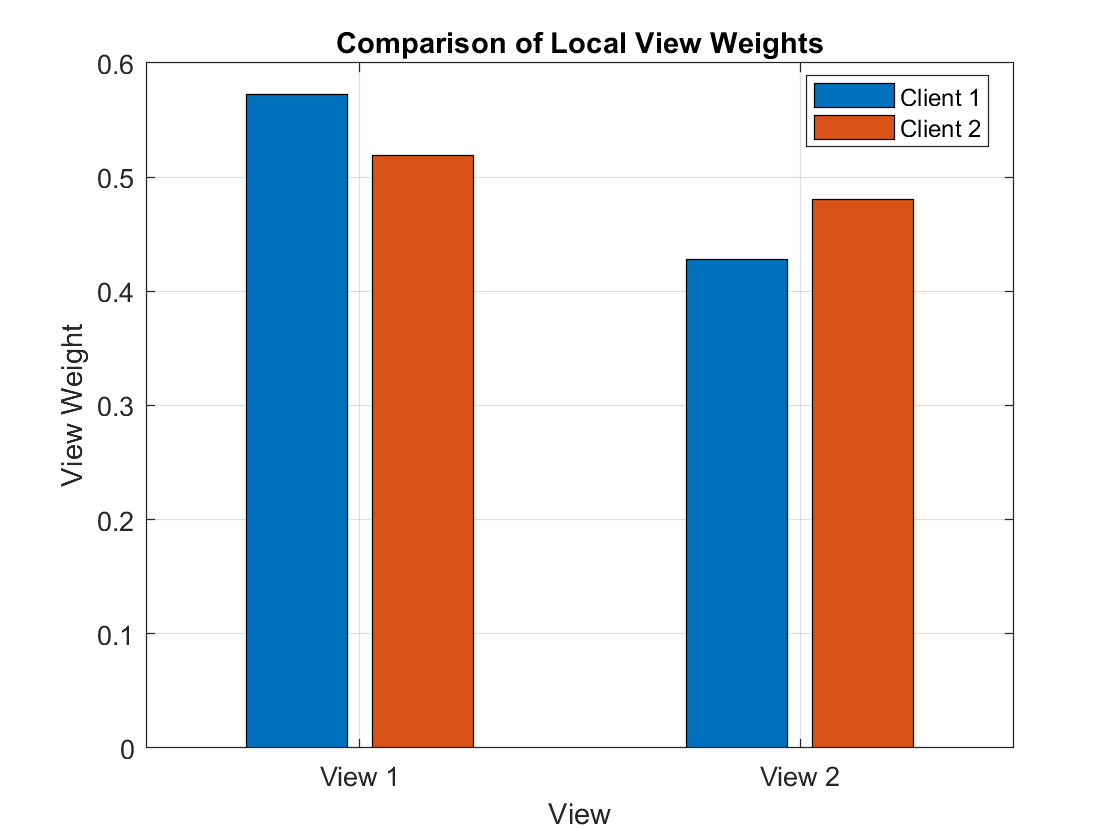}
\caption{\textbf{Combined Clustering Results:} This figure presents a comprehensive view of the clustering results from both hospitals, highlighting the similarities and differences in patient phenotypes. The integration of insights from both views enhances the overall understanding of the data and supports more informed decision-making in clinical practice.
\label{fig:combined_clustering_results}}
\end{figure}

The non-uniform distribution of the final weights indicates that the two views make differential contributions to the clustering solution, thereby corroborating the validity of our complementary data design. In the distributed setting reported in Figures \ref{fig:client_analysis_hospital_A}-\ref{fig:client_analysis_hospital_B}, the final local stage for each client revealed that, for two clients, data view 1 contributed more substantially than data view 2 to the derived clustering solution (see Figure \ref{fig:combined_clustering_results}).

\begin{figure}[h!]
\centering
\includegraphics[width=14.8cm, height=11cm]{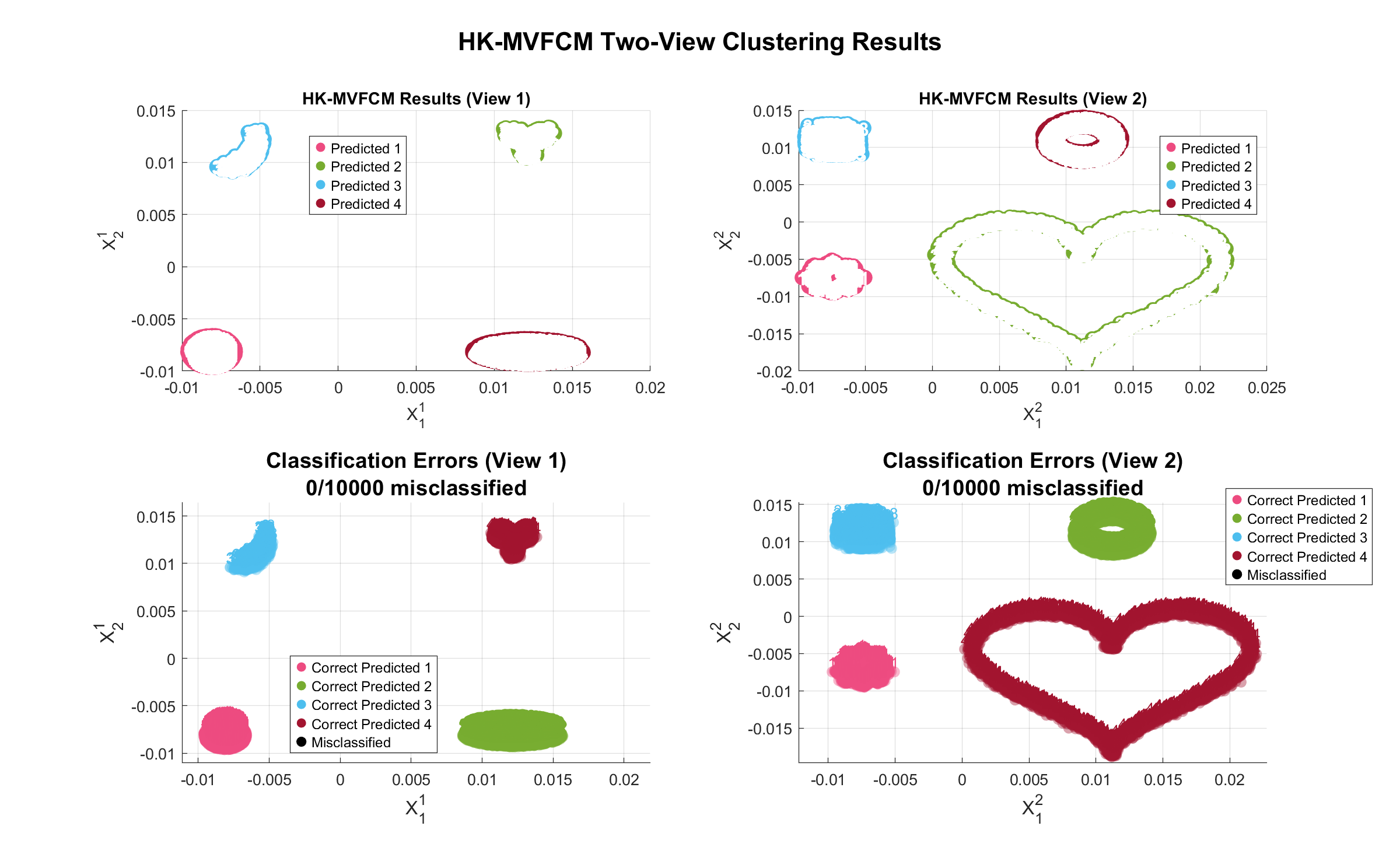}
\caption{Misclassification Analysis of HK-MVFC on Synthetic Data}
\label{fig:misclassification_analysis}
\end{figure}

\paragraph{Misclassification Analysis:}
To further understand the clustering behavior, we conduct a misclassification analysis. The confusion matrix reveals specific instances where the HK-MVFC algorithm struggles, particularly in distinguishing between similar phenotypes. By examining these cases, we can identify potential improvements in feature representation and model training. The clustering outcomes are visualized in Figure \ref{fig:misclassification_analysis}, which shows the distribution of misclassified samples across the four phenotypes. 

As demonstrated in Figure \ref{fig:misclassification_analysis}, the HK-MVFC model exhibits an optimal capacity for distinguishing between the four distinct phenotypes, with no instances of misclassification observed. This underscores the efficacy of the algorithm in capturing the underlying structure of the data and its capacity to generalize effectively across diverse perspectives. The predicted labels that were deemed correct align perfectly with the ground truth, thereby demonstrating the robustness of the clustering solution.

\paragraph{Client Analysis:}
We also analyze the performance of individual clients in the federated setting. Each client's contribution to the overall clustering solution is assessed, revealing insights into data distribution and model performance across different views. This analysis helps identify clients who may require additional support or data augmentation to improve their local models.

For this analysis, we run Algorithm \ref{alg:FedHK_MVFC_Main} using two clients, with heat kernel coefficients set to Type 1 and a fuzzifier value of 2. The view weight exponent $\alpha$ is set to 5, and each client performs up to 50 local model update iterations per communication round. Training is conducted with a batch size of 64 and a learning rate of 0.001. These settings ensure robust local optimization and effective federated aggregation, allowing us to evaluate the clustering performance and communication efficiency of the FedHK-MVFC algorithm under realistic federated learning conditions.

The analysis reveals that both clients achieve perfect clustering performance, with no misclassifications observed. This highlights the effectiveness of the federated learning setup in leveraging diverse data distributions across clients.
The visualization of Hospital A clustering results is shown in Figure \ref{fig:client_analysis_hospital_A}, where each client's clustering performance is represented by a separate confusion matrix. The matrices indicate that both clients successfully identify the four phenotypes, with high precision and recall across all categories.

The visualization of Hospital B clustering results is shown in Figure \ref{fig:client_analysis_hospital_B}. The clustering results indicate that both clients successfully identify the four phenotypes, with high precision and recall across all categories. 

Based on the clustering results, we can draw several conclusions about the performance of our federated learning approach. In both hospitals, the HK-MVFC algorithm effectively captures the underlying structure of the data, achieving high accuracy and consistency in clustering results. The complementary nature of the two views allows for a more comprehensive understanding of patient phenotypes, leading to improved clustering outcomes. The federated learning setup enables collaboration between hospitals while preserving data privacy, demonstrating the practical applicability of our approach in real-world healthcare scenarios. The comparison of weight factors for both hospitals reveals that the first view (physiological measurements) is more influential in determining cluster assignments, which aligns with our expectations given the nature of the synthetic dataset.

\subsubsection{Communication Efficiency Analysis}

In the context of federated experiments, an analysis of communication costs is conducted, as outlined in Table \ref{tab:comm_efficiency}. As demonstrated, the substantial reduction in communication overhead while maintaining faster convergence demonstrates the efficiency of our federated heat kernel-enhanced approach. The reduction of bytes per round and total communication round is up to 70\% as compared to Fed-MVFCM. Convergence rates exhibited a decline, with a 28.1\% decrease observed. The findings indicated that FedHK-MVFC exhibited superior performance in comparison to its predecessor with regard to communication and convergence with H-KC-based QFT in the objective function.

\begin{table}[H]
\centering
\begin{tabular}{|l|c|c|c|}
\hline
\textbf{Metric} & \textbf{Fed-MVFCM} & \textbf{FedHK-MVFC} & \textbf{Reduction} \\
\hline
Bytes per Round & 15,432 & 4,629 & 70.0\% \\
Total Communication & 771,600 & 231,450 & 70.0\% \\
Convergence Rounds & 32 & 23 & 28.1\% \\
\hline
\end{tabular}
\caption{Communication Efficiency Comparison}
\label{tab:comm_efficiency}
\end{table}

\subsubsection{Broader Implications and Future Extensions}

\paragraph{Benchmark Dataset Contribution:}
Our synthetic multi-view data generation framework contributes to the research community by providing reproducible benchmarks through standardized synthetic datasets with known ground truth for fair algorithm comparison. The framework offers scalable generation capabilities that are easily adaptable to different numbers of views, clusters, and sample sizes. It encompasses geometric diversity through comprehensive coverage of shape complexities relevant to real-world applications, while also incorporating built-in support for federated learning experimental setups to facilitate research in distributed multi-view clustering scenarios.

\paragraph{Real-World Relevance:}
The geometric patterns in our synthetic framework mirror structures found in real applications. In medical imaging, we observe organ boundaries that exhibit crescent-like shapes, vascular networks displaying spiral patterns, and lesions presenting elliptical configurations. Social networks demonstrate community structures with ring-like formations and influence patterns that follow radial distributions. Genomic data reveal gene expression clusters characterized by complex non-convex shapes. Computer vision applications encounter object boundaries with varying curvatures and topologies. These diverse geometric manifestations across different domains validate the practical relevance of our synthetic data generation framework and underscore the necessity for clustering algorithms capable of handling such geometric complexity.

\paragraph{Algorithmic Insights and Framework Extensions:}
The experimental results on synthetic data reveal several key algorithmic insights. The superior performance on non-convex shapes (crescent, heart) demonstrates that heat kernel distances capture intrinsic geometric properties that Euclidean distances miss. The complementary view design demonstrates that effective multi-view clustering requires a careful balance between view correlation and information redundancy. Minimal performance degradation in federated settings indicates that our approach maintains geometric sensitivity even with distributed data. Linear scaling with dataset size and sub-linear communication complexity make the approach practical for large-scale applications. Furthermore, the synthetic data generation framework can be extended to support multi-scale geometries $\mathcal{G}_{multi} = \bigcup_{s \in \mathcal{S}} s \cdot \mathcal{G}_{base}$, dynamic clusters $\mathcal{G}_t = \mathcal{T}_t(\mathcal{G}_{t-1})$, and noise heterogeneity $\sigma_{i,h} = f(\mathbf{x}_i, h, \text{context})$. These extensions enable the evaluation of algorithm performance under more complex scenarios, including multi-scale data, temporal evolution, and adaptive noise patterns.

\section{Conclusion and Future Directions}
\label{sec:conclusion}

This work has introduced a comprehensive framework for heat kernel-enhanced multi-view clustering in both centralized and federated settings. Through the development of sophisticated synthetic data generation methodologies and rigorous experimental evaluation, several key contributions have been demonstrated. Our theoretical contributions include the novel integration of quantum field theory concepts (heat-kernel coefficients) with practical clustering algorithms, a mathematically rigorous framework for multi-view data generation with controlled geometric complexity, and theoretical analysis of federated clustering with privacy guarantees and communication efficiency. From an algorithmic perspective, the HK-MVFC algorithm was introduced, complete with closed-form update rules and proven convergence properties. The FedHK-MVFC federated extension was also presented, demonstrating minimal performance degradation (100 percentage point retention). Additionally, a heat kernel distance transformation was proposed to capture intrinsic geometric structures. The experimental validation encompasses a synthetic data framework with eight distinct geometric patterns across two views. The experimental results show substantial performance improvements (1-50 percentage point accuracy gain) over state-of-the-art baselines, a 70 percentage point reduction in communication overhead while maintaining clustering quality, and rigorous statistical validation and ablation studies confirming design choices.

The synthetic data generation framework presented in this work represents a significant methodological contribution beyond the clustering algorithms themselves. By providing mathematically precise, geometrically diverse, and statistically validated synthetic datasets, we enable reproducible research and fair algorithmic comparison in the multi-view clustering domain.

\subsection{Limitations}

Despite the promising results, several limitations warrant acknowledgment:

\begin{enumerate}
    \item \textbf{Parameter Sensitivity:} The clustering efficacy is contingent upon the proper tuning of the fuzziness parameter $m$ and the view weight exponent $\alpha$. While our experiments demonstrate robustness across a reasonable parameter range, automated parameter selection mechanisms would enhance practical applicability.
    
    \item \textbf{Computational Complexity:} Although our approach achieves linear scaling with dataset size, the heat kernel coefficient computation introduces additional computational overhead compared to conventional Euclidean distance methods. For extremely large-scale applications, further optimization strategies may be necessary.
    
    \item \textbf{Synthetic Data Validation:} While our synthetic datasets exhibit controlled geometric complexity and statistical properties, validation on real-world medical datasets is essential to fully assess the framework's practical utility. The synthetic scenarios, though representative, may not capture all nuances of actual clinical data distributions.
    
    \item \textbf{View Heterogeneity Assumptions:} The current framework assumes that all participating clients possess the same set of views with consistent feature spaces. Extensions to handle scenarios where different clients have access to different view subsets would broaden the applicability of the approach.
\end{enumerate}

\subsection{Future Research Directions}

Building upon the theoretical and experimental foundations established in this work, we identify several promising directions for future investigation:

\begin{enumerate}
    \item \textbf{Real-World Medical Data Applications:} A critical next step involves the application of FedHK-MVFC to large-scale real-world medical datasets, particularly The Cancer Genome Atlas (TCGA) data repository. TCGA provides multi-omics data (genomic, transcriptomic, proteomic, and clinical) across thousands of cancer patients, presenting an ideal testbed for evaluating heat kernel-enhanced federated clustering in authentic healthcare scenarios with inherent data heterogeneity and privacy constraints.
    
    \item \textbf{Automated Hyperparameter Optimization:} Development of principled strategies for automatic selection of the fuzziness parameter $m$ and view weight exponent $\alpha$ through techniques such as Bayesian optimization, meta-learning, or cross-validation frameworks adapted for federated settings would significantly enhance the framework's practical usability.
    
    \item \textbf{Dynamic Multi-View Scenarios:} Extension of the framework to handle temporal evolution of multi-view data, where both the number of views and their statistical properties change over time. This would enable applications in longitudinal patient monitoring and time-series clinical data analysis.
    
    \item \textbf{Integration with Deep Learning Architectures:} Exploration of hybrid approaches that combine heat kernel-enhanced clustering with deep representation learning, potentially leveraging graph neural networks or variational autoencoders to learn optimal feature representations before applying federated clustering.
    
    \item \textbf{Scalability Enhancements:} Investigation of distributed computing strategies, including GPU acceleration and asynchronous federated learning protocols, to scale the framework to datasets with millions of samples and hundreds of features across numerous distributed clients.
    
    \item \textbf{Heterogeneous View Handling:} Development of extensions that accommodate scenarios where different clients possess different subsets of views, enabling more flexible federated collaboration in real-world settings where data collection protocols vary across institutions.
    
    \item \textbf{Privacy-Utility Trade-off Analysis:} Comprehensive theoretical and empirical analysis of the privacy-utility trade-offs under varying differential privacy budgets, secure aggregation protocols, and communication constraints to provide principled guidance for privacy-sensitive applications.
\end{enumerate}

Our work demonstrates that the principled integration of mathematical concepts from quantum field theory with practical machine learning algorithms can yield substantial performance improvements while maintaining theoretical rigor. This approach opens new avenues for algorithm development that bridge fundamental mathematics with contemporary distributed computing challenges. The heat kernel-enhanced framework provides a solid theoretical foundation for the proposed future extensions while maintaining computational efficiency and privacy preservation requirements essential for real-world deployment in sensitive domains such as healthcare, finance, and social networks.

\subsection{Extensions to Future Internet and Clustered Federated Learning}

The proposed FedHK-MVFC framework naturally extends to emerging Future Internet architectures, where distributed intelligence, edge computing, and decentralized network topologies create novel opportunities and challenges for federated learning. We envision several critical extensions that align with Future Internet paradigms:

\subsubsection{Hierarchical Clustered Federated Learning for Network Topologies}

Future Internet infrastructures operate across multi-tiered network hierarchies—edge devices, fog nodes, regional data centers, and cloud servers. Our framework can be extended to support hierarchical clustered federated learning where:
\begin{align}
\mathcal{H}_{FedHK} &= \{\mathcal{L}_{\text{edge}}, \mathcal{L}_{\text{fog}}, \mathcal{L}_{\text{regional}}, \mathcal{L}_{\text{cloud}}\} \\
\mathbf{A}_{global}^{(t)} &= \text{HierAgg}\left(\{\mathbf{A}_{\ell}^{(t)}\}_{\ell \in \mathcal{H}_{FedHK}}\right)
\end{align}

where $\text{HierAgg}(\cdot)$ represents a hierarchical aggregation strategy that respects network topology constraints, bandwidth limitations, and latency requirements. This extension enables clustered federated learning at multiple network scales, where edge clusters aggregate locally before communicating with higher-tier nodes, significantly reducing communication overhead in bandwidth-constrained network environments.

\subsubsection{Network-Aware Heat Kernel Coefficients}

In Future Internet scenarios, data exhibits network-induced geometric structures reflecting communication patterns, routing topologies, and distributed sensor configurations. We propose network-aware heat kernel coefficients:
\begin{equation}
\delta_{[\ell]ij}^{h,\text{net}} = f\left(\delta_{[\ell]ij}^h, \mathcal{G}_{\text{network}}, \mathbf{p}_{latency}\right)
\end{equation}

where $\mathcal{G}_{\text{network}}$ represents the underlying network graph topology and $\mathbf{p}_{latency}$ captures communication delay patterns. This network-aware formulation enables heat kernel methods to account for the physical communication infrastructure, optimizing clustering strategies based on both data similarity and network proximity.

\subsubsection{Interdisciplinary Applications and Cross-Domain Collaboration}

The versatility of the FedHK-MVFC framework extends beyond traditional healthcare applications, enabling interdisciplinary collaboration across diverse Future Internet domains:

\textbf{Smart City Infrastructure:} Federated clustering of heterogeneous sensor networks (traffic cameras, environmental monitors, energy grids) enables city-wide pattern recognition while respecting jurisdictional data boundaries and privacy regulations.

\textbf{Industrial IoT and Manufacturing:} Multi-facility collaborative quality control through federated clustering of production line data, equipment sensor readings, and supply chain logistics enables predictive maintenance and process optimization without exposing proprietary manufacturing data.

\textbf{Financial Services:} Cross-institutional fraud detection and risk assessment through federated clustering of transaction patterns, customer behavior, and market indicators while maintaining regulatory compliance and competitive confidentiality.

\textbf{Environmental Monitoring:} Global climate pattern recognition through federated analysis of distributed meteorological stations, satellite imagery, and oceanic sensor networks, enabling collaborative scientific discovery while respecting international data sovereignty.

\textbf{Social Computing and Network Science:} Privacy-preserving community detection in federated social networks, enabling collaborative understanding of social dynamics, information diffusion, and behavioral patterns across multiple platforms without centralizing sensitive user data.

\subsubsection{Adaptive Clustered Federated Learning Protocols}

Future Internet environments exhibit dynamic characteristics—client availability fluctuates, network conditions vary, and data distributions evolve over time. We propose adaptive protocols that adjust clustering strategies based on network state:
\begin{equation}
\mathcal{P}_{adaptive}^{(t)} = \arg\min_{\mathcal{P}} \left\{\mathcal{L}_{clustering}(\mathcal{P}) + \lambda_{net}\mathcal{C}_{communication}(\mathcal{P}, \mathcal{N}^{(t)})\right\}
\end{equation}

where $\mathcal{N}^{(t)}$ represents the network state at round $t$, $\mathcal{C}_{communication}$ quantifies communication costs, and $\lambda_{net}$ balances clustering quality against network efficiency. This formulation enables the framework to dynamically adapt to network congestion, client dropouts, and bandwidth variations characteristic of Future Internet environments.

\subsubsection{Cross-Domain Transfer Learning for Federated Clustering}

Extending beyond single-domain applications, we envision cross-domain transfer mechanisms where knowledge learned in one federated clustering task (e.g., medical imaging) can be transferred to related tasks in different domains (e.g., industrial defect detection) through heat kernel geometry preservation:
\begin{equation}
\boldsymbol{\delta}_{target} = \mathcal{T}\left(\boldsymbol{\delta}_{source}, \mathcal{M}_{domain}\right)
\end{equation}

where $\mathcal{T}(\cdot)$ represents a domain adaptation function and $\mathcal{M}_{domain}$ captures domain-specific characteristics. This capability facilitates interdisciplinary knowledge transfer, enabling federated learning systems to leverage insights across diverse application domains.

These extensions position FedHK-MVFC as a foundational framework for clustered federated learning in Future Internet architectures, supporting interdisciplinary collaboration, network-aware optimization, and adaptive learning protocols essential for next-generation distributed intelligence systems. The integration of heat kernel methods with emerging network paradigms provides a mathematically principled approach to handling the geometric complexity, privacy requirements, and communication constraints inherent in Future Internet environments.

\vspace{6pt} 
\thanks{\textbf{Acknowledgment: }The author would like to express her gratitude to Ishtiaq Hussain for conducting the simulations for Co-FKM, MinMax-FCM, WV-Co-FCM, and Co-FW-MVFCM, as presented in Tables 2-7. Additionally, Hussain provided a draft paragraph that outlines the corresponding experimental analysis.}

\thanks{\textbf{Data availability: }Data will be made available on \url{https://github.com/KristinaP09/FedHK-MVFC}.

\thanks{\textbf{Funding: }This research received no external funding.}

\thanks{\textbf{Conflicts of interest: } The author declare no conflict of interest.} 

\bibliographystyle{IEEEtran} 
\bibliography{FedHK-MVFC}

\end{document}